\documentclass[11pt]{article}
\usepackage[margin=1in]{geometry}
\usepackage{amsmath,amssymb,amsthm,amsfonts,mathrsfs}
\usepackage{graphicx}
\usepackage{float}
\usepackage{enumitem}
\usepackage{siunitx}
\usepackage{calc}
\usepackage{url}
\usepackage{booktabs}
\usepackage{subcaption}
\usepackage{titlesec}
\usepackage[labelfont=bf,textfont=it,size=small]{caption}
\usepackage{parskip}
\usepackage{authblk}
\usepackage{placeins}

\usepackage[dvipsnames,table,xcdraw]{xcolor}
\definecolor{DarkBlue}{RGB}{8,20,200}
\definecolor{LightBlue}{RGB}{222,235,247}
\definecolor{LightGreen}{RGB}{229,245,224}
\definecolor{DarkGreen}{RGB}{0,109,44}
\definecolor{lightgreen}{rgb}{0.3,0.9,0.6}
\definecolor{ISURed}{RGB}{153,0,0}
\definecolor{ISUGold}{RGB}{241,190,72}
\definecolor{LightGray}{RGB}{245,245,245}
\definecolor{BoxBlue}{RGB}{210,228,245}

\usepackage[T1]{fontenc}
\usepackage[utf8]{inputenc}
\usepackage[scaled]{helvet}
\usepackage{times}

\usepackage{tikz}
\usepackage{adjustbox}
\usepackage{pgfplots}
\usepackage[most]{tcolorbox}
\usepgfplotslibrary{fillbetween}
\usepgfplotslibrary{statistics}
\pgfplotsset{compat=1.18}
\usetikzlibrary{shapes.geometric, arrows.meta, positioning, fit, calc, backgrounds}

\usepackage[authoryear,round]{natbib}

\usepackage{multirow}
\usepackage{tabularx}

\DeclareGraphicsExtensions{.pdf,.png,.jpg}
\graphicspath{{Figures/}}

\titlespacing\section{0pt}{6pt plus 1pt minus 1pt}{3pt plus 1pt minus 1pt}
\titlespacing\subsection{0pt}{6pt plus 1pt minus 1pt}{6pt plus 1pt minus 1pt}
\titlespacing\subsubsection{0pt}{3pt plus 1pt minus 1pt}{3pt plus 1pt minus 1pt}
\titlespacing\paragraph{0pt}{6pt}{0.5em}
\titleformat{\section}{\large\bfseries\sffamily}{\thesection}{1em}{}
\titleformat{\subsection}{\normalsize\bfseries\sffamily}{\thesubsection}{1em}{}
\titleformat{\subsubsection}{\small\sffamily\bfseries}{\thesubsubsection}{1em}{}
\titleformat{\paragraph}[runin]{\normalfont\normalsize\bfseries}{\theparagraph}{0em}{}

\renewcommand\thesection{\arabic{section}}
\usepackage[pdfborder={0 0 0},colorlinks,allcolors=DarkBlue]{hyperref}
\hypersetup{
    colorlinks,
    linkcolor={DarkBlue},
    citecolor={DarkBlue},
    urlcolor={DarkBlue!80!black},
    plainpages=true
}
\newcommand{\colref}[3]{\hyperref[#2]{#1~\ref*{#2}{#3}}}
\newcommand{\cref}[2]{\hyperref[#2]{#1~\ref*{#2}}}
\newcommand{\figref}[1]{\cref{Figure}{#1}}
\newcommand{\secref}[1]{\cref{Section}{#1}}
\newcommand{\tableref}[1]{\cref{Table}{#1}}

\begin{document}

\title{Automated Maize Ear Phenotyping Using 3D Reconstructions}

\author[1]{\small Ritwesh A. Kumar}
\author[2]{\small  Som Tripathi}
\author[2,3]{\small  Peja Matthews}
\author[2,4]{\small  Srikar Reddy}
\author[2]{\small Talukder Zaki Jubery}
\author[5]{\small Patrick Schnable}
\author[2,6,*]{\small Adarsh Krishnamurthy}
\author[2,6,*]{\small Baskar Ganapathysubramanian}

\affil[1]{Department of Electrical and Computer Engineering, Iowa State University, Ames, IA 50011, USA}
\affil[2]{Translational AI Research and Education Center, Iowa State University, Ames, IA 50011, USA}
\affil[3]{Department of Mathematics and Computer Science, Fayetteville State University, Fayetteville, NC 28301, USA}
\affil[4]{School of Mechanical Engineering, Purdue University, West Lafayette, IN 47907, USA}
\affil[5]{Department of Agronomy, Iowa State University, Ames, IA 50011, USA}
\affil[6]{Department of Mechanical Engineering, Iowa State University, Ames, IA 50011, USA}
\affil[*]{Corresponding Authors: adarsh | baskar@iastate.edu}
\date{}
\maketitle

% ============================================================
\begin{abstract}
% The abstract should be a single paragraph of 250 words or less. It should be
% specific, telling why and how the study was made, what the results were, and
% why they were important. The abstract should read like a "mini-manuscript"
% with 1 to 2 sentences each for a justification/rationale, objective(s),
% methods, results, and conclusion.
% ============================================================

Maize kernel traits such as row number, kernels per row, and kernel size vary largely for genetic reasons and are consistently associated with regions of the genome that influence yield. Manual measurement of these traits, however, cannot keep pace with the volume of maize generated in a breeding program. To address this, we developed and validated a fully automated pipeline for extracting these traits from 3D point clouds of corn ears, built on a recently developed video-to-point-cloud platform. Raw video frames are processed through COLMAP and NeRF, the ear is isolated via density-based separation, and the point cloud is distance-calibrated to physical units. The calibrated ear point cloud was Z-axis aligned via PCA and cylindrically unwrapped to a 2D image. We enhanced contrast and performed zero-fine-tuning instance segmentation using Cellpose-SAM. A triple-juxtaposed unwrap strategy was used to prevent double-counting at the seam. The pipeline achieved kernel count $R^{2} = 0.921$ ($\mathrm{MAPE} = 10.33\%$) and kernel row number within $\pm$2 rows for 95.2\% of ears ($\mathrm{MAE} = 0.75$ rows) on a 168-ear held-out set from the 268-ear labeled dataset. The resulting multi-trait dataset has known genotype identity for each ear, positioning it for phenotype-to-genotype association analyses.

\end{abstract}

% ============================================================
\section*{Plain Language Summary}
% Please include a plain language summary (limit 1000 characters). The summary
% should be clear, concise, and free from jargon.

Counting and measuring kernels on corn ears by hand is slow and error-prone, limiting how quickly plant breeders can evaluate new crop varieties. We developed a high-throughput automated analysis pipeline that extracts eleven traits related to kernel count, size, shape, color, and packing geometry along the ear from base to tip, building on the low-cost turntable imaging and 3D reconstruction platform of \citet{young2026lowcost}. The system was applied to 1{,}091 corn ears, with accuracy validated on the 168-ear held-out set, achieving a kernel count $R^{2}$ of 0.921 and correctly identifying the number of kernel rows to within two rows for 160 out of 168 ears. Each ear has a known genotype identity. The resulting dataset of 1{,}091 ears spans heritable traits (e.g., kernel row number) and management-sensitive traits (e.g., kernels per row). This dataset is ready for genetic breeding and agronomic decision-making to pursue higher yield.

\pagebreak

\subsubsection*{Abbreviations}
\vspace{-1em}
\begin{table}[h!]
\begin{tabular}{l|l}
AABB& Axis-Aligned Bounding Box\\
AP& Axial Profile (mean hue and mean aspect ratio vs.\ height)\\
AR& Bounding Box Aspect Ratio\\
BB& Bounding Box\\
BPA& Ball-Pivoting Algorithm\\
CLAHE& Contrast Limited Adaptive Histogram Equalization\\
COLMAP& Structure-from-Motion and Multi-View Stereo pipeline\\
CPSAM& Cellpose-SAM\\
EMD& Earth-Mover Distance\\
FFT& Fast Fourier Transform\\
GT& Ground Truth\\
HSV& Hue Saturation Value\\
Hue& Circular Mean Kernel Hue\\
KC& Kernel Count\\
KNN& K Nearest Neighbor\\
KPR& Kernels Per Row\\
KRN& Kernel Row Number\\
MAE& Mean Absolute Error\\
MAPE& Mean Absolute Percentage Error\\
NeRF& Neural Radiance Field\\
Pack& Kernel Packing Traits (Gabriel graph neighbor count
      and mean neighbor distance)\\
PCA& Principal Component Analysis\\
RMSE& Root Mean Square Error\\
SA& Kernel Surface Area\\
SD& Standard Deviation\\
Signed ME& Signed Mean Error\\
Vol& Kernel Convex Hull Volume Proxy\\
\end{tabular}
\end{table}

% ============================================================
\section{Introduction}
%Keep the introduction short, but include (i) a brief statement of the problem that justifies doing the work, or the hypothesis on which it is based; (ii) the findings of others that will be further developed or challenged; and (iii) an explanation of the general approach and objectives. This last part may indicate the means by which the question was examined, especially if the methods are new.
% ============================================================

Corn (\textit{Zea mays} L.) is the world's most produced cereal crop and a primary target of modern breeding programs aimed at improving yield under increasingly variable environmental conditions~\citep{fao2023}. Yield in maize is largely determined by ear and kernel traits, including kernel row number (KRN), kernels per row (KPR), kernel size, and total kernel count, that carry substantial genetically determined variance and have been repeatedly associated with yield-related loci in quantitative trait locus (QTL) mapping and genome-wide association studies (GWAS)~\citep{peng2011, liu2015krn}. These traits, however, are not interchangeable. KRN is established early in ear development and exhibits high broad-sense heritability with consistent QTL associations~\citep{peng2011, liu2015krn}. In contrast, KPR reflects the degree to which each floret position is successfully filled during grain fill and is therefore more sensitive to environmental stress, resource availability, and management conditions~\citep{andrade2002}. The spatial organization of kernels on the ear reflects the timing and completeness of floret initiation and grain fill, and its genotype-by-environment (G$\times$E) component has been directly linked to the positional patterning of kernel development and abortion along the ear~\citep{oury2022}. Despite this informativeness, measuring these traits at the scale required by modern breeding programs remains a manual, labor-intensive bottleneck. It limits phenotypic throughput and introduces observer-dependent measurement error~\citep{makanza2018, araus2018}.

Automated phenotyping has advanced rapidly, with platforms spanning two-dimensional (2D) digital imaging, rotational scanning, and three-dimensional (3D) reconstruction. A fundamental limitation of single-image 2D approaches is that only one face of the ear is visible, leaving roughly half the kernels unobserved. Methods differ in how they handle this: some apply a blanket doubling factor assuming perfect bilateral symmetry~\citep{zhao2014}. In contrast, others count only the visible face and treat it as a yield proxy~\citep{makanza2018, shi2022}. Rotational scanning systems partially address occlusion by capturing the full ear surface. \citet{warman2021} developed a maize ear scanner that rotates an ear under a fixed camera, followed by digital flattening of the resulting video using a 2D panoramic projection, recovering per-kernel positional information from all sides of the ear. More recently, \citet{fan2026openear} introduced OpenEar, a DIY 360\textdegree{} video-based platform that uses YOLOv11-based deep learning segmentation on cylindrical surface projections to extract several ear- and kernel-level traits (including ear length, ear diameter, ear volume, kernel number, KRN, and KPR) on consumer-grade hardware. These approaches demonstrate that task-specific deep learning pipelines can achieve high-throughput, low-cost ear phenotyping. They nevertheless share two constraints: they require supervised training on manually annotated kernel and ear masks rather than zero-shot segmentation, and they operate on a 2D projection of the ear surface.

The reliance on a 2D projection is the more fundamental of these two limitations, as it constrains the set of traits that can be recovered. Projection of the ear surface onto a plane collapses the radial dimension, such that tightly packed or overlapping kernels cannot be distinguished by shape alone, and kernel length, width, and fill are conflated within a single silhouette~\citep{gonzalez2022, lu2025}. Reconstruction of the ear as a 3D surface avoids this loss. Although the cob-facing portion of each kernel is necessarily occluded from all lines of sight, the recovered outer surface patch retains radial depth and inter-kernel spacing, and thereby permits the extraction of relative shape descriptors and packing geometry that are not accessible from a projected representation. Such surface-based reconstruction provides a low-cost, readily deployable alternative to sub-surface imaging modalities such as micro-CT, albeit one that characterizes the exposed kernel cap rather than the complete kernel.

% ============================================================
% FULLY VERIFIED, OPTIMIZED METHOD COMPARISON (TABLE 1)
% ============================================================
\begin{table}[t!]
\centering
\small
\setlength{\tabcolsep}{3pt}
\caption{Comparative analysis of automated maize ear phenotyping methods. Abbreviations marked $^{a}$ are outputs of the cited method only; all others are defined in the Abbreviations section.}
\label{tab:method_comparison}
\renewcommand{\arraystretch}{0.95}
\begin{tabularx}{\textwidth}{@{} >{\raggedright\arraybackslash}p{3.5cm} >{\raggedright\arraybackslash}p{2.3cm} >{\raggedright\arraybackslash}p{1.6cm} >{\centering\arraybackslash}p{1.5cm} >{\raggedright\arraybackslash}X @{}}
\toprule
\textbf{Method} & \textbf{Modality} & \textbf{Hardware} & \textbf{Training} & \textbf{Outputs} \\
\midrule
\citet{makanza2018}
& 2D image & Consumer & No & EL$^{a}$, EW$^{a}$, KC, KW$^{a}$ \\[2pt]
\citet{warman2021}
& Rotational scan & Consumer & Yes & KPhen$^{a}$ \\[2pt]
\citet{gonzalez2022}
& 2D image & Consumer & No & EL$^{a}$, EW$^{a}$, TF$^{a}$, Tap$^{a}$, Cur$^{a}$, Col$^{a}$, KRN \\[2pt]
\citet{shi2022}
& 2D image & Specialist & Yes & KC, KRN, KPR \\[2pt]
\citet{maizeearSAM2025}
& 2D image & Consumer & No & KPR \\[2pt]
\citet{lu2025}
& 2D image & Consumer & Yes (PQT) & KC, KRN, KPR \\[2pt]
\citet{sun2026mep3d}
& Structured light & Specialist & No & KC, KRN, KPR, EL$^{a}$, ED$^{a}$, Barren Tip Length \\[2pt]
\citet{zhao2025}
& Micro-CT & Specialist & No & Vol, SA, ENDI$^{a}$, ENII$^{a}$, EMVSR$^{a}$, SCTI$^{a}$, ENDUI$^{a}$ \\
\citet{fan2026openear}
& 360$^\circ$ video (2D projection) & Consumer & Yes & EL$^{a}$, ED$^{a}$, EV$^{a}$, EW$^{a}$, KC, KRN, KPR, KT$^{a}$, KW$^{a}$, TKW$^{a}$ \\[2pt]
\midrule
\textbf{Ours} &
\textbf{360$^\circ$ video} &
\textbf{Consumer} &
\textbf{No} &
\textbf{\boldmath KC, KRN, KPR, SA, Vol, Pack$^{g}$, Hue, AR, AP$^{g}$} \\
\bottomrule
\multicolumn{5}{l}{\footnotesize $^{a}$Output of cited method only; see respective reference for definition.}\\
\multicolumn{5}{l}{\footnotesize $^{g}$Pack and AP each comprise two distinct traits (see Abbreviations), giving eleven in total.}\\
\end{tabularx}
\end{table}

Three-dimensional methods address the projection limitation directly, but existing systems sit at one of two extremes. Structured-light scanning achieves sub-millimeter kernel segmentation and morphological trait extraction from full-surface point clouds~\citep{sun2026mep3d}, and micro-CT imaging enables simultaneous recovery of internal kernel volume, surface area, and density parameters across diverse inbred populations~\citep{zhao2025}. Both recover kernel-level 3D geometry, but require specialist instrumentation that constrains deployment scale. At the opposite extreme, a low-cost stationary-camera neural radiance field (NeRF) pipeline has been used to recover skeleton length, convex hull volume, and cross-sectional shape descriptors from a maize diversity panel (300 ears processed, 250 passing quality control) using consumer-grade hardware and a cylindrical fiducial for automatic metric scaling~\citep{young2026lowcost}. This approach demonstrates that 3D reconstruction is viable at breeding scale, though the traits recovered are ear-level rather than kernel-level. Instance segmentation has been applied separately to maize ears, but only for a single kernel-level trait~\citep{maizeearSAM2025}. No existing pipeline jointly measures kernel count, volumetric shape proxies, and packing traits (specifically the number of physically touching kernel neighbors and their mean inter-kernel distances) from a single, consumer-accessible imaging workflow without supervised training. Kernel color and shape are informative but remain unaddressed by existing 3D pipelines: carotenoid concentration varies with kernel position along the cob~\citep{calvobrenes2019}, so axial gradients in hue and elongation carry a developmental signal that a single per-ear mean discards. \tableref{tab:method_comparison} summarizes existing automated phenotyping methods and situates the present work among them.

In this study, we develop and validate a fully automated pipeline for extracting eleven ear- and kernel-level phenotypic traits from 3D point clouds of corn ears, extending the corn ear NeRF reconstruction method of \citet{young2026lowcost} for trait extraction. Our contributions fall along three axes (method, validation, and resource), as detailed below. The specific contributions of this work are as follows:
\begin{enumerate}[leftmargin=*, itemsep=2pt, topsep=3pt]
\item \textbf{A training-free segmentation pipeline for maize kernels.} The calibrated ear point cloud is aligned to its principal axis via principal component analysis (PCA), cylindrically unwrapped to a 2D image, contrast-enhanced via contrast limited adaptive histogram equalization (CLAHE), and segmented into individual kernel instances using the zero-fine-tuned Cellpose-SAM model. A triple-juxtaposed unwrap strategy resolves the seam introduced by cylindrical projection, ensuring each kernel is counted exactly once.
\item \textbf{Joint extraction of eleven ear- and kernel-level traits.} Segmented kernels are re-projected to 3D point sets, from which kernel count, kernel row number (KRN), kernels per row (KPR), per-kernel surface area, per-kernel volume proxy, kernel packing geometry, kernel hue, and kernel aspect ratio are measured. Hue and aspect ratio are additionally profiled in 10\% height bands along the ear's principal axis, characterizing axial gradients in kernel color and shape from base to tip.
\item \textbf{Isolated per-trait validation against known ground-truth geometry.} Accuracy is assessed on a 27-ear synthetic dataset with exactly known generating geometry, a six-ear manually annotated subset annotated exhaustively at the level of individual kernels, and a 168-ear held-out set of manually annotated ears. This multi-tiered approach isolates segmentation and trait-extraction error from physical measurement error.
\item \textbf{A breeding-scale phenotypic dataset.} The pipeline is applied to 1{,}091 ears of known genotype identity, yielding a multi-trait dataset positioned for phenotype-to-genotype association analyses in future studies.
\end{enumerate}
Together, these contributions establish that eleven traits spanning count, shape, color, and packing geometry are recoverable from 360\textdegree{} video alone. The pipeline requires neither specialist instrumentation nor supervised, crop-specific models. Its throughput is compatible with breeding-scale deployment.

% Figure 1
\begin{figure}[t!]
  \centering
  \usetikzlibrary{positioning,arrows.meta,backgrounds,calc}

\definecolor{cRecon}{HTML}{2C6FBB}
\definecolor{cSeg}{HTML}{0097A7}
\definecolor{cMethod}{HTML}{D9821A}
\definecolor{cHub}{HTML}{6E54AB}
\definecolor{cTrait}{HTML}{9C6B12}
\definecolor{cData}{HTML}{2E8B57}
\definecolor{cPrior}{HTML}{2C6FBB}

\resizebox{\linewidth}{!}{
\renewcommand{\familydefault}{\sfdefault}
\begin{tikzpicture}[
  font=\footnotesize,
  >={Stealth[length=2.1mm]},
  base/.style ={rounded corners=2pt, draw, align=center, inner sep=3pt,
                minimum height=13mm, text width=24mm, line width=0.5pt},
  input/.style ={base, draw=cPrior, fill=cPrior!18, minimum height=14mm, text width=26mm, font=\footnotesize},
  recon/.style ={base, draw=cRecon,  fill=cRecon!22},
  seg/.style   ={base, draw=cSeg,    fill=cSeg!24},
  method/.style={base, draw=cMethod, fill=cMethod!30},
  hub/.style   ={base, draw=cHub,    fill=cHub!24},
  trait/.style ={rounded corners=2pt, draw=cData, fill=cData!30, align=center,
                 inner sep=4pt, minimum height=8mm, text width=40mm,
                 line width=0.5pt, font=\scriptsize\bfseries},
  card/.style  ={rounded corners=2pt, draw=cData, fill=cData!30, align=left,
                 inner sep=4pt, minimum height=8mm, text width=40mm,
                 line width=0.5pt, font=\scriptsize},
  data/.style  ={rounded corners=2pt, draw=cData, fill=cData!30, align=center,
                 inner sep=4pt, line width=0.9pt, text width=138mm, font=\footnotesize},
  arr/.style   ={->, line width=0.6pt, draw=black!72},
  elab/.style  ={font=\scriptsize\itshape, text=black!55, inner sep=1.3pt,
                 fill=white, fill opacity=0.85, text opacity=1, align=center, text width=18mm},
  legend_sw/.style={rectangle, draw=black!40, line width=0.4pt, minimum width=4mm, minimum height=3mm},
]

% ============ Stage 1 : Alignment (left -> right) [y = 0.0] ============
\node[input] (vid)    at (1.5,0) {\textbf{Input:}\\Metric-calibrated (mm)\\3D ear point cloud};
\node[seg]   (pca)    at (12,0)  {PCA\\$Z$-axis\\alignment};

% ============ Stage 2 : unwrap + segment (right -> left) [y = -2.5] ============
\node[seg] (unwrap) at (12,-2.5) {Cylindrical\\unwrap\\(+ index map)};
\node[seg] (clahe)  at (9,-2.5)  {CLAHE\\(L channel)};
\node[seg] (triple) at (6,-2.5)  {Triple-\\juxtapose\\(seam handling)};
\node[seg] (sam)    at (3,-2.5)  {Cellpose-SAM\\zero-shot};
\node[seg] (map)    at (0,-2.5)  {2D\,$\rightarrow$\,3D\\segment\\mapping};

\node[trait](kc)    at (3,-4.5)  {Total kernel count};

% ============ Stage 3 : per-kernel trait extraction [Baseline y = -9.0] ============
\node[hub]   (pts) at (0,-9.0)  {Per-kernel\\3D point set};
\node[hub]   (bpa) at (3,-9.0)  {BPA\\mesh};

% --- Upper Fork ---
\node[hub]   (sk)  at (9,-6.0)  {Surviving\\points $\mathcal{S}_k$};
\node[card] (morph) at (12.8,-6.0)
  {\textbf{Per-kernel morphology}\\[1pt]
   \textbullet\ Surface area\\
   \textbullet\ Volume \,(convex hull)\\
   \textbullet\ Mean hue \,(circular)\\
   \textbullet\ Aspect ratio \,(AABB)\\
   \textbullet\ Hue vs.\ height \,(10\% bands)\\
   \textbullet\ Aspect ratio vs.\ height \,(10\% bands)};
   
% --- Lower Fork ---
\node[hub]   (ck)  at (6,-12.0) {Area-weighted\\centroids $\mathbf{c}_k$};

\node[method](gab) at (9,-10.0) {Gabriel graph\\(periodic)};
\node[method](fft) at (9,-12.0) {$\theta$-histogram\\+ adaptive FFT};
\node[method](rc)  at (9,-14.0) {Row-chain\\graph};

\node[card] (pack) at (12.8,-10.0)
  {\textbf{Kernel packing}\\[1pt]
   \textbullet\ Neighbor count\\
   \textbullet\ Mean neighbor distance};
\node[trait](krn) at (12.8,-12.0) {KRN \,(row number)};
\node[trait](kpr) at (12.8,-14.0) {KPR \,(kernels / row)};

% ============ Outcome banner ============
\node[data] (out) at (6.9,-16.0)
  {\hyphenpenalty=10000\exhyphenpenalty=10000
   \textbf{Output: Eleven ear- and kernel-level traits} (count, KRN, KPR, 6 morphology, 2 packing)\\
   $\Rightarrow$ multi-trait dataset over 1{,}091 ears with known \mbox{genotype}
   $\Rightarrow$ phenotype $\rightarrow$ \mbox{genotype} (QTL / GWAS) association};

% ===================== connectors =====================
\draw[arr] (vid) -- (pca);
\draw[arr] (pca) -- (unwrap);
\draw[arr] (unwrap) -- (clahe);
\draw[arr] (clahe) -- (triple);
\draw[arr] (triple) -- (sam);
\draw[arr] (sam) -- (map);
\draw[arr] (sam.south) -- (kc.north);
\draw[arr] (map.south) -- (pts.north);
\draw[arr] (pts) -- (bpa);

\draw[arr] (bpa.east) -- (4.5,-9.0) |- (sk.west)
  node[elab, above] at ($(4.5,-6.0)!0.5!(sk.west)$) {largest\\connected\\component};
\draw[arr] (bpa.east) -- (4.5,-9.0) |- (ck.west);

\draw[arr] (sk.east) -- (morph.west);

\draw[arr] (ck.east) -- (7.5,-12.0) |- (gab.west);
\draw[arr] (ck.east) -- (fft.west);
\draw[arr] (ck.east) -- (7.5,-12.0) |- (rc.west);

\draw[arr] (gab.east) -- (pack.west);
\draw[arr] (fft.east) -- (krn.west);
\draw[arr] (rc.east)  -- (kpr.west);

% ===================== background bands + dashed box =====================
\begin{scope}[on background layer]
  \fill[cRecon!4,  rounded corners=3pt] (-1.4,-0.85) rectangle (15.1, 0.85);
  \fill[cSeg!4,    rounded corners=3pt] (-1.4,-5.10) rectangle (15.1,-1.65);
  \fill[cMethod!3, rounded corners=3pt] (-1.4,-15.15) rectangle (15.1,-5.30);

  % Outer boundary scaled up safely to encase the inputs with clean padding margins
  \draw[
    cPrior,
    dashed,
    line width=0.9pt,
    rounded corners=5pt
  ] (-0.2,-1.0) rectangle (3.2, 1.0);

  \node[
    anchor=south,
    font=\scriptsize\itshape,
    text=cPrior,
    fill=white,
    fill opacity=0.85,
    text opacity=1,
    inner sep=2pt
  ] at (1.5, 1.0) {adapted from Young et al.\ (2025)};
\end{scope}

% ===================== legend: color -> section mapping =====================
% Shifted to x=-3.2 to completely clear the "Area-weighted centroids" column
\begin{scope}[shift={(-3.2,-11.0)}]
  \node[legend_sw, draw=cPrior,   fill=cPrior!18, dashed, line width=0.6pt] (lg0) at (0,0.42) {};
  \node[anchor=west, font=\scriptsize] at (lg0.east) {\ Calibrated input stage};

  \node[legend_sw, draw=cSeg,    fill=cSeg!24]    (lg1) at (0,0)      {};
  \node[anchor=west, font=\scriptsize] at (lg1.east) {\ \secref{sec:preprocessing}\ Preprocessing Steps};

  \node[legend_sw, draw=cHub,    fill=cHub!24]    (lg2) at (0,-0.42)  {};
  \node[anchor=west, font=\scriptsize] at (lg2.east) {\ Intermediate 3D geometries};

  \node[legend_sw, draw=cMethod, fill=cMethod!30] (lg3) at (0,-0.84)  {};
  \node[anchor=west, font=\scriptsize] at (lg3.east) {\ Trait extraction algorithms};

  \node[legend_sw, draw=cData,   fill=cData!30]   (lg4) at (0,-1.26)  {};
  \node[anchor=west, font=\scriptsize] at (lg4.east) {\ \textbf{Output:} \secref{sec:ear_traits}\ Ear- and \secref{sec:kernel_traits}\ kernel-level traits};
\end{scope}

\end{tikzpicture}
}
  \caption[Full pipeline for 3D maize ear phenotyping and kernel trait extraction.]{Full pipeline for 3D maize ear phenotyping and kernel trait extraction, with the video-to-point-cloud platform stage adapted from \protect\citet{young2026lowcost}, and the subsequent 2D--3D projection, kernel segmentation, and spatial graph trait extraction stages representing the novel contributions of this work.}
  \label{fig:pipeline}
\end{figure}
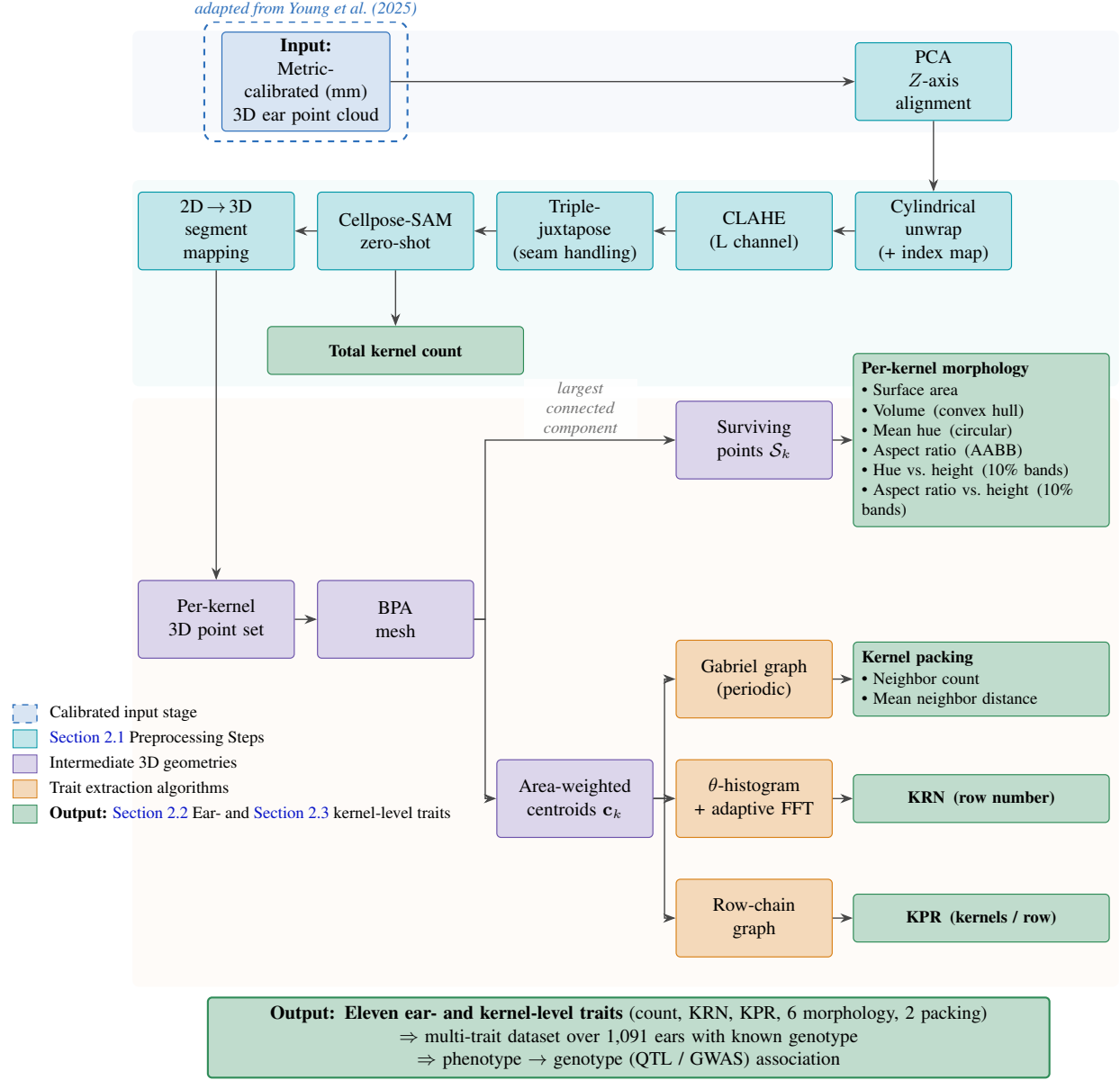

The rest of the paper is structured as follows. \secref{sec:methods} details the video acquisition and 3D point cloud reconstruction pipeline, following \citet{young2026lowcost}, and describes the zero-shot segmentation, per-kernel trait extraction, and validation protocols developed in this work. \secref{sec:results} presents results on the 27-ear synthetic dataset, the 168-ear held-out set, and the six-ear manually annotated subset. Finally, \secref{sec:discussion} interprets these results in light of the pipeline's structural occlusion limitations, examines the synthetic dataset's row-detection discrepancy, and discusses the implications of this work for breeding-scale phenotyping.

% ============================================================
\section{Materials and Methods}
\label{sec:methods} 
%In the Materials and Methods section, give enough detail to allow a competent scientist to repeat the experiments, mentally or in fact.

The overall workflow, from raw 360\textdegree{} video to eleven ear- and kernel-level phenotypic traits, is summarized in \figref{fig:pipeline}. The 3D reconstruction procedure converting raw 360\textdegree{} video into a point cloud representation was adapted from the approach described by \citet{young2026lowcost} and is detailed in the \hyperref[sec:video_to_pcd]{Supplemental Material, ``Corn Ear 3D Reconstruction Using Stationary Camera''}. The pipeline components developed in this work are described in the four sections that follow. \secref{sec:preprocessing} describes the 3D point cloud alignment, cylindrical unwrapping, contrast enhancement, and zero-shot kernel instance segmentation steps. \secref{sec:ear_traits} outlines the extraction of ear-level phenotypic traits, and \secref{sec:kernel_traits} details the extraction of individual kernel geometric, volumetric, and packing traits. \secref{sec:metrics} then defines the statistical metrics and computational cost measurement protocol used to evaluate pipeline performance.

\subsubsection*{Dataset Composition}
\label{sec:dataset_composition}

Three distinct ear populations were used in this study. The pipeline was applied at scale to a population of 1{,}091 reconstructed ears of known genotype identity (\secref{sec:results}). Of these, 268 ears had manually annotated kernel count (KC) and kernel row number (KRN) ground truth available and constitute the labeled dataset; the others were not annotated manually, reflecting the labor-intensive nature of manual phenotyping that motivates this work. The 268-ear labeled dataset was further split into a 100-ear tuning set, used to optimize all pipeline parameters (\secref{sec:paramtuning}), and a 168-ear held-out set, reserved for final validation (\secref{sec:168_results}). Separately, a six-ear manually annotated subset was used for exhaustive kernel-level validation of spatial packing and row-tracking accuracy (\secref{sec:manual_validation}) and to calibrate default parameters for the synthetic dataset (\secref{sec:synthetic_methods}). These six ears were drawn from an independent field study and have no overlap with either the 1{,}091-ear reconstructed population or the 268-ear labeled dataset. The two datasets differ in annotator count and annotation depth; full detail is provided in the \hyperref[sec:supplemental]{Supplemental Material, ``Manual Annotation Protocol''} (\figref{fig:annotation_protocol}). Finally, a 27-ear synthetic dataset with exactly known generating geometry was procedurally constructed for per-trait validation independent of manual annotation (\secref{sec:synthetic_methods}). \tableref{tab:dataset_composition} summarizes these populations and their relationships.

\begin{table}[t!]
\centering
\small
\setlength{\tabcolsep}{3pt}
\caption{Summary of the ear populations used in this study and their relationships.}
\label{tab:dataset_composition}
\renewcommand{\arraystretch}{0.95}
\begin{tabularx}{\textwidth}{@{} >{\raggedright\arraybackslash}p{2.7cm} >{\raggedright\arraybackslash}p{1.6cm} >{\raggedright\arraybackslash}p{4.2cm} >{\raggedright\arraybackslash}X @{}}
\toprule
\textbf{Population} & \textbf{Size} & \textbf{Role} & \textbf{Relationship} \\
\midrule
Reconstructed population & 1{,}091 ears & Full breeding-scale dataset; applied pipeline output & Superset; contains the 268-ear labeled dataset \\[2pt]
Labeled dataset & 268 ears & Ears with manual KC/KRN ground truth & Subset of the 1{,}091; split into tuning + held-out \\[2pt]
\quad Tuning set & 100 ears & Parameter optimization (\secref{sec:paramtuning}) & Subset of the 268-ear labeled dataset \\[2pt]
\quad Held-out set & 168 ears & Final validation (\secref{sec:168_results}) & Subset of the 268-ear labeled dataset; disjoint from tuning set \\[2pt]
Manually annotated subset & 6 ears & Exhaustive kernel-level validation (\secref{sec:manual_validation}); synthetic parameter calibration & Independent field study; no overlap with the 1{,}091 or 268-ear sets \\[2pt]
Synthetic dataset & 27 ears & Per-trait validation against exact generating geometry (\secref{sec:synthetic_methods}) & Procedurally generated; default parameters calibrated from aggregate statistics of the six-ear subset, but no individual synthetic ear is derived from a real scan \\
\bottomrule
\end{tabularx}
\end{table}

% ============================================================

\subsection{Preprocessing Steps}
\label{sec:preprocessing}

\subsubsection*{Z-Axis Alignment via PCA}
\label{sec:pca}

\begin{figure}[b!]
\centering
\begin{subfigure}[b]{0.22\textwidth}
    \includegraphics[width=\textwidth]{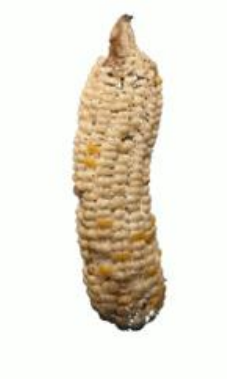}
    \caption{24-P2042\_3}
\end{subfigure}\hfill
\begin{subfigure}[b]{0.22\textwidth}
    \includegraphics[width=\textwidth]{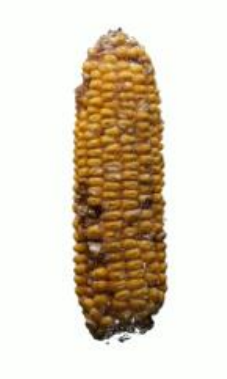}
    \caption{24-P2071\_2}
\end{subfigure}\hfill
\begin{subfigure}[b]{0.22\textwidth}
    \includegraphics[width=\textwidth]{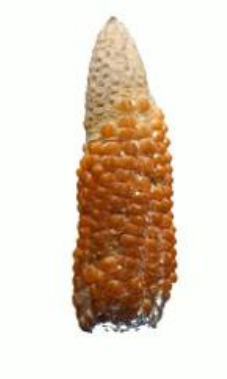}
    \caption{24-P2146\_2}
\end{subfigure}\hfill
\begin{subfigure}[b]{0.22\textwidth}
    \includegraphics[width=\textwidth]{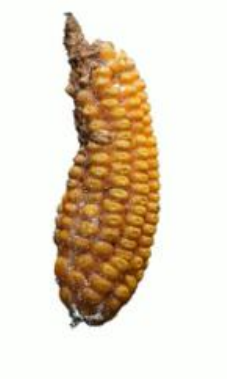}
    \caption{24-P2199\_1}
\end{subfigure}
\caption{Isolated and calibrated 3D point clouds for the four representative ears. Density-based separation and distance calibration to physical units follow \protect\citet{young2026lowcost}; Z-axis alignment via PCA is contributed by this work.}
\label{fig:pointcloud}
\end{figure}

To standardize the orientation of the ear for subsequent processing, the ear point cloud was aligned so that its principal axis coincided with the Z-axis. Principal Component Analysis (PCA) was applied to the point cloud; the first principal component, corresponding to the long axis of the ear, was identified. A rotation matrix was computed to align this principal axis with $[0, 0, 1]^T$ using the axis-angle rotation formula, and all points were transformed accordingly. \figref{fig:pointcloud} shows the resulting isolated, calibrated, and Z-axis-aligned point clouds for the four representative ears.

\subsubsection*{Cylindrical Unwrapping}
\label{sec:unwrap}

Per-kernel traits in 3D require segmenting individual kernels first. Because segmentation was performed on 2D imagery, the point cloud was first unwrapped to a 2D image, the kernels were segmented in that image, and the resulting 2D segments were re-projected to 3D point sets. The aligned 3D point cloud was mapped onto a 2D image via cylindrical unwrapping. Each point $p_i = (x_i, y_i, z_i)$ was first projected onto the XY plane, and the geometric centroid $(c_x, c_y)$ was computed from the moments of the XY convex hull of $\{(x_i, y_i)\}$. Each point was then mean-centered:
\begin{equation}
    x_i' = x_i - c_x, \qquad y_i' = y_i - c_y.
\end{equation}
The per-point radius and effective cylinder radius were computed as:
\begin{equation}
    r_i = \sqrt{x_i'^2 + y_i'^2}, \qquad R = P_{98}(\{r_i\}),
\end{equation}
where $R$ was set empirically to the $98$th percentile of all radii. This choice formed the first step of tuning stage 1 (\secref{sec:paramtuning}): percentile values from 95 to 99 were swept on the 100-ear tuning set, and 98 gave the closest agreement between predicted and manually counted total kernel counts. Next, the azimuthal angle was computed as:
\begin{equation}
    \theta_i = \mathrm{atan2}(y_i',\, x_i'),
\label{eq:theta}
\end{equation}
then mapped to $[0, 2\pi)$ by adding $2\pi$ where $\theta_i < 0$. Each point was then mapped to a pixel at column $u_i$ and row $v_i$:
\begin{equation}
    u_i = \left\lfloor \frac{\theta_i}{2\pi} (W-1) \right\rfloor,
    \qquad
    v_i = \left\lfloor \frac{z_i - z_{\min}}{H_{\mathrm{mm}}} (H-1)
    \right\rfloor,
\end{equation}
where $W$ and $H$ are the image width and height in pixels and $H_{\mathrm{mm}} = z_{\max} - z_{\min}$. Both image dimensions were set proportionally to the physical circumference $2\pi R$ and the ear height $H_{\mathrm{mm}}$ at a fixed resolution of $0.25\ \mathrm{mm\ pixel}^{-1}$. That resolution was selected in the second step of tuning stage 1 (\secref{sec:paramtuning}), sweeping values from 0.1 to 1.0 in steps of 0.05 on the 100-ear tuning set to maximize kernel boundary separability and segmentation accuracy against manually measured total kernel count. Where multiple points projected to the same pixel, the point with the minimum RGB value was retained, enhancing kernel boundary contrast and improving subsequent segmentation accuracy. The resulting 2D image represents the full $360^\circ$ surface of the ear laid flat, with the circumference axis horizontal and the ear height axis vertical. Each pixel retains an index mapping back to its source 3D point, so that 2D segments can later be re-projected to 3D point sets.

\subsubsection*{Contrast Enhancement via Contrast Limited Adaptive Histogram Equalization (CLAHE)}

\begin{figure}[t!]
\centering
\begin{subfigure}[b]{0.48\textwidth}
    \centering
    \includegraphics[width=\textwidth]{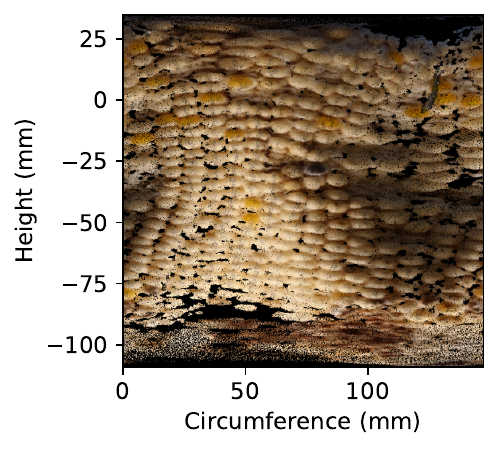}
    \caption{Raw unwrapped image}
\end{subfigure}%
\hfill
\begin{subfigure}[b]{0.48\textwidth}
    \centering
    \includegraphics[width=\textwidth]{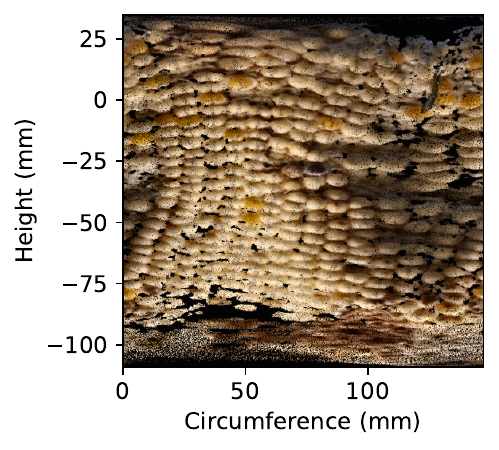}
    \caption{After CLAHE}
\end{subfigure}
\caption{Cylindrically unwrapped image of ear 24-P2042\_3 before (a) and after (b) CLAHE contrast enhancement. The full 360\textdegree{} surface is laid flat with circumference horizontal and ear height vertical.}
\label{fig:clahe}
\end{figure}

To normalize local contrast and improve kernel boundary visibility prior to segmentation, the unwrapped image was converted from RGB to LAB color space. Contrast Limited Adaptive Histogram Equalization (CLAHE)~\citep{zuiderveld1994clahe} was applied exclusively to the L channel. The image was then converted back to RGB, yielding a contrast-enhanced unwrapped image used as input to segmentation. A clip limit of 1.0 and a tile grid size of $8 \times 8$ pixels were used. Both parameters were tuned sequentially on the 100-ear tuning set as tuning stage 2 (\secref{sec:paramtuning}), minimizing MAE against manual kernel counts, before the Cellpose-SAM (CPSAM) segmentation parameters were tuned. Because CLAHE operates on the luminance channel without reference to segmentation boundaries, optimizing it for downstream count accuracy rather than intrinsic contrast quality means the chosen parameters reflect the needs of the full pipeline rather than local contrast alone. In the first step of stage 2, the clip limit was swept from 0.5 to 2.0 in steps of 0.5, yielding an optimal value of 1.0. In the second step, with the clip limit fixed at 1.0, the tile grid size was swept from $4 \times 4$ to $16 \times 16$ in steps of $2 \times 2$, yielding an optimal size of $8 \times 8$. \figref{fig:clahe} shows the resulting contrast enhancement for a representative ear, with kernel boundaries substantially more separable after CLAHE than in the raw unwrapped image.

\subsubsection*{Triple-Juxtaposed Unwrap Strategy}

\begin{figure}[t!]
\centering
\includegraphics[width=\textwidth]{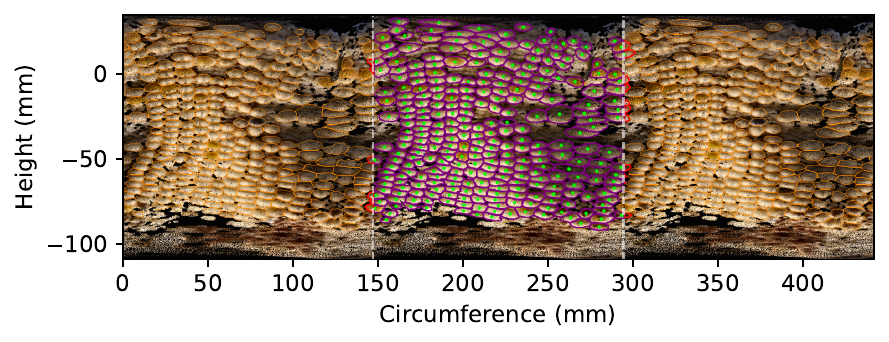}
\caption{Triple-juxtaposed unwrapped image with CPSAM segmentation overlaid (shown for ear 24-P2042\_3). Three copies of the contrast-enhanced image are concatenated horizontally; only segments whose center of mass falls within the middle copy (highlighted in purple) are retained, ensuring kernels straddling the seam boundary are counted exactly once (seam-straddling segments shown in red).}
\label{fig:triple}
\end{figure}

Cylindrical unwrapping introduces a seam at $\theta = 0$ where kernels that straddle the seam boundary are split across the left and right edges of the image, causing them to be double-counted if segmented naively. To resolve this, the contrast-enhanced unwrapped image was tripled horizontally by concatenating three copies side by side prior to segmentation. Instance segmentation was then performed on this tripled image. After segmentation, only segments whose center of mass fell within the middle copy ($x \in [W, 2W)$) were retained, counting each segment, whether fully contained within the middle copy or straddling a seam, exactly once. The total kernel count was defined as the number of retained segment centers. \figref{fig:triple} shows the tripled image with retained and discarded segments overlaid, illustrating how a kernel straddling the seam is counted once from its occurrence in the middle copy.

\subsubsection*{Segmentation via CPSAM}

\begin{figure}[t!]
\centering
\begin{subfigure}[b]{0.45\textwidth}
    \centering
    \includegraphics[width=\textwidth]{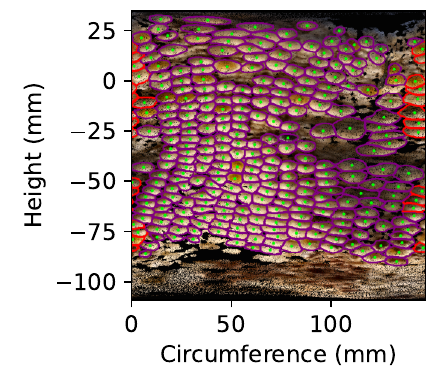}
    \caption{24-P2042\_3}
\end{subfigure}\hfill
\begin{subfigure}[b]{0.45\textwidth}
    \centering
    \includegraphics[width=\textwidth]{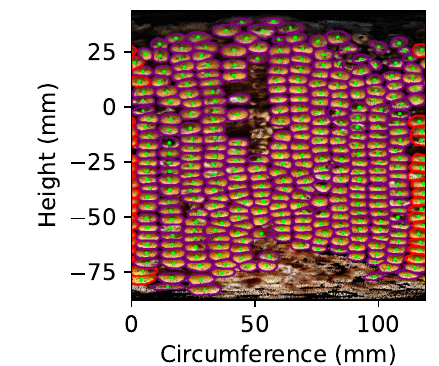}
    \caption{24-P2071\_2}
\end{subfigure}

\vspace{0.5em}

\begin{subfigure}[b]{0.45\textwidth}
    \centering
    \includegraphics[width=\textwidth]{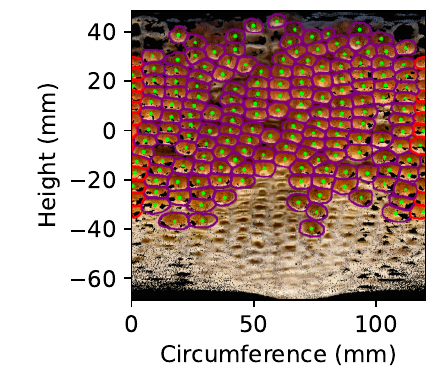}
    \caption{24-P2146\_2}
\end{subfigure}\hfill
\begin{subfigure}[b]{0.45\textwidth}
    \centering
    \includegraphics[width=\textwidth]{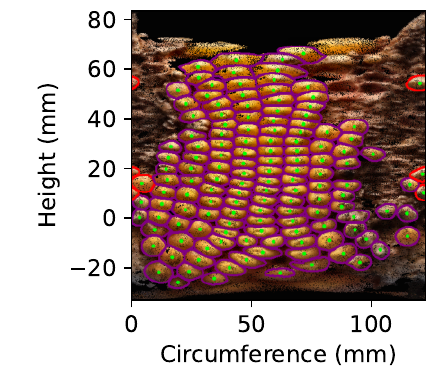}
    \caption{24-P2199\_1}
\end{subfigure}
\caption{CPSAM kernel instance segmentation results for the four representative ears. The total number of kernels is shown with count equal to the total number of distinct segments.}
\label{fig:segmentation}
\end{figure}

Individual kernels were delineated using CPSAM \citep{stringer2025cellposeSAM}, a foundation-model-based instance segmentation method. CPSAM was applied zero-shot to the tripled, contrast-enhanced unwrapped image, with no task-specific fine-tuning on corn kernel imagery, using a cell probability threshold of 2.0, a flow threshold of 0.6, and a minimum segment size of 150 pixels. These three parameters were tuned sequentially, in the order listed, on the 100-ear tuning set as tuning stage 3 (\secref{sec:paramtuning}), minimizing MAE against manual kernel counts with the upstream CLAHE configuration held fixed. In the first step of stage 3, the cell probability threshold was swept over \{1.0, 1.5, 2.0, 2.5, 3.0\}, yielding an optimal value of 2.0. In the second step, with the cell probability threshold fixed at 2.0, the flow threshold was swept over \{0.4, 0.5, 0.6, 0.7, 0.8\}, yielding an optimal value of 0.6. In the third step, with both prior parameters fixed, the minimum segment size was swept over \{50, 100, 150, 200, 250\} pixels, yielding an optimal value of 150 pixels. \figref{fig:segmentation} shows the resulting kernel instance segmentation for the four representative ears, with the retained kernel count reported for each.

\subsubsection*{Mapping 2D Segments to 3D Point Sets}

Each 2D kernel segment was used to collect the corresponding 3D points from the aligned ear point cloud. For every pixel within a segment that had a valid entry in the index map (i.e., a 3D point was projected to that pixel during unwrapping), the corresponding 3D point index was recorded. The unique set of 3D point indices associated with each segment thus defined a per-kernel 3D point set, forming the basis for all subsequent volumetric and geometric trait computations.

\subsection{Ear-Level Traits}
\label{sec:ear_traits}

\subsubsection*{Kernel Count, Kernel Row Number, and Kernels Per Row}

\begin{figure}[p]
\centering
\begin{subfigure}[b]{\textwidth}
    \centering
    \includegraphics[height=0.18\textheight, keepaspectratio=true]{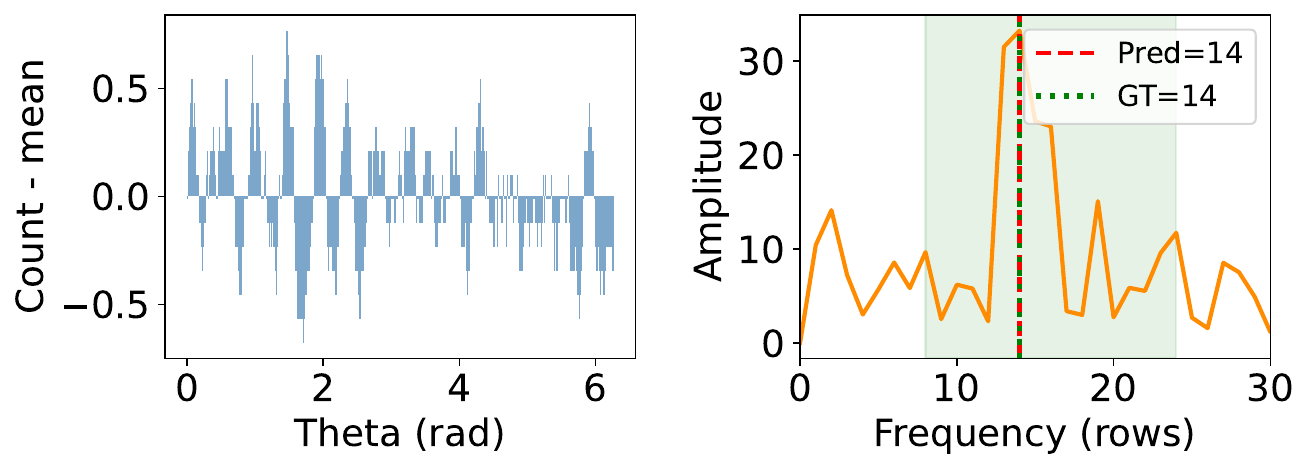}
    \caption{24-P2042\_3}
\end{subfigure}
\vspace{0.1em}
\begin{subfigure}[b]{\textwidth}
    \centering
    \includegraphics[height=0.18\textheight, keepaspectratio=true]{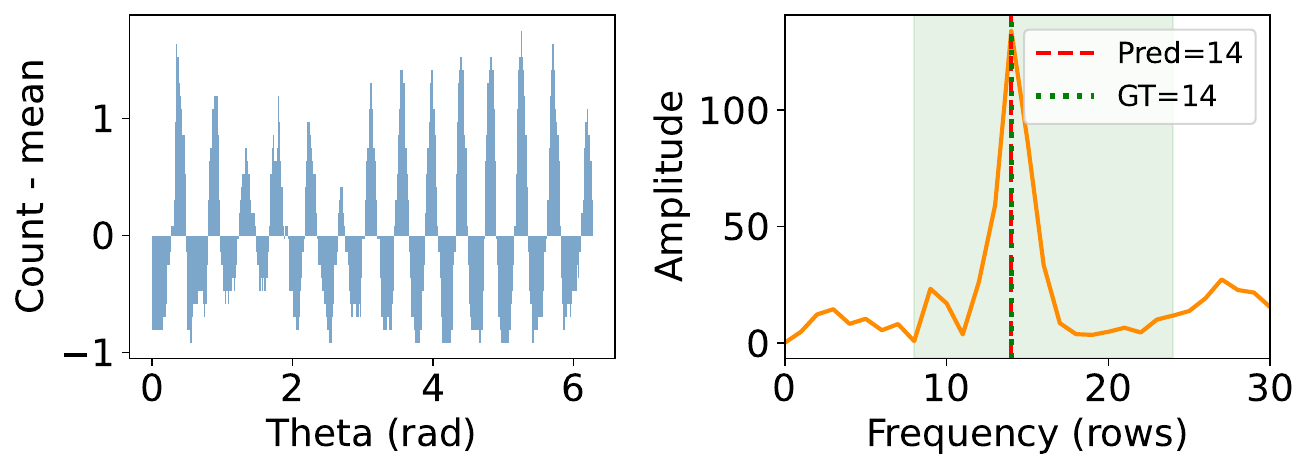}
    \caption{24-P2071\_2}
\end{subfigure}
\vspace{0.1em}
\begin{subfigure}[b]{\textwidth}
    \centering
    \includegraphics[height=0.18\textheight, keepaspectratio=true]{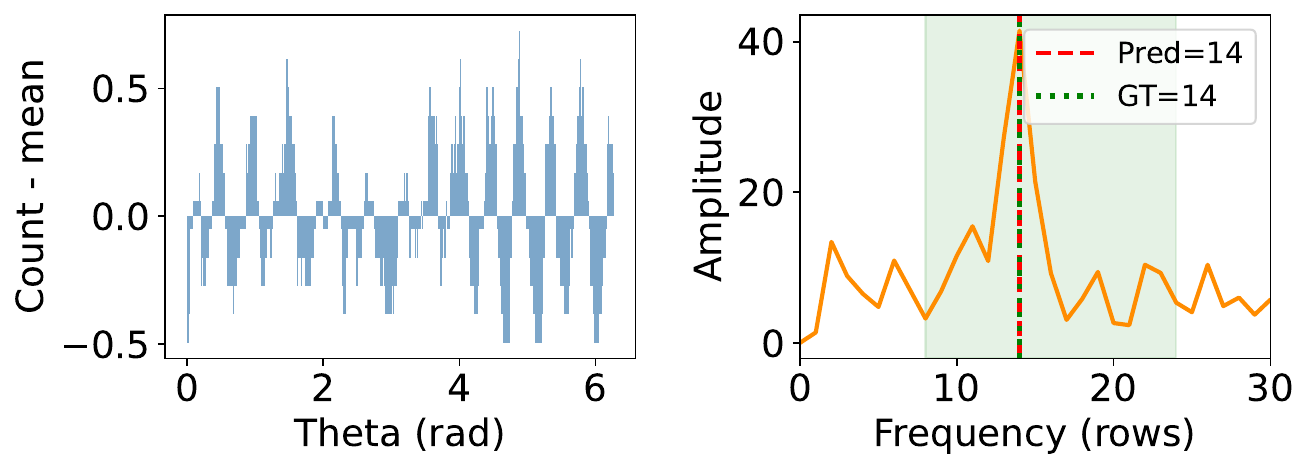}
    \caption{24-P2146\_2}
\end{subfigure}
\vspace{0.1em}
\begin{subfigure}[b]{\textwidth}
    \centering
    \includegraphics[height=0.18\textheight, keepaspectratio=true]{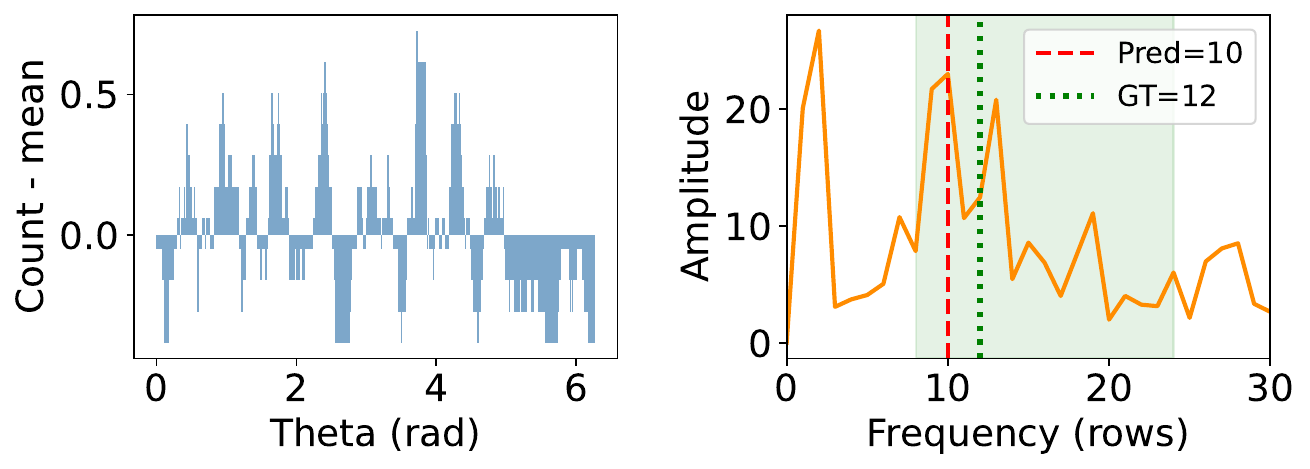}
    \caption{24-P2199\_1}
    \label{fig:fft_d}
\end{subfigure}
\vspace{0.5em}
\caption{$\theta$ histograms and FFT spectra for each representative ear. In panel (d), two competing peaks fall within the search range and the FFT selects the lower-frequency peak, underpredicting KRN as 10 against a ground truth of 12.}
\label{fig:fft}
\end{figure}

\begin{figure}[t!]
\centering
\begin{subfigure}[b]{0.22\textwidth}
    \includegraphics[width=\textwidth]{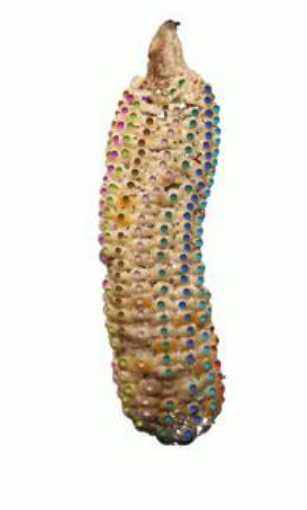}
    \caption{24-P2042\_3}
\end{subfigure}\hfill
\begin{subfigure}[b]{0.22\textwidth}
    \includegraphics[width=\textwidth]{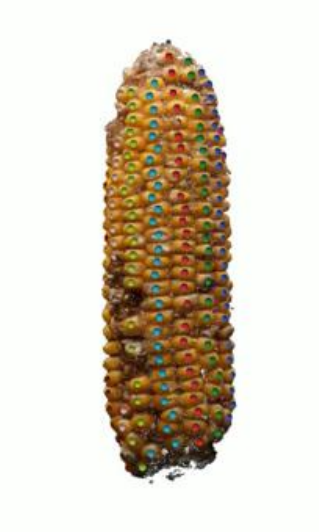}
    \caption{24-P2071\_2}
\end{subfigure}\hfill
\begin{subfigure}[b]{0.22\textwidth}
    \includegraphics[width=\textwidth]{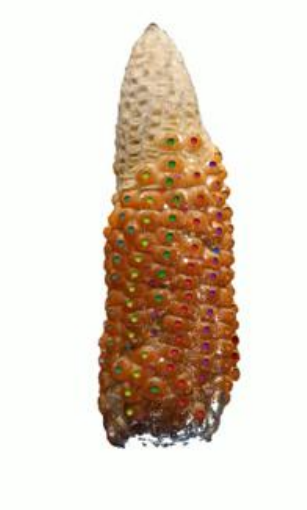}
    \caption{24-P2146\_2}
\end{subfigure}\hfill
\begin{subfigure}[b]{0.22\textwidth}
    \includegraphics[width=\textwidth]{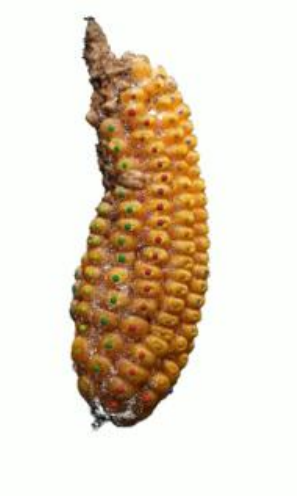}
    \caption{24-P2199\_1}
\end{subfigure}
\caption{Row chain assignments visualized on the 3D ear surface for each representative ear. Each color represents a distinct row chain produced by the angular spanning forest algorithm.}
\label{fig:rowchains}
\end{figure}

Beyond the per-kernel traits described in \secref{sec:kernel_traits}, three traits characterize the ear at the whole-ear level: total kernel count (KC), kernel row number (KRN), and kernels per row (KPR). KC was obtained directly as the total number of retained kernel segments following the triple-juxtaposed segmentation strategy described in \secref{sec:preprocessing}. KRN and KPR were estimated via an adaptive FFT (\figref{fig:fft}) and a row chain graph (\figref{fig:rowchains}), respectively, as described in the two subsections that follow.

\subsubsection*{Row Count Estimation via Adaptive FFT}

Kernels arranged in the same longitudinal row on the ear share similar azimuthal angles $\theta$ in the unwrapped space. Consequently, the number of kernel rows can be estimated by the number of evenly distributed peaks in the histogram of $\theta$ values across all kernel centroids, which in turn equals the dominant frequency in the Fast Fourier Transform (FFT) of that histogram. The azimuthal angles of all kernel centroids were therefore binned into a histogram of $B = 360$ bins over $[0, 2\pi)$. The coefficient of variation of the histogram counts $\{h_b\}_{b=1}^{B}$ was computed as:
\begin{equation}
    \mathrm{CV} = \frac{\sigma_h}{\mu_h},
\end{equation}
where $\mu_h$ and $\sigma_h$ are the mean and standard deviation of the bin counts respectively. The smoothing kernel size and CV threshold were jointly optimized on the 100-ear tuning set as tuning stage 4, independently of the three sequential kernel count tuning stages (\secref{sec:paramtuning}). The sweep covered odd kernel sizes from 3 to 15 bins and 200 evenly spaced CV threshold values spanning the full range of CV values observed on that set. The combination minimizing MAE against manual row counts, with RMSE as a tie-breaker, gave a kernel size of $15$ bins and a threshold of $1.8725$.

If $\mathrm{CV} \geq 1.8725$, indicating a highly uneven angular distribution, the histogram was smoothed with a uniform moving-average kernel of width 15 bins prior to FFT to suppress noise; otherwise, no smoothing was applied. The mean was subtracted from the histogram to remove the DC component:
\begin{equation}
    \tilde{h}_b = h_b - \mu_h,
\end{equation}
and the real-valued FFT was computed as $\mathcal{F} = |\mathrm{RFFT}(\{\tilde{h}_b\})|$. The dominant frequency within the expected row range $f \in [8,24]$ was identified as
\begin{equation}
    \hat{R}=\underset{f\in[8,\,24]}{\arg\max}\;\mathcal{F}(f),
\end{equation}
where $\hat{R}$ is the initial predicted kernel row number (KRN). For a histogram containing $N=360$ angular bins, the FFT frequency axis is defined as
\begin{equation}
    f_k=\frac{k}{Nd},
    \qquad
    k=0,1,\ldots,\frac{N}{2},
\end{equation}
where $d$ is the sample spacing. Since the histogram was uniformly sampled over one revolution, $d=1/N$, yielding
\begin{equation}
    f_k=\frac{k}{N(1/N)}=k.
\end{equation}
Therefore, the FFT frequencies correspond exactly to the integer harmonics $0,1,\ldots,N/2$, making the dominant FFT frequency, and hence $\hat{R}$, inherently an integer. The search range $f\in[8,24]$ corresponds to the biologically observed KRN range in cultivated maize. To account for the predominance of even kernel row numbers in maize, the final KRN prediction was computed as
\begin{equation}
\hat{R}_{\mathrm{final}}=
\begin{cases}
\hat{R}, & \text{if } \hat{R}\text{ is even},\\
\hat{R}+1, & \text{if } \hat{R}\text{ is odd},
\end{cases}
\label{eq:even_round}
\end{equation}
where $\hat{R}_{\mathrm{final}}$ denotes the final predicted KRN used in subsequent analysis.
This biologically motivated post-processing rule was selected from evaluation on the 100-ear tuning set (\tableref{tab:even_ablation}), where rounding odd-valued predictions up to the next even integer lowered MAE relative to the unrounded estimate. Because the rule only ever increases a prediction, it shifts the signed mean error (Signed ME) in the positive direction, and Signed ME is reported alongside MAE to quantify that shift. The raw estimate is nearly unbiased (Signed ME $= -0.08$ rows), while the even-rounded estimate carries a small positive bias (Signed ME $= +0.26$ rows). This bias is the rule working as intended: KRN in cultivated maize is predominantly even, so a positive shift toward the nearest even integer moves odd predictions closer to the ground-truth distribution, and MAE improves under the induced bias.

\begin{table}[t!]
\centering
\small
\caption{Effect of the even-parity rounding rule (Eq.~\ref{eq:even_round}) on KRN prediction accuracy for the 100-ear tuning set.}
\label{tab:even_ablation}
\begin{tabular}{lcc}
\toprule
\textbf{KRN Estimate} & \textbf{MAE (rows)} & \textbf{Signed ME (rows)} \\
\midrule
Raw FFT peak, $\hat{R}$ (no rounding) & 0.88 & $-0.08$ \\
Even-rounded, $\hat{R}_{\mathrm{final}}$ (used) & 0.86 & $+0.26$ \\
\bottomrule
\end{tabular}
\end{table}

\subsubsection*{Row Chain Graph and Per-Row Kernel Counts}

To assign individual kernels to rows and compute kernels per row, a row chain graph was constructed on the 3D kernel centroids. The azimuthal angle $\theta_i$ of each kernel centroid was computed as in Equation~\ref{eq:theta}, and the circular angular distance between two kernels was defined as $\delta(\theta_i, \theta_j) = \min(|\theta_i - \theta_j|,\, 2\pi - |\theta_i - \theta_j|)$. An angular spanning forest was built by connecting each kernel to its nearest angular neighbors within a per-ear optimized angular threshold $\alpha$, with the degree of each node constrained to at most two throughout, forming chains of kernels that trace individual rows from base to tip.

After the initial graph was constructed, vertically discontinuous chains were reconnected. A candidate link joined the top node $u$ of one chain to the bottom node $v$ of another, subject to $z_u < z_v$, $\delta(\theta_u, \theta_v) \leq \alpha$, and both $u$ and $v$ having degree one. The candidate pair minimizing $\delta(\theta_u, \theta_v)$ was linked, and this step was repeated until no further connections were possible, with the degree-two constraint enforced after each addition. The threshold $\alpha$ was determined per ear at inference time by sweeping from $5^\circ$ to $30^\circ$ in $1^\circ$ increments. Writing $C(\alpha)$ for the number of connected components produced by the spanning forest at threshold $\alpha$, the selected $\alpha$ was the one minimizing $C(\alpha) - \hat{R}$ among all angles producing $C(\alpha) \geq \hat{R}$.

If the resulting number of chains still exceeded $\hat{R}$, the smallest chain by kernel count was repeatedly merged into the chain with the nearest mean angular position $\bar{\theta} = \mathrm{atan2}\!\left(\sum_{i}\sin\theta_i,\, \sum_{i}\cos\theta_i\right)$ under $\delta$, until exactly $\hat{R}$ chains remained. The size of each final chain, that is, the number of kernels in it, was recorded as the kernels-per-row measurement for that row. Because the number of retained chains is fixed at $\hat{R}$, KPR accuracy is conditional on correct KRN estimation: an underestimated $\hat{R}$ forces kernels from distinct true rows into a single predicted chain.

\FloatBarrier

\subsection{Kernel-Level Traits}
\label{sec:kernel_traits}

\subsubsection*{Kernel Surface Area}

\begin{figure}[t!]
\centering
\begin{subfigure}[b]{0.22\textwidth}
    \includegraphics[width=\textwidth]{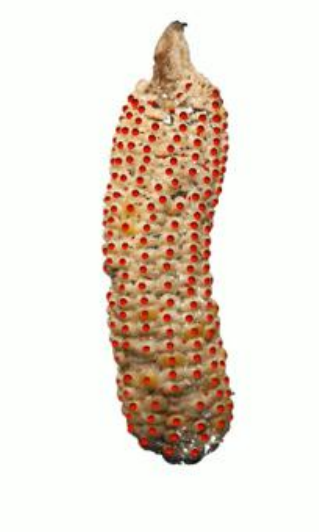}
    \caption{24-P2042\_3}
\end{subfigure}\hfill
\begin{subfigure}[b]{0.22\textwidth}
    \includegraphics[width=\textwidth]{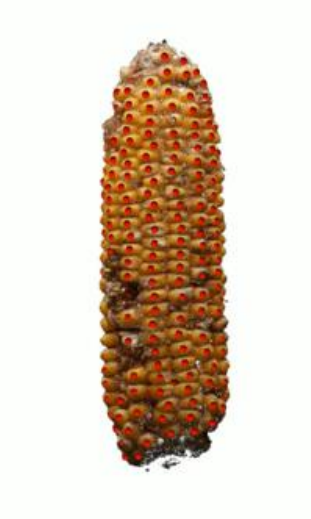}
    \caption{24-P2071\_2}
\end{subfigure}\hfill
\begin{subfigure}[b]{0.22\textwidth}
    \includegraphics[width=\textwidth]{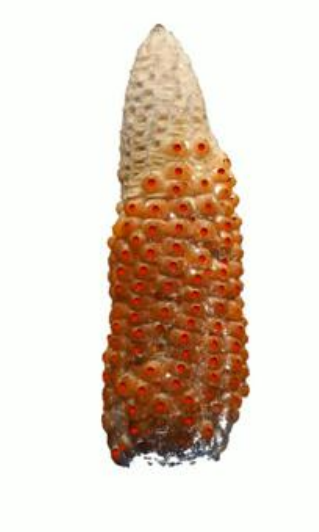}
    \caption{24-P2146\_2}
\end{subfigure}\hfill
\begin{subfigure}[b]{0.22\textwidth}
    \includegraphics[width=\textwidth]{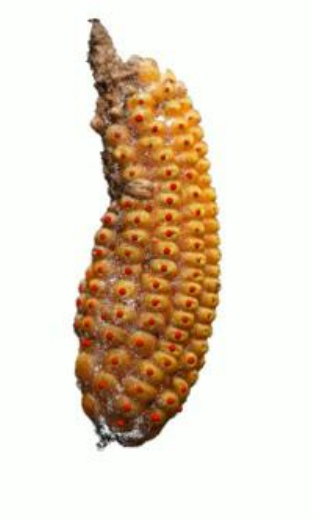}
    \caption{24-P2199\_1}
\end{subfigure}
\caption{Per-kernel 3D centroids (red spheres) overlaid on each reconstructed ear point cloud.}
\label{fig:centers}
\end{figure}

Per-kernel surface area was estimated from each kernel's 3D point set using the Ball-Pivoting Algorithm (BPA)~\citep{bernardini1999ball}.  For each kernel, surface normals were estimated using a K Nearest Neighbor (KNN) search with $k = \min(30,\, n_k - 1)$, where $n_k$ is the number of points in kernel $k$, and oriented consistently using tangent plane propagation. Letting $d_i$ denote the distance from each point to its nearest neighbor, the two BPA ball radii were set to:
\begin{equation}
    r_1 = P_{98}(\{d_i\}), \qquad r_2 = 2\,r_1,
\end{equation}
where $P_{98}(\{d_i\})$ denotes the 98th percentile of the nearest-neighbor distance distribution for each kernel point set, chosen so that the pivoting ball is large enough to bridge typical inter-point gaps while remaining small enough to resolve individual kernel surface geometry. Denoting the vertices of triangle $t$ as $\mathbf{v}_0^t,\, \mathbf{v}_1^t,\, \mathbf{v}_2^t$, the area of each triangle was computed as:
\begin{equation}
    a_t = \frac{1}{2}
    \left\|(\mathbf{v}_1^t - \mathbf{v}_0^t)
    \times
    (\mathbf{v}_2^t - \mathbf{v}_0^t)\right\|,
\end{equation}
and the total surface area of kernel $k$ as:
\begin{equation}
    A_k = \sum_{t} a_t.
\end{equation}
The area-weighted 3D centroid of kernel $k$ was then computed as:
\begin{equation}
    \mathbf{c}_k = \frac{\sum_{t} a_t
    \left(\dfrac{\mathbf{v}_0^t + \mathbf{v}_1^t +
    \mathbf{v}_2^t}{3}\right)}{\sum_{t} a_t},
\end{equation}
and used as the 3D centroid of each kernel for downstream computations. \figref{fig:centers} shows the resulting per-kernel centroids overlaid on each representative ear's point cloud. The vertices of the largest connected component of the BPA mesh for kernel $k$ are denoted $\mathcal{S}_k$; the following subsections use $\mathcal{S}_k$ and the per-kernel centroids $\mathbf{c}_k$ to extract the pipeline's remaining kernel-level traits, describing kernel size, color, shape, and packing.

\subsubsection*{Kernel Convex Hull Volume Proxy}

We define the {\bf convex hull volume proxy} of kernel $k$ as the convex hull of the 3D point set $\mathcal{S}_k$, computed using the Quickhull algorithm as implemented in \texttt{scipy.spatial.ConvexHull}~\citep{virtanen2020scipy}.  Because the kernel interior and the proximal cob-facing surface are occluded from all camera views, the convex hull closes across the occluded region, enclosing only the exposed outer cap geometry. The resulting convex hull volume proxy is therefore a relative size descriptor of the visible kernel geometry rather than an estimate of the kernel's true physical volume.

\subsubsection*{Mean Kernel Hue}
\label{sec:mean_hue}

The circular mean hue of each kernel was computed from the RGB values of $\mathcal{S}_k$. Each point's RGB value was converted to HSV and its hue angle $h_i \in [0^\circ, 360^\circ)$ recorded. The circular mean was then computed as:
\begin{equation}
    \bar{h}_k = \mathrm{atan2}\!\left(
        \tfrac{1}{|\mathcal{S}_k|}\textstyle\sum_{i} \sin h_i,\;
        \tfrac{1}{|\mathcal{S}_k|}\textstyle\sum_{i} \cos h_i
    \right) \bmod 360^\circ,
\end{equation}
to correctly handle the wraparound at $0^\circ/360^\circ$.

\subsubsection*{3D Bounding Box Aspect Ratio}
\label{sec:aspect_ratio}

The axis-aligned bounding box (AABB) of each kernel was computed from $\mathcal{S}_k$. The three edge lengths
\begin{equation}
    \ell_1 = x_{\max} - x_{\min}, \qquad
    \ell_2 = y_{\max} - y_{\min}, \qquad
    \ell_3 = z_{\max} - z_{\min}
\end{equation}
were sorted in descending order as $\ell_{(1)} \geq \ell_{(2)} \geq \ell_{(3)}$, and the aspect ratio defined as:
\begin{equation}
    \mathrm{AR}_k = \frac{\ell_{(1)}}{\ell_{(2)}} \geq 1.
\end{equation}
Because each kernel's 3D segment bounding box captures only the exposed outer cap, its dimensions reflect visible surface depth. The resulting aspect ratio ($\mathrm{AR}_k$) functions as a consistent relative shape descriptor optimized for non-destructive, line-of-sight phenotyping.

\subsubsection*{Mean Hue and Mean Aspect Ratio vs.\ Height}

To characterize how kernel color and shape vary along the ear from base to tip, kernel centroids were partitioned into ten equal height bands of 10\% each along the PCA-aligned Z-axis (base to tip). For each band $b$, the set of kernels $\mathcal{K}_b$ whose Z coordinate fell within the corresponding interval of the total ear height was identified. The mean hue vs.\ height profile was computed as the circular mean of the per-kernel circular mean hues $\bar{h}_k$ (\secref{sec:mean_hue}) over all $k \in \mathcal{K}_b$, giving each kernel equal weight:
\begin{equation}
    \bar{h}_b = \mathrm{atan2}\!\left(
        \tfrac{1}{|\mathcal{K}_b|}\textstyle\sum_{k \in \mathcal{K}_b} \sin \bar{h}_k,\;
        \tfrac{1}{|\mathcal{K}_b|}\textstyle\sum_{k \in \mathcal{K}_b} \cos \bar{h}_k
    \right) \bmod 360^\circ.
\end{equation}
The mean aspect ratio vs.\ height profile was computed as the arithmetic mean of $\mathrm{AR}_k$ (\secref{sec:aspect_ratio}) over all $k \in \mathcal{K}_b$.

\subsubsection*{Gabriel Graph Construction}

To characterize the spatial packing structure of kernels on the ear, the area-weighted 3D centroids $\mathbf{c}_k$ computed during surface area estimation were first projected back onto the 2D unwrapped image coordinate system. The azimuthal angle $\theta_k = \mathrm{atan2}(y_k - c_y,\, x_k - c_x)$ of each centroid was mapped to a column coordinate and its $z$ coordinate to a row coordinate using the same cylindrical mapping applied during unwrapping, yielding a 2D center $\mathbf{u}_k = (u_k, v_k)$ for each kernel. A Gabriel graph \citep{gabriel1969new} was then constructed on these projected 2D centers. In a Gabriel graph, two nodes $p$ and $q$ are connected by an edge if and only if no other node lies within the circle whose diameter is the segment $pq$.

\begin{figure}[t!]
\centering
\begin{subfigure}[b]{0.4\textwidth}
    \centering
    \includegraphics[width=\textwidth]{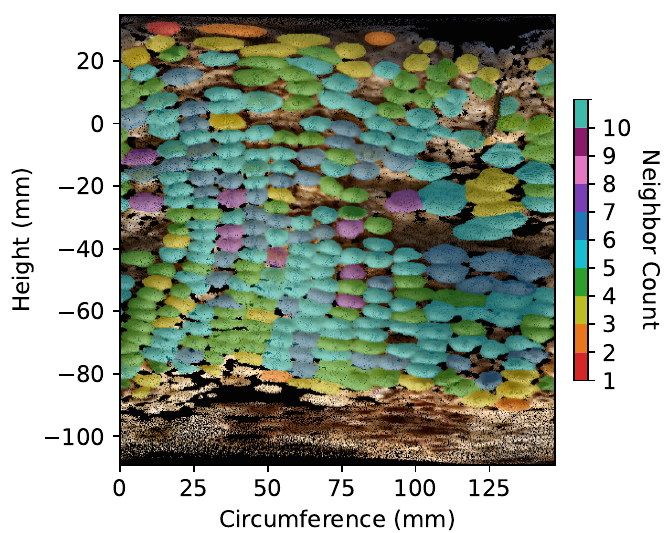}
    \caption{24-P2042\_3}
\end{subfigure}
\begin{subfigure}[b]{0.4\textwidth}
    \centering
    \includegraphics[width=\textwidth]{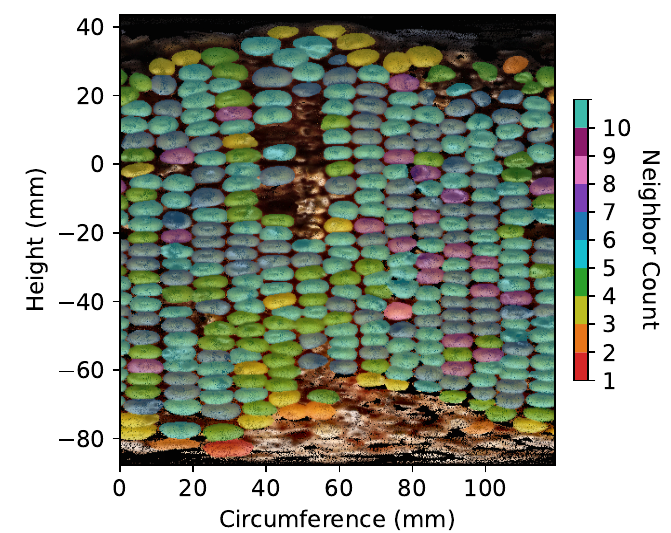}
    \caption{24-P2071\_2}
\end{subfigure}
\begin{subfigure}[b]{0.4\textwidth}
    \centering
    \includegraphics[width=\textwidth]{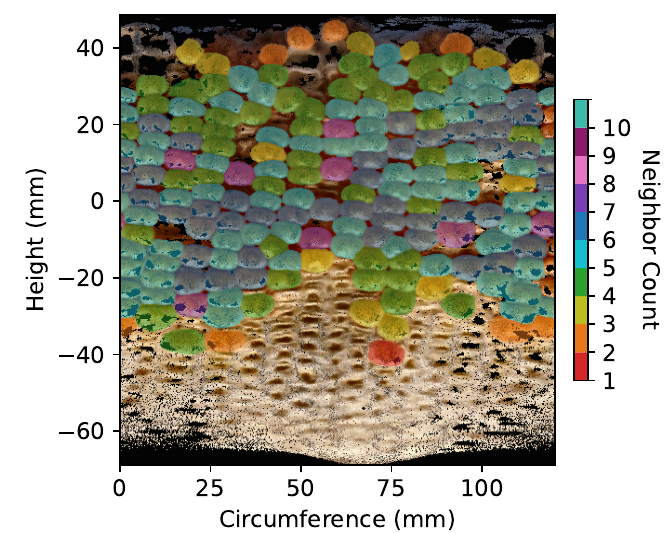}
    \caption{24-P2146\_2}
\end{subfigure}
\begin{subfigure}[b]{0.4\textwidth}
    \centering
    \includegraphics[width=\textwidth]{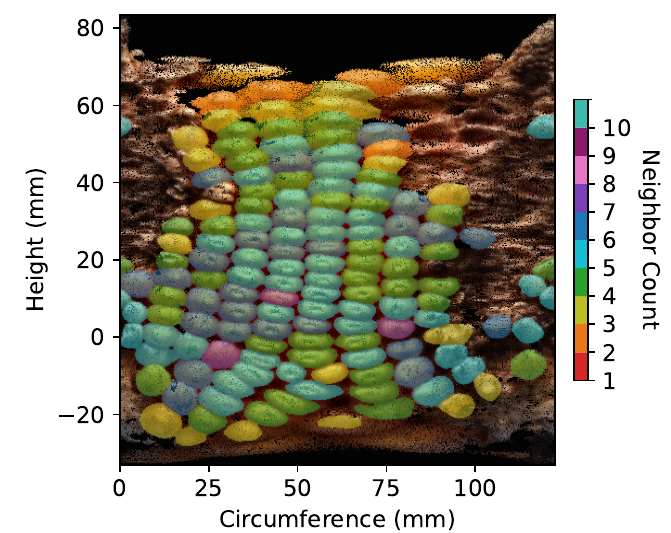}
    \caption{24-P2199\_1}
\end{subfigure}
\caption{Kernel neighbor count maps derived from the Gabriel graph, overlaid on the unwrapped image for each representative ear. Each segmented kernel is colored by its number of Gabriel graph neighbors, providing a spatial view of local packing density across the ear surface.}
\label{fig:neighbors}
\end{figure}

Because the unwrapped image is periodic in the horizontal (circumferential) direction, the graph was computed on a cylindrically-periodic domain: three copies of the kernel centers (offset by $-W$, $0$, and $+W$ pixels) were used to construct a Delaunay triangulation, from which candidate Gabriel edges were extracted and tested for the empty-circle condition across the periodic domain. Neighbor relationships were counted only for middle-copy nodes, ensuring each edge was recorded exactly once. Finally, the number of Gabriel graph neighbors was recorded for each kernel as a measure of local packing density. Because this construction operates directly on kernel centroids and does not consult $\hat{R}$, the Gabriel graph packing traits are independent of the row detection step. \figref{fig:neighbors} maps the resulting per-kernel neighbor counts onto the unwrapped image for each representative ear.

\subsubsection*{Mean Nearest-Neighbor Distance}

For each kernel, the mean Euclidean distance in 3D between its centroid $\mathbf{c}_k$ and the centroids of all its Gabriel graph neighbors was computed. The Gabriel graph was stored as a sparse adjacency structure, from which the direct-neighbor edge weights in 3D were retrieved and averaged per kernel. This quantity characterizes the average physical spacing between a kernel and its proximal neighbors, providing a continuous measure of local kernel packing density.

\subsection{Evaluation Metrics}
\label{sec:metrics}

The evaluation metrics used throughout the evaluations below were computed as follows. For a set of $N$ paired predicted and true values $\{(x_i, y_i)\}_{i=1}^{N}$, the mean absolute error (MAE) was computed as
\begin{equation}
    \mathrm{MAE} = \frac{1}{N} \sum_{i=1}^{N} \left| x_i - y_i \right|.
\end{equation}
The root mean square error (RMSE), used as a tie-breaking criterion during FFT parameter tuning (\secref{sec:paramtuning}) and reported alongside kernel count accuracy, was computed as
\begin{equation}
    \mathrm{RMSE} = \sqrt{\frac{1}{N} \sum_{i=1}^{N} (x_i - y_i)^2}.
\end{equation}
The mean absolute percentage error (MAPE), used to report kernel count accuracy on the 168-ear held-out set, was computed as
\begin{equation}
    \mathrm{MAPE} = \frac{100\%}{N} \sum_{i=1}^{N} \left| \frac{x_i - y_i}{y_i} \right|,
\end{equation}
where $y_i > 0$ is the true (manually annotated) value for ear $i$ and $x_i$ the corresponding prediction. For circular quantities (hue, in degrees), the circular MAE was computed using the wrapped angular difference,
\begin{equation}
    \mathrm{MAE}_{\mathrm{circ}} = \frac{1}{N} \sum_{i=1}^{N} \min\!\left(\left|x_i - y_i\right|,\; 360^{\circ} - \left|x_i - y_i\right|\right).
\end{equation}
The coefficient of determination ($R^{2}$), used to assess agreement between predicted and true per-ear kernel counts, was computed as
\begin{equation}
    R^{2} = 1 - \frac{\sum_{i=1}^{N} (y_i - x_i)^2}{\sum_{i=1}^{N} (y_i - \bar{y})^2},
\end{equation}
where $\bar{y}$ is the mean of the true values $\{y_i\}$. KRN accuracy on the held-out set was additionally summarized as a tolerance-based accuracy rate: the fraction of ears whose predicted KRN fell within $k$ rows of the manually annotated value,
\begin{equation}
    \mathrm{Acc}_{\pm k} = \frac{1}{N} \sum_{i=1}^{N} \mathbb{1}\!\left[\, \left| x_i - y_i \right| \leq k \,\right],
\end{equation}
where $\mathbb{1}[\cdot]$ is the indicator function; $k = 2$ was used when reporting the headline KRN accuracy figure.

Computational cost was quantified as the end-to-end wall-clock runtime (in seconds) required to process a single ear from the input point cloud to the final extraction of all eleven phenotypic traits, measured on the hardware reported in \tableref{tab:compute_cost}.

To characterize axial variation in kernel shape across the population, each kernel centroid was assigned a normalized longitudinal position,
\begin{equation}
    z_{\mathrm{norm}}
    =
    \frac{z-z_{\min}}{z_{\max}-z_{\min}},
\end{equation}
where $z$ is the kernel centroid $z$-coordinate, and $z_{\min}$ and $z_{\max}$ are the minimum and maximum kernel centroid $z$-coordinates of the corresponding ear. The normalized coordinate was partitioned into three equal intervals corresponding to the bottom, middle, and top thirds of the ear.

For ear $i$, the mean kernel aspect ratio within height band $b$ was computed as
\begin{equation}
    \bar{AR}_{i,b}
    =
    \frac{1}{N_{i,b}}
    \sum_{k=1}^{N_{i,b}}
    AR_{i,b,k},
\end{equation}
where $AR_{i,b,k}$ denotes the aspect ratio of kernel $k$ within height band $b$, and $N_{i,b}$ is the number of kernels assigned to that band for ear $i$.

Population-level summaries for each height band were then obtained by averaging the corresponding per-ear means,
\begin{equation}
    \bar{AR}_{b}
    =
    \frac{1}{M}
    \sum_{i=1}^{M}
    \bar{AR}_{i,b},
\end{equation}
where $M$ is the total number of analyzed ears. The reported standard deviation (SD) for each height band was computed over the corresponding set of per-ear mean aspect ratios,
\begin{equation}
    SD_b
    =
    \sqrt{
    \frac{1}{M-1}
    \sum_{i=1}^{M}
    \left(
    \bar{AR}_{i,b}
    -
    \bar{AR}_{b}
    \right)^2
    }.
\end{equation}

To capture spatial differences in kernel shape development, the bounding box aspect ratio (AR) was also computed separately over the base, middle, and tip thirds of each ear. The resulting aspect ratio by thirds is compiled in \tableref{tab:ar_by_thirds} and interpreted in the Results and Discussion sections.

\subsubsection*{Distributional Agreement via Earth-Mover Distance}
\label{sec:emd}

With all eleven traits defined, the remaining subsections describe the validation protocols: a distributional agreement metric, the synthetic ground-truth protocol (\secref{sec:synthetic_methods}), kernel matching via Hungarian assignment, and trait-specific evaluation notes. To quantify agreement between automated and manually annotated kernels-per-row (KPR) and Gabriel graph neighbor-count distributions on the six-ear manually annotated subset, the Earth-Mover Distance (EMD), equivalent to the first Wasserstein distance \citep{rubner2000earth}, was computed separately for each ear and each trait between the empirical distribution of manually annotated values and the corresponding automated predictions. For two one-dimensional empirical distributions with cumulative distribution functions $F_P$ and $F_Q$, the EMD is defined as
\begin{equation}
    \mathrm{EMD}(P, Q) = \int_{-\infty}^{\infty} \left| F_P(x) - F_Q(x) \right| \, dx,
\label{eq:emd}
\end{equation}
computed here using the \texttt{scipy.stats.wasserstein\_distance} implementation~\citep{virtanen2020scipy}. Lower EMD values indicate closer agreement between the automated and manual distributions, with an EMD of zero indicating identical distributions. Since the EMD preserves the physical units of the underlying trait, an EMD value below 1.0 indicates that the cumulative distribution of automated predictions deviates from the manual annotation distribution by less than one unit of the trait across the population.

\subsubsection*{Synthetic Ground-Truth Validation Protocol}
\label{sec:synthetic_methods}

Manual annotation of all eleven ear- and kernel-level traits at exact, pixel-perfect accuracy is infeasible at scale. Per-trait extraction accuracy was therefore additionally assessed against a complementary dataset of 27 procedurally generated synthetic ears. Each synthetic ear was built directly from a set of known geometric and lattice parameters (ear length, curvature, cross-sectional taper, kernel row number (KRN), kernels per row (KPR), and per-kernel ellipsoid dimensions), with no fitting to any individual real scan. Construction placed ellipsoidal kernels on a parametric cob: a constant-curvature centerline defined the ear's axial spine, and a radius profile governed cross-sectional size from the base through a maximum-width position to the tip. Kernels were arranged on a $\mathrm{KRN} \times \mathrm{KPR}$ lattice (KRN even, in $[8,24]$), with an optional alternate-row circumferential offset and per-kernel jitter in size and circumferential angle; the angular jitter also perturbs each kernel's position along its row.

Kernels were removed by a tip-biased Bernoulli dropout, with each kernel's drop probability increasing linearly with its normalized axial position $\tau_k \in [0.06, 0.97]$ (row positions were sampled with a small margin left at the base and tip; 0 = base, 1 = tip):
\begin{equation}
p_{\mathrm{drop}}(\tau_k) = f_{\mathrm{miss}}\,(0.4 + 1.2\,\tau_k),
\end{equation}
where $f_{\mathrm{miss}}$ is the ear's missing-fraction parameter. Each kernel was retained independently with probability $1-p_{\mathrm{drop}}(\tau_k)$, so a kernel at the base row was roughly $0.47f_{\mathrm{miss}}$ likely to be dropped and one at the tip row roughly $1.56f_{\mathrm{miss}}$ likely to be dropped, emulating incomplete tip fill. The realized fraction of missing kernels on a given ear therefore fluctuates stochastically around $f_{\mathrm{miss}}$ instead of matching it exactly.

For each kernel, only the outward-facing hemispherical cap of its ellipsoid, the portion visible to an external camera, was sampled into 3D points, mirroring the cob-facing occlusion present in the empirical pipeline. Gaussian along-normal position noise was then added to approximate reconstruction noise. Default parameter values were calibrated once, offline, to match the aggregate geometry, color, and row count of the independent six-ear manually annotated subset described in \secref{sec:manual_validation}, while no real point cloud entered the generation of any individual synthetic ear. Ground truth was therefore known exactly by construction rather than measured post hoc.

Ground-truth kernel count, KRN, and per-row kernel count (KPR) were recorded directly from the generating lattice; per-kernel surface area and volume were computed analytically from each ellipsoid's three known semi-axes using the Thomsen approximation for surface area~\citep{kresta2015mixing} (below), halved to reference the same outward cap geometry recovered by the empirical pipeline.
\begin{equation}
A(a,b,c) = 4\pi\left(\frac{(ab)^p + (ac)^p + (bc)^p}{3}\right)^{1/p},
\qquad p = 1.6075,
\end{equation}
with ground-truth cap surface area and volume referenced to the outward-facing hemisphere as $A_{\mathrm{cap}} = A/2$ and $V_{\mathrm{cap}} = \tfrac{4}{3}\pi abc/2$. The 27 synthetic ears were generated as a full $3\times3\times3$ factorial sweep over kernel row number ($\mathrm{KRN}\in\{12,16,20\}$), spine curvature ($\in\{0,\,0.002,\,0.004\}\ \mathrm{rad\,mm^{-1}}$), and tip-biased missing-kernel fraction ($\in\{0,\,0.05,\,0.15\}$), with all other parameters held at their calibrated defaults and a distinct reproducible seed assigned to each of the 27 combinations.

Two ground-truth quantities carry documented limitations relative to the pipeline's corresponding outputs. First, ground-truth hue is the population mean hue parameter used to generate each kernel's color, not the specific per-kernel hue actually sampled (each kernel's hue is an independent draw around this mean); it is therefore a distributional reference for aggregate hue comparisons rather than an exact per-kernel target. Second, ground-truth neighbor count and mean neighbor distance are computed from lattice-topological adjacency (row/column neighbors on the generating grid, including brick-offset diagonals, with periodic wraparound in the circumferential direction) instead of from a Gabriel graph. These packing-trait comparisons should therefore be read as evaluating agreement in overall packing density and topology rather than an exact edge-by-edge match to the pipeline's cylindrical Gabriel graph construction.

The full pipeline (PCA-based Z-axis alignment, cylindrical unwrapping, CLAHE contrast enhancement, CPSAM segmentation, BPA-based surface area, convex hull volume, hue and aspect ratio extraction, Gabriel graph packing, and row detection via adaptive FFT and the row chain graph) was executed end-to-end on each synthetic point cloud. The same implementation was used as for the 168-ear held-out set, with no access to the ground-truth kernel table at any stage.

\subsubsection*{Kernel Matching via Hungarian Assignment}

For a given ear, let $\{\mathbf{p}_i\}_{i=1}^{N_{\mathrm{pred}}}$ denote the predicted 3D kernel centroids and $\{\mathbf{g}_j\}_{j=1}^{N_{\mathrm{gt}}}$ the ground-truth centroids, both expressed in the shared PCA-aligned reference frame. The pairwise cost matrix was defined as the Euclidean centroid distance,
\begin{equation}
    C_{ij} = \left\lVert \mathbf{p}_i - \mathbf{g}_j \right\rVert_2,
\end{equation}
and the globally optimal one-to-one correspondence $\pi^{*}$ between predicted and ground-truth kernels, over the full set of predicted and ground-truth kernels for that ear, was obtained by solving the linear assignment problem
\begin{equation}
    \pi^{*} = \underset{\pi}{\arg\min} \sum_{i} C_{i,\pi(i)},
\end{equation}
via the Hungarian algorithm ~\citep{kuhn1955hungarian,crouse2016assignment}. Each optimal pair $(i, \pi^{*}(i))$ was then filtered post hoc against a per-ear distance cutoff,
\begin{equation}
    C_{i,\pi^{*}(i)} \leq \tau_{\mathrm{match}}, \qquad \tau_{\mathrm{match}} = \tfrac{1}{2}\, \mathrm{median}_{j}\!\left(\bar{d}^{\,\mathrm{gt}}_{j}\right),
\end{equation}
where $\bar{d}^{\,\mathrm{gt}}_{j}$ is the ground-truth lattice-topological mean neighbor distance of kernel $j$, so that $\tau_{\mathrm{match}}$ scales naturally with each ear's kernel density. Pairs exceeding $\tau_{\mathrm{match}}$ were rejected as spurious; the corresponding predicted kernel was recorded as a false positive and the corresponding ground-truth kernel as a false negative. This post hoc distance filter prevents distant, incidental pairings, which the Hungarian algorithm may still return as globally optimal when $N_{\mathrm{pred}} \neq N_{\mathrm{gt}}$, from being scored as correct matches.

\subsubsection*{Trait-Specific Evaluation Notes}

For each of the eleven traits, MAE (or circular MAE, for hue) was computed across matched kernel pairs using the definitions in \secref{sec:metrics}, then averaged across all 27 ears. Kernels-per-row (KPR) MAE was computed after aligning predicted row labels to ground-truth row labels via a second, independent linear assignment, which maximizes shared kernel count over a row-membership contingency table instead of minimizing centroid distance. This MAE was averaged only over rows present in both the aligned predicted and ground-truth sets, so rows absent from one side after alignment do not contribute.

That is,
\begin{equation}
\mathrm{MAE}_{\mathrm{KPR}} = \frac{1}{|\mathcal{R}|}\sum_{r\in\mathcal{R}} \left| n^{\mathrm{pred}}_r - n^{\mathrm{gt}}_r \right|,
\end{equation}
where $\mathcal{R}$ is the set of row labels present in both the aligned predicted rows and the ground-truth rows. Ground-truth aspect ratio was derived from each kernel's three known ellipsoid semi-axes, sorted in descending order and ratioed as
\begin{equation}
\ell^{\mathrm{gt}}_{(1)} \geq \ell^{\mathrm{gt}}_{(2)} \geq \ell^{\mathrm{gt}}_{(3)} = \mathrm{sort}\!\left(a_{\mathrm{circ}}, a_{\mathrm{axial}}, a_{\mathrm{radial}}\right), \qquad \mathrm{AR}^{\mathrm{gt}}_k = \frac{\ell^{\mathrm{gt}}_{(1)}}{\ell^{\mathrm{gt}}_{(2)}}.
\end{equation}
The predicted aspect ratio was instead computed from the axis-aligned bounding box of the BPA-surviving point set (\secref{sec:aspect_ratio}), so the two constructions are analogous without being identical. The axial hue and aspect ratio profiles were evaluated by comparing predicted and ground-truth values within matched 10\%-height bands along the PCA-aligned ear axis, each computed independently from each set's own height range.

% ============================================================
\section{Results}
\label{sec:results}
% ============================================================

\begin{figure}[t!]
  \centering
  \usetikzlibrary{positioning,arrows.meta,backgrounds,calc}

\definecolor{cRecon}{HTML}{2C6FBB}
\definecolor{cSeg}{HTML}{0097A7}
\definecolor{cMethod}{HTML}{D9821A}
\definecolor{cHub}{HTML}{6E54AB}
\definecolor{cData}{HTML}{2E8B57}

\resizebox{0.95\linewidth}{!}{
\renewcommand{\familydefault}{\sfdefault}
\begin{tikzpicture}[
  font=\footnotesize,
  >={Stealth[length=2.1mm]},
  base/.style ={rounded corners=2pt, draw, align=center, inner sep=3pt,
                minimum height=8mm, minimum width=58mm, text width=52mm,
                line width=0.5pt},
  hub/.style   ={base, draw=cHub,   fill=cHub!24, minimum width=46mm, text width=42mm},
  recon/.style ={base, draw=cRecon, fill=cRecon!22},
  seg/.style   ={base, draw=cSeg,   fill=cSeg!24},
  method/.style={base, draw=cMethod,fill=cMethod!30},
  card/.style  ={rounded corners=2pt, draw=cData, fill=cData!30, align=left,
               inner sep=5pt, minimum width=58mm,
               text width=58mm, line width=0.5pt, font=\scriptsize,
               anchor=north},
  wide/.style  ={rounded corners=2pt, draw=cData, fill=cData!30, align=left,
                 inner sep=5pt, minimum width=110mm, text width=110mm,
                 line width=0.5pt, font=\scriptsize, anchor=north},
  data/.style  ={rounded corners=2pt, draw=cData, fill=cData!30, align=center,
                 inner sep=5pt, line width=0.9pt, minimum width=110mm,
                 text width=110mm, font=\footnotesize, anchor=north},
  arr/.style   ={->, line width=0.6pt, draw=black!72},
]

\def\cL{0}   \def\cM{6.4}  \def\cR{12.8}

\node[hub] (results) at (\cM,0) {\textbf{Results:} Three validation tiers};

\node[recon]  (synthetic) at (\cL,-2.2)  {\textbf{Synthetic} \\ 27 ears, exact geometry};
\node[seg]    (heldout)   at (\cM,-2.2)  {\textbf{Held-Out} \\ 168 ears, manual annotation};
\node[method] (manualv)   at (\cR,-2.2)  {\textbf{Manual Validation} \\ 6 ears, exhaustive};

\node[card] (synth_card) at (\cL,-3.4)
  {\textbf{Per-trait accuracy vs.\ exact ground truth}\\[2pt]
   \textbullet\ Kernel count agreement (\tableref{tab:pooled_metrics})\\
   \textbullet\ Geometry, color, shape traits (\tableref{tab:pooled_metrics})\\
   \textbullet\ Row-structure traits (\tableref{tab:pooled_metrics})};

\node[card] (held_card1) at (\cM,-3.4)
  {\textbf{Kernel count \& KRN accuracy}\\[2pt]
   \textbullet\ Predicted vs.\ manual kernel counts (\figref{fig:kernel_count_results})\\
   \textbullet\ Predicted vs.\ manual row counts (\figref{fig:kernel_rows_results})\\
   \mbox{}};

\node[card] (manual_card) at (\cR,-3.4)
  {\textbf{KPR \& neighbor-count agreement}\\[2pt]
   \textbullet\ Distributional agreement, EMD (\figref{fig:sample1_accuracy})\\
   \textbullet\ Predicted vs.\ ground-truth KRN (\tableref{tab:krn_accuracy})\\
   \mbox{}};

\node[card] (held_card2) at (\cM,-6.6)
  {\textbf{Benchmarking \& compute cost}\\[2pt]
   \textbullet\ Comparison against prior methods (\tableref{tab:benchmarking})\\
   \textbullet\ Per-ear processing time by stage (\tableref{tab:compute_cost})\\
   \mbox{}};

\node[wide] (held_card3) at (\cM,-9.6)
  {\textbf{Qualitative per-ear distributions} (4 representative ears)\\[3pt]
   \begin{minipage}[t]{180mm}
   \textbullet\ Surface area (\figref{fig:area_3d}, \figref{fig:surface_area}, \figref{fig:surface_area_thirds})\\[1pt]
   \textbullet\ Volume proxy (\figref{fig:volume_3d}, \figref{fig:volume}, \figref{fig:volume_thirds})\\[1pt]
   \textbullet\ Packing: neighbor count \& distance (\figref{fig:nndist_3d}, \figref{fig:neighbor_dist}, \figref{fig:mean_distance})\\[1pt]
   \textbullet\ Row-level kernel counts (\figref{fig:row_distribution})\\[1pt]
   \textbullet\ Hue \& aspect ratio (\figref{fig:hue}, \figref{fig:aspect_ratio})\\[1pt]
   \textbullet\ Axial hue \& aspect profiles (\figref{fig:hue_by_height}, \figref{fig:ar_by_height})\\[1pt]
   \textbullet\ Full-dataset trait distributions (\figref{fig:combined_distributions})
   \end{minipage}};

\node[data] (out) at (\cM,-14.0)
  {\textbf{Multi-tiered validation of all eleven traits}\\
   Pipeline accuracy under exact ground truth (synthetic)\\
   Breeding-scale accuracy on real ears (held-out)\\
   Spatial tracking fidelity at kernel resolution (manual) };

% ---------- hub fan-out: orthogonal only ----------
\coordinate (fan) at ($(results.south)+(0,-0.55)$);
\draw[line width=0.6pt, draw=black!72] (results.south) -- (fan);
\draw[line width=0.6pt, draw=black!72] (synthetic.north |- fan) -- (manualv.north |- fan);
\draw[arr] (synthetic.north |- fan) -- (synthetic.north);
\draw[arr] (heldout.north   |- fan) -- (heldout.north);
\draw[arr] (manualv.north   |- fan) -- (manualv.north);

% ---------- spine ----------
\draw[arr] (synthetic.south)  -- (synth_card.north);
\draw[arr] (heldout.south)    -- (held_card1.north);
\draw[arr] (held_card1.south) -- (held_card2.north);
\draw[arr] (held_card2.south) -- (held_card3.north);
\draw[arr] (manualv.south)    -- (manual_card.north);

% ---------- collector bus below the wide card ----------
\coordinate (bus) at ($(held_card3.south)+(0,-0.7)$);
\draw[line width=0.6pt, draw=black!72] (synth_card.south)  |- (bus);
\draw[line width=0.6pt, draw=black!72] (manual_card.south) |- (bus);
\draw[line width=0.6pt, draw=black!72] (held_card3.south)  -- (bus);
\draw[line width=0.6pt, draw=black!72] (synth_card.south -| bus) -- (manual_card.south -| bus);
\draw[arr] (bus) -- (out.north);

% ---------- background bands ----------
\begin{scope}[on background layer]
  \fill[cRecon!4,  rounded corners=3pt]
    ($(synthetic.north west)+(-0.3,0.3)$) rectangle ($(synth_card.south east)+(0.3,-0.3)$);
  \fill[cSeg!4,    rounded corners=3pt]
    ($(heldout.north west)+(-0.3,0.3)$)   rectangle ($(held_card2.south east)+(0.3,-0.3)$);
  \fill[cMethod!3, rounded corners=3pt]
    ($(manualv.north west)+(-0.3,0.3)$)   rectangle ($(manual_card.south east)+(0.3,-0.3)$);
\end{scope}

\end{tikzpicture}
}
  \caption{Overview of the multi-tiered validation framework presented in \secref{sec:results}: exact ground-truth validation using a 27-ear synthetic dataset (\secref{sec:synthetic_validation}), breeding-scale accuracy using a 168-ear held-out set (\secref{sec:168_results}), and kernel-resolution tracking validation on a six-ear manually annotated subset (\secref{sec:manual_validation}).}
  \label{fig:results_tree}
\end{figure}
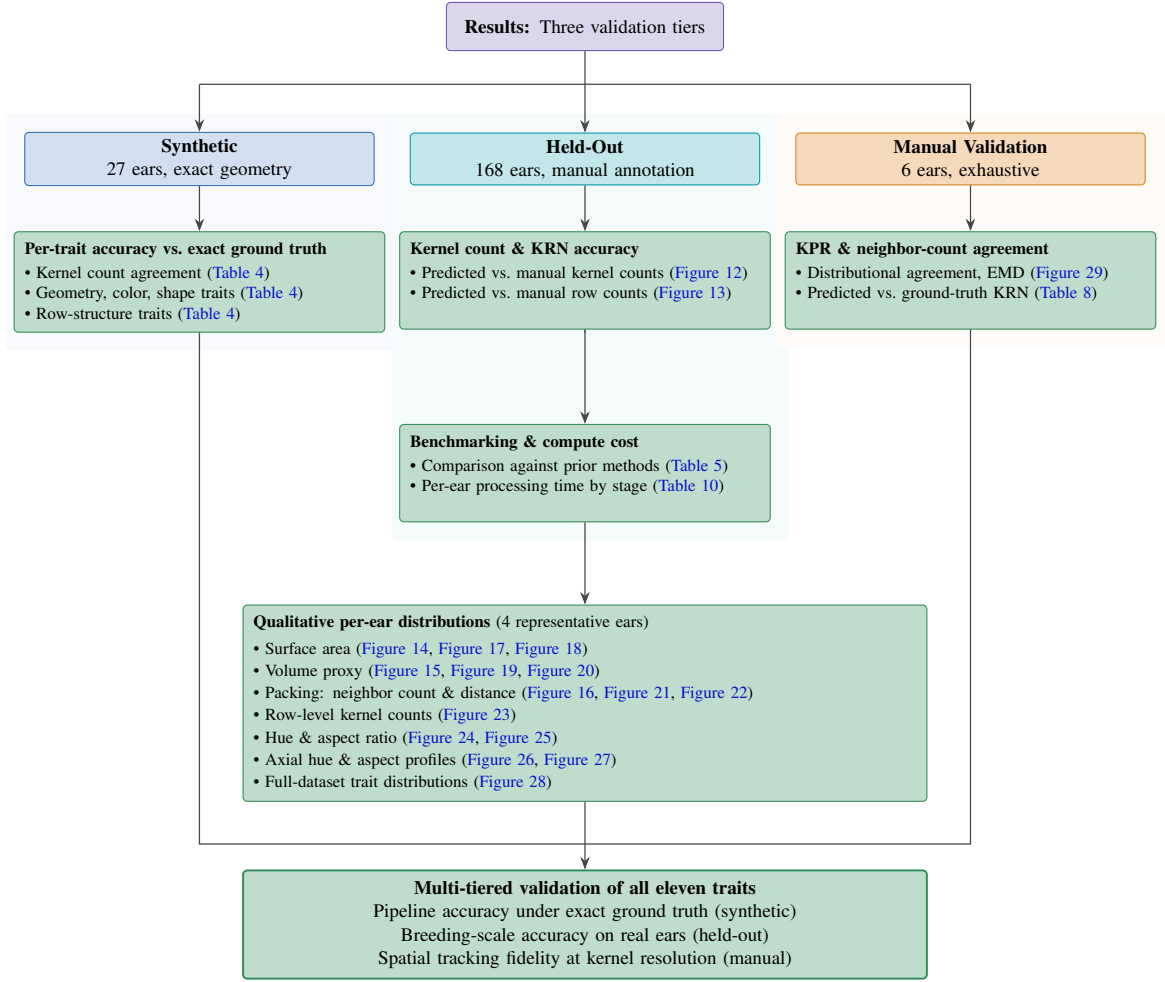

Results are presented across three complementary validation tiers, structurally summarized in \figref{fig:results_tree}, preceded by an overview of the parameter tuning configuration and followed by a computational cost analysis. \secref{sec:paramtuning} details the optimization of pipeline parameters. \secref{sec:synthetic_validation} evaluates per-trait accuracy against the 27-ear synthetic dataset, where exact ground truth isolates pipeline performance from physical measurement error. \secref{sec:168_results} reports kernel count and KRN accuracy on the 168-ear held-out set, situates the pipeline against existing benchmarks, and provides qualitative per-ear trait distributions for four representative ears. \secref{sec:manual_validation} assesses spatial packing and row-tracking fidelity against the six-ear manually annotated subset. \secref{sec:compute} details the computational execution times and hardware scalability of the workflow.

\subsection{Parameter Tuning Dataset}
\label{sec:paramtuning}

All empirically tuned parameters were optimized on the 100-ear tuning set, drawn from the 268-ear labeled dataset, with the remaining 168-ear held-out set reserved for final evaluation. The 100-ear tuning set was constructed to be representative of the full dataset in two steps. First, 50 ears were selected manually by choosing samples spanning the full range of ear curvature, prioritizing highly curved and moderately curved ears (assessed visually prior to selection) to ensure adequate coverage of challenging geometries. Second, the remaining 50 ears were selected to match the kernel row number distribution of the full 268-ear labeled dataset: at least three ears of each observed row count category were included (or all available ears where fewer than three existed), and the remaining slots were filled by sampling each category in proportion to its frequency in the full dataset, with specific ears chosen randomly within each category. All parameters reported in subsequent sections were tuned on this 100-ear tuning set and evaluated on the 168-ear held-out set. Because the tuning set was deliberately enriched for highly curved and moderately curved ears, it is not a random sample of the labeled dataset and constitutes a more challenging population than the held-out set. Performance on the tuning set therefore places no upper bound on held-out performance.

Parameters targeting kernel count accuracy were optimized sequentially in pipeline order across three stages (stage 1: cylindrical unwrapping parameters; stage 2: CLAHE parameters; stage 3: CPSAM parameters), minimizing MAE against manual kernel counts on the 100-ear tuning set. Within each stage, individual parameters were swept one at a time in the order listed, with each parameter sweep constituting a step within that stage. The FFT smoothing kernel size and CV threshold, which govern row count accuracy, were optimized jointly in stage 4, independently of the three sequential kernel count tuning stages, as the two-dimensional parameter space was small enough to enumerate exhaustively.

\subsection{Synthetic Ground-Truth Validation}
\label{sec:synthetic_validation}

Per-trait extraction accuracy, assessed against the 27-ear synthetic dataset described in \secref{sec:synthetic_methods}, is summarized in \tableref{tab:pooled_metrics}. For each of the eleven traits, mean absolute error (MAE), or circular MAE for hue, was computed across matched kernel pairs within an ear, then averaged across all 27 ears.

\begin{table}[t!]
\centering
\caption{Performance metrics across phenotypic traits on the 27-ear synthetic dataset. KC achieved an $R^{2}$ of 0.769.}
\label{tab:pooled_metrics}
\footnotesize
\begin{tabular}{lrl}
\toprule
\textbf{Trait} & \textbf{MAE} & \textbf{Unit} \\
\midrule
KC & 26.48 & kernels \\
KRN & 7.26 & rows \\
KPR & 11.28 & kernels \\
SA & 9.19 & mm\textsuperscript{2} \\
Vol & 6.62 & mm\textsuperscript{3} \\
Neighbor Count & 2.05 & neighbors \\
Mean NN Distance & 1.23 & mm \\
Circular Hue & 3.24 & degrees \\
Aspect Ratio & 0.14 & unitless \\
Circular Hue vs. Height & 2.30 & degrees \\
Aspect Ratio vs. Height & 0.04 & unitless \\
\bottomrule
\end{tabular}
\end{table}

Quantitative evaluation on the synthetic dataset reveals a higher error rate in row-number detection compared to the real-ear dataset, with a pooled KRN MAE of 7.26 rows (\tableref{tab:pooled_metrics}). To investigate the root cause of this discrepancy, we analyzed individual failure cases in the frequency analysis step.

\begin{figure}[htbp]
    \centering
    \includegraphics[width=0.9\linewidth]{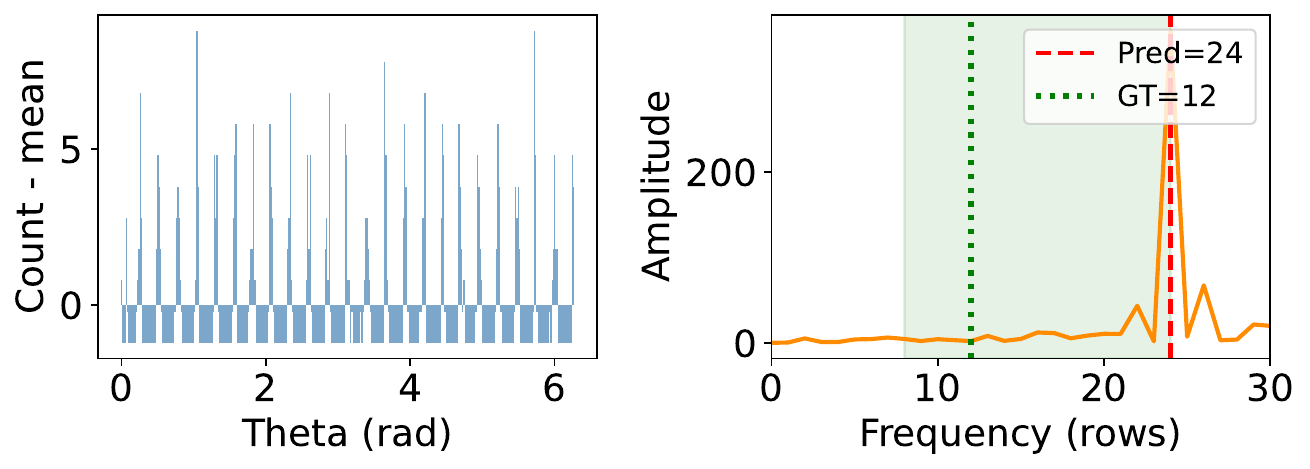}
    \caption{FFT row-detection failure due to harmonic locking on a synthetic ear lattice ($\text{KRN}=12$, zero curvature, fully-filled ear with no missing kernels). (Left) The highly regular smoothed azimuthal histogram. (Right) The corresponding FFT spectrum showing a prominent secondary structural harmonic at $f = 24$. The adaptive peak-selection step locks onto this harmonic peak ($\text{Pred}=24$) rather than the true fundamental frequency ($\text{GT}=12$).}
    \label{fig:synthetic_fft_failure}
\end{figure}

Real ears carry natural biological variation and structural irregularities that smooth the spatial frequency spectrum into a single dominant peak. The synthetic ears instead possess a perfectly regular, rigid lattice structure. This mathematical regularity introduces discrete grid-aliasing artifacts and prominent structural harmonics. Because the adaptive peak-selection step searches for a dominant frequency, these sharp secondary harmonic spikes can mislead it. When a high-frequency harmonic spike carries a greater amplitude than the true fundamental frequency, the adaptive selection step locks onto it, resulting in a predicted row count that represents the harmonic rather than the ground truth. This specific failure mode is illustrated in \figref{fig:synthetic_fft_failure} using a synthetic ear generated with a true kernel row number of 12, where the peak selection locks onto the dominant harmonic spike at a frequency of 24.

The geometric and color traits (surface area, convex hull volume proxy, hue, aspect ratio, and both axial profiles) show close agreement with the generating parameters, confirming that the segmentation and per-kernel trait extraction stages recover fine-grained relative shape descriptors under known ground truth. The row-structure traits (KRN and KPR) show substantially larger error on this dataset than on the real held-out set; this discrepancy is examined in \secref{sec:discussion}.

\FloatBarrier

\subsection{Held-Out Set Validation and Benchmarking}
\label{sec:168_results}

\begin{figure}[t!]
\centering
\begin{subfigure}[b]{0.49\textwidth}
\includegraphics[width=0.99\textwidth]{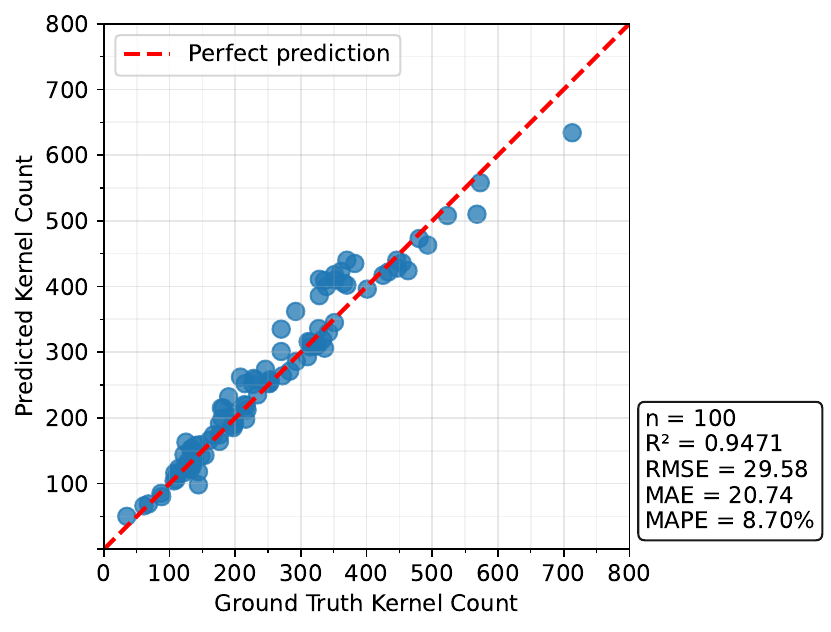}
\caption{100-Ear Tuning Set}
\end{subfigure}
\begin{subfigure}[b]{0.49\textwidth}
\includegraphics[width=0.99\textwidth]{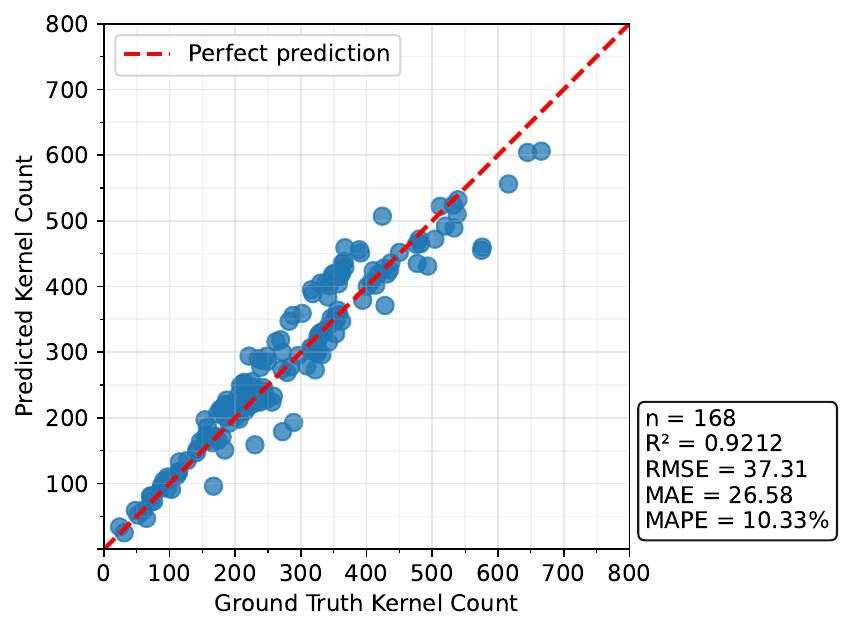}
\caption{168-Ear Held-Out Set}
\end{subfigure}
\caption{Kernel count prediction accuracy for (a) the 100-ear tuning set and (b) the 168-ear held-out set.}
\label{fig:kernel_count_results}
\end{figure}

\begin{figure}[t!]
\centering
\begin{subfigure}[b]{0.49\textwidth}
\includegraphics[width=0.99\textwidth]{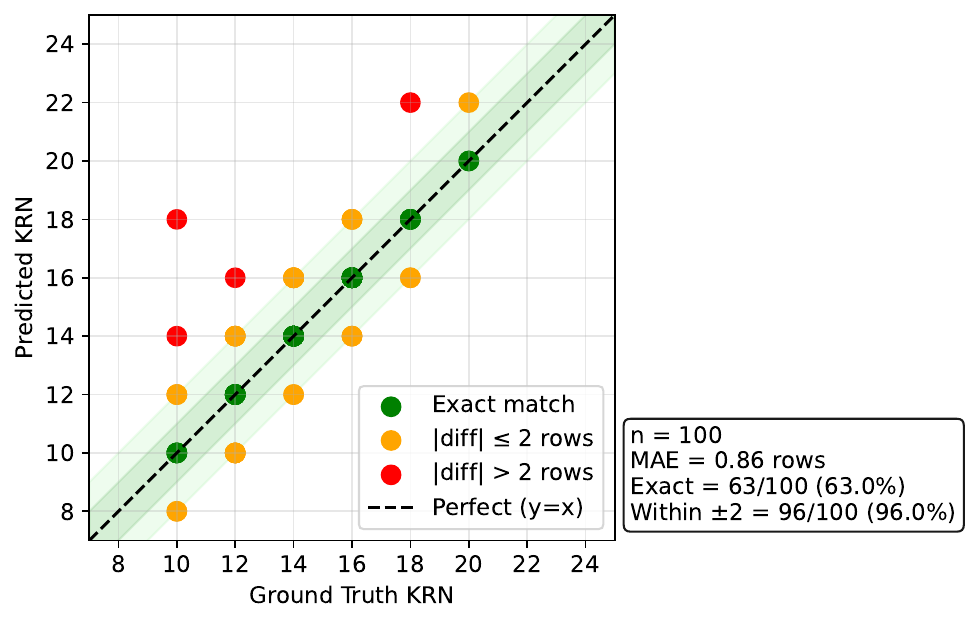}
\caption{100-Ear Tuning Set}
\end{subfigure}
\begin{subfigure}[b]{0.49\textwidth}
\includegraphics[width=0.99\textwidth]{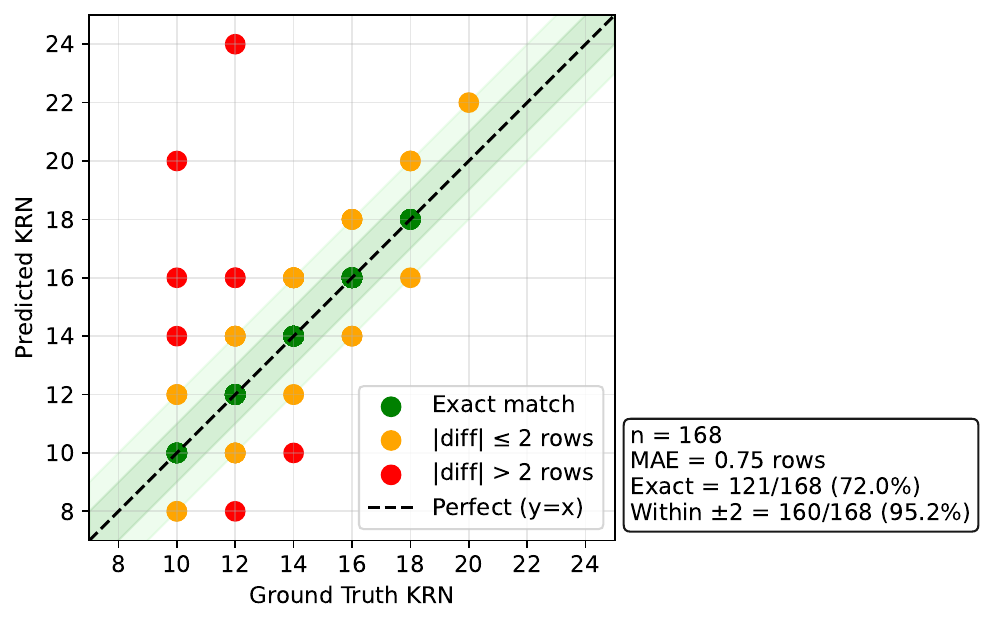}
\caption{168-Ear Held-Out Set}
\end{subfigure}
\caption{KRN prediction accuracy for (a) the 100-ear tuning set and (b) the 168-ear held-out set.}
\label{fig:kernel_rows_results}
\end{figure}

Of the 268-ear labeled dataset, the 168-ear held-out set was reserved for validation. Performance is reported side-by-side for both the tuning and validation sets in \figref{fig:kernel_count_results} and \figref{fig:kernel_rows_results}.

% ------ Kernel Count Accuracy ------

% ------ Row Count Accuracy ------

KRN was predicted exactly for 121 of 168 held-out ears (72.0\%) and within $\pm$2 rows for 160 of 168 (95.2\%), against 63 of 100 (63.0\%) exact and 96 of 100 (96.0\%) within $\pm$2 on the tuning set. The higher exact-match rate on the held-out set reflects the composition of the tuning set, not an anomaly of the model: as described in \secref{sec:paramtuning}, the tuning set was deliberately enriched for highly curved ears, on which the angular regularity assumption underlying the FFT row detection is most often violated.

\tableref{tab:benchmarking} compares hardware class, dataset size, trait count, and reported accuracy across existing automated ear phenotyping methods and the present pipeline.

% ============================================================
% ULTRA-COMPACT UNIFIED BENCHMARKING TABLE
% ============================================================
\begin{table}[t!]
\caption{Comparative benchmarking of automated maize ear phenotyping methods.
Abbreviations are defined in the Abbreviations section.
$^{b}$~Single-path KPR evaluation~\protect\citep{maizeearSAM2025}.
$^{c}$~Evaluated across the full 574 test ears for KC, and selected benchmark subsets of 150 and 50 ears for KPR and KRN, respectively~\protect\citep{lu2025}.
$^{d}$~Distinct kernel-level phenotypic outputs.
$^{e}$~Pearson's correlation coefficient $r$. \protect\citet{warman2021} and \protect\citet{zhao2025} are excluded as neither reports KC, KRN, or KPR regression accuracy.
$^{f}$~105 ears used for primary trait (KC, KRN) validation; a 43-ear subset used for KPR~\protect\citep{fan2026openear}.}
\label{tab:benchmarking}
\centering
\small
\setlength{\tabcolsep}{3pt}
\renewcommand{\arraystretch}{0.85}
\begin{tabularx}{\textwidth}{@{} l >{\raggedright\arraybackslash}p{2.4cm} c c >{\raggedright\arraybackslash}X @{}}
\toprule
\textbf{Method} & \textbf{Hardware Class} & \textbf{Traits}$^{d}$ & \textbf{Dataset Size} & \textbf{Reported Metric Performance} \\
\midrule

\citet{makanza2018}
& Consumer Camera & 4 & 180 ears
& KC $r = 0.98^{e}$ \\

\citet{gonzalez2022}
& Document Scanner & 7 & 250 ears
& KRN Classification Rate $=0.68$ \\

\citet{shi2022}
& Specialist Camera & 3 & 300 ears
& KC $R^{2}=0.942$; KRN MAE $=0.32$; KPR MAE $=1.07$ \\

\citet{maizeearSAM2025}
& Smartphone Camera & 1 & 136 ears
& KPR $R^{2}=0.87^{b}$ \\

\citet{lu2025}
& Smartphone Camera & 3 & 574 ears$^{c}$
& KC $R^{2}=0.764$; KRN 80.3\% exact (MAE $=0.23$); KPR $R^{2}=0.645$ \\

\citet{sun2026mep3d}
& Industrial Scanner & 6 & 30 ears
& KC MAPE $=0.91\%$ ($R^{2}=0.992$); KRN 100\% exact \\

\citet{fan2026openear}
& Consumer-grade & 10 & 148 ears$^{f}$
& KC $R^{2}=0.980$; KRN $R^{2}=0.888$; KPR $R^{2}=0.852$ \\

\midrule

\textbf{Ours}
& \textbf{Consumer Video}
& \textbf{11}
& \textbf{168 ears}
& \textbf{KC MAPE = 10.33\% ($R^{2}=0.921$); KRN MAE = 0.75 (95.2\% within $\pm2$ rows)} \\

\bottomrule
\end{tabularx}
\end{table}

Disparate imaging protocols, varied crop varieties, and custom validation datasets limit direct performance comparisons across these methods; \tableref{tab:benchmarking} therefore serves as a high-level comparative assessment, not a controlled baseline experiment on identical samples. For instance, \citet{shi2022}, \citet{lu2025}, and this work report regression-based KC, while others rely on correlation coefficients or geometric indices. For \citet{gonzalez2022}, yield-component verification focuses on discrete KRN tier classification rather than continuous regression curves, yielding a 0.68 classification rate over a highly diverse germplasm pool. For \citet{maizeearSAM2025}, the tabulated $R^{2}=0.87$ denotes primary-path KPR evaluation (see footnote~$^{b}$).

While dedicated industrial 3D profiling frameworks~\citep{sun2026mep3d} achieve near-perfect counting precision, they rely on rigid, high-cost optical scanning configurations that restrict deployment scale. By contrast, 2D imaging methods~\citep{makanza2018, gonzalez2022, shi2022, maizeearSAM2025, lu2025, fan2026openear} maximize accessibility but are limited to surface-projected metrics and do not recover volumetric or packing geometry. The present pipeline occupies the intermediate position: it delivers spatial packing data and volumetric shape proxies from standard consumer video streams without specialized hardware calibration, at the cost of a higher kernel count error than the best-performing projection-based methods. The sources of that error are examined in \secref{sec:discussion}.

Held-out KC error was stratified by ground-truth KRN (\tableref{tab:krn_stratified_kc_error}) on the 168-ear held-out set to characterize where this error concentrates.

\begin{table}[t!]
\centering
\small
\caption{KC prediction error stratified by ground-truth KRN, 168-ear held-out
set. $n$ denotes the number of ears in each KRN class.}
\label{tab:krn_stratified_kc_error}
\begin{tabular}{ccccc}
\toprule
\textbf{KRN} & \textbf{n} & \textbf{MAE} & \textbf{Signed ME} & \textbf{MAPE} \\
\midrule
10 & 20 & 22.15 & $+10.55$ & 14.62\% \\
12 & 44 & 15.27 & $+4.95$  & 8.58\%  \\
14 & 46 & 20.43 & $-0.48$  & 8.26\%  \\
16 & 44 & 41.41 & $+7.27$  & 11.79\% \\
18 & 13 & 40.54 & $+22.38$ & 11.50\% \\
20$^{\dagger}$ & 1 & 62.00 & $+62.00$ & 16.89\% \\
\bottomrule
\multicolumn{5}{l}{\footnotesize $^{\dagger}$Only one ear in this KRN class ($n=1$); excluded from interpretation.}\\
\end{tabular}
\end{table}
\FloatBarrier

Mean bounding box aspect ratio was computed separately for the base, middle, and tip thirds of each ear to evaluate how kernel shape varies systematically along the longitudinal axis of the cob. \tableref{tab:ar_by_thirds} presents the resulting aspect ratio by thirds, in which mean kernel elongation is highest in the middle third and lowest at the tip.

\begin{table}[htbp]
\centering
\caption{Ear-level averaged kernel AR across vertical height thirds for 1{,}091 ears.}
\label{tab:ar_by_thirds}
\begin{tabular}{lc}
\toprule
\textbf{Height Band (Third)} & \textbf{Aspect Ratio (Mean $\pm$ SD)} \\
\midrule
Bottom Third (Base) & $1.29 \pm 0.11$ \\
Middle Third (Mid-ear) & $1.33 \pm 0.12$ \\
Top Third (Tip) & $1.26 \pm 0.09$ \\
\bottomrule
\end{tabular}
\end{table}

\begin{figure}[t!]
\centering
\begin{subfigure}[b]{0.24\textwidth}
    \centering
    \includegraphics[width=\textwidth]{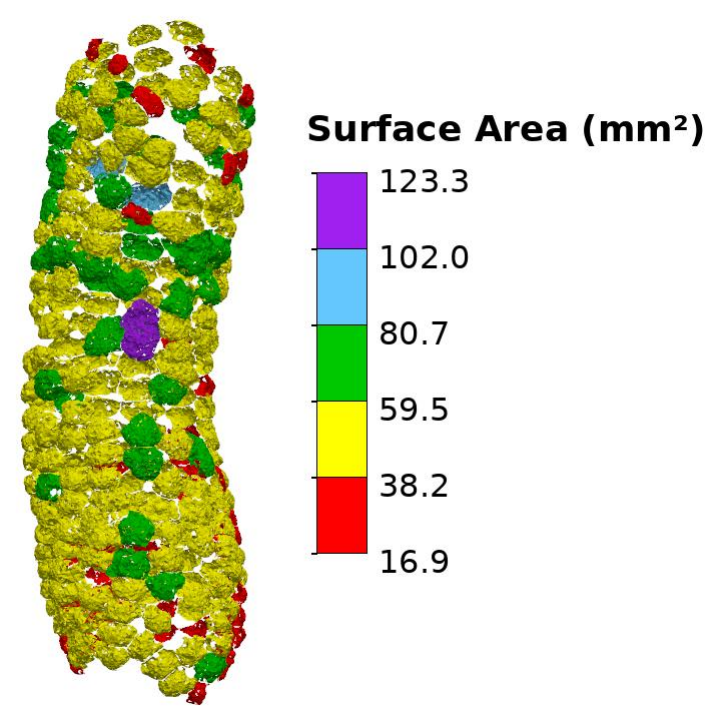}
    \caption{24-P2042\_3}
\end{subfigure}\hfill
\begin{subfigure}[b]{0.24\textwidth}
    \centering
    \includegraphics[width=\textwidth]{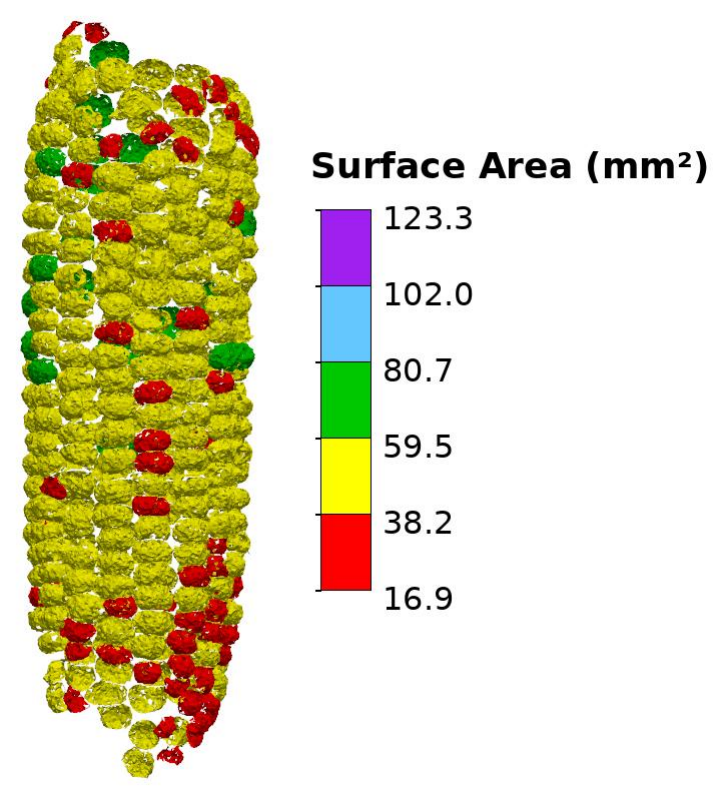}
    \caption{24-P2071\_2}
\end{subfigure}
\begin{subfigure}[b]{0.24\textwidth}
    \centering
    \includegraphics[width=\textwidth]{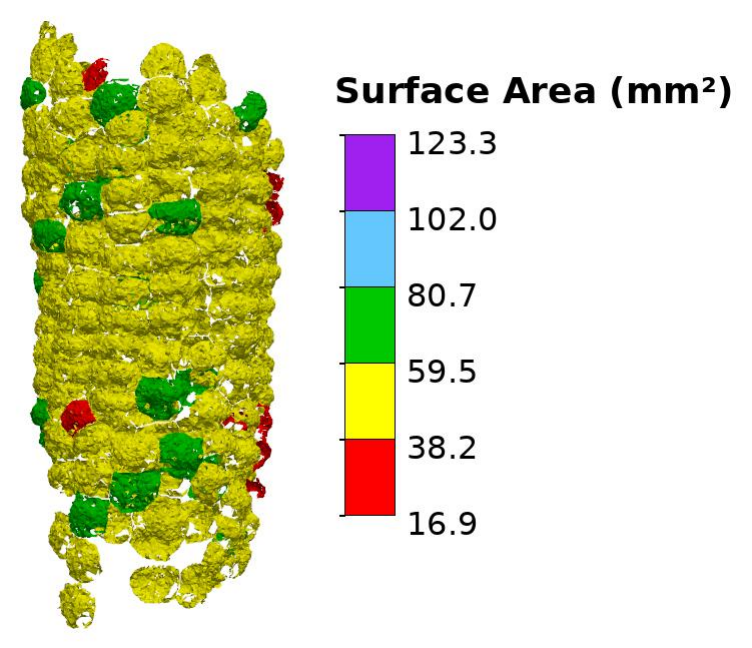}
    \caption{24-P2146\_2}
\end{subfigure}\hfill
\begin{subfigure}[b]{0.24\textwidth}
    \centering
    \includegraphics[width=\textwidth]{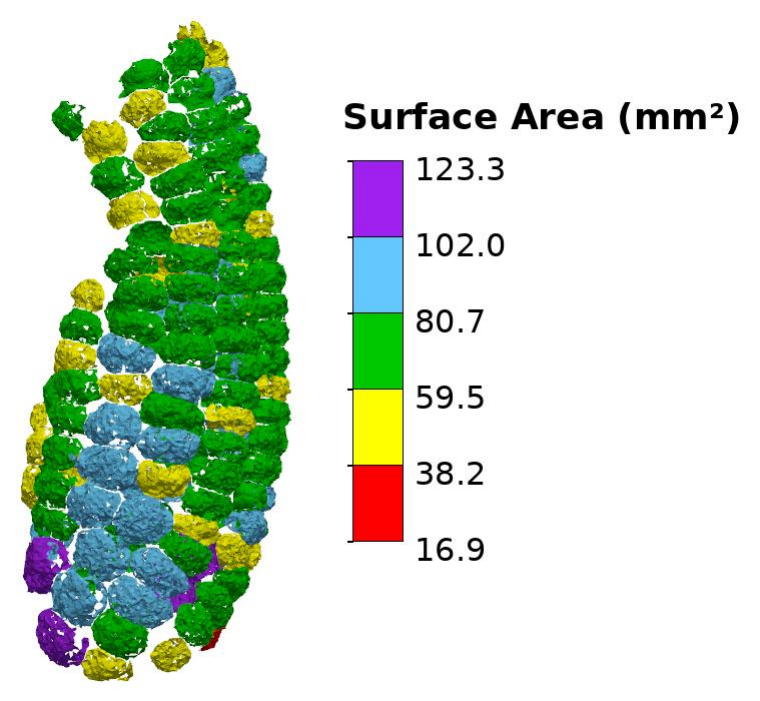}
    \caption{24-P2199\_1}
\end{subfigure}

\caption{Per-kernel surface area mapped onto the 3D ear point cloud for each representative ear. Color encodes surface area (mm\textsuperscript{2}).}
\label{fig:area_3d}
\end{figure}

\begin{figure}[t!]
\centering
\begin{subfigure}[b]{0.24\textwidth}
    \centering
    \includegraphics[width=\textwidth]{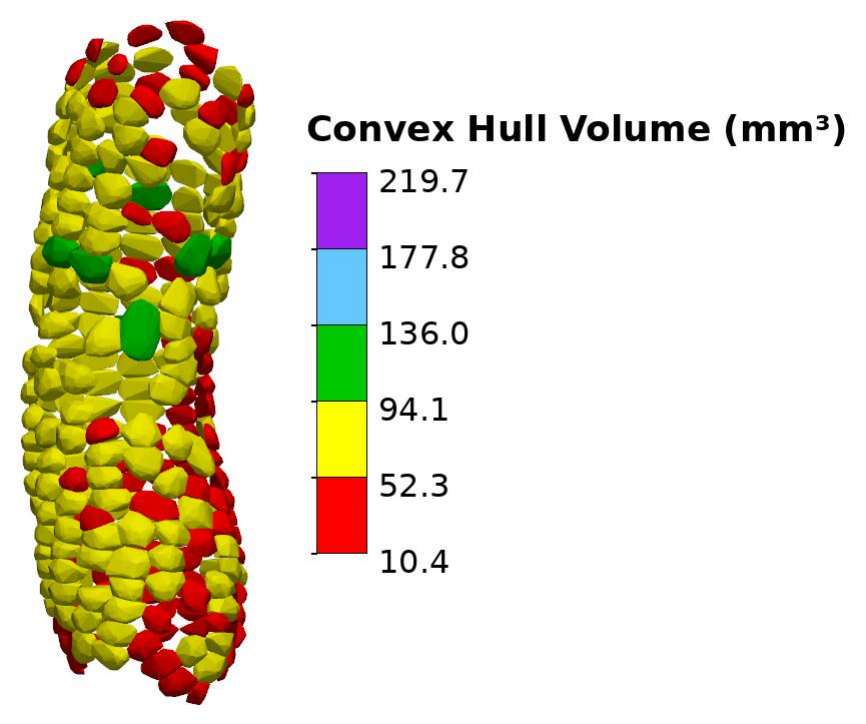}
    \caption{24-P2042\_3}
\end{subfigure}\hfill
\begin{subfigure}[b]{0.24\textwidth}
    \centering
    \includegraphics[width=\textwidth]{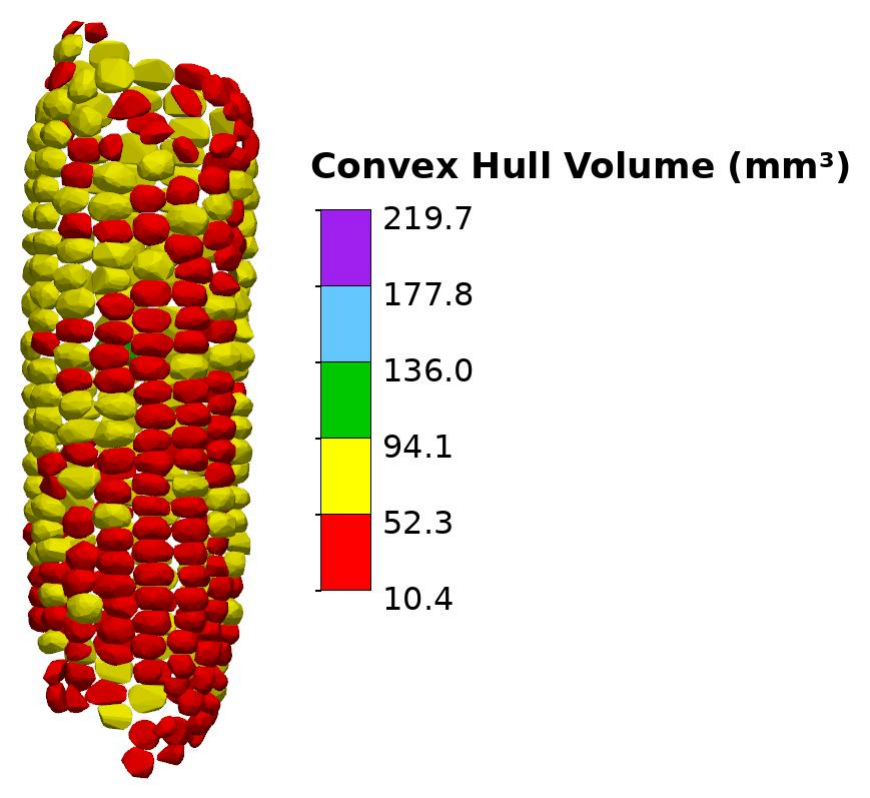}
    \caption{24-P2071\_2}
\end{subfigure}
\begin{subfigure}[b]{0.24\textwidth}
    \centering
    \includegraphics[width=\textwidth]{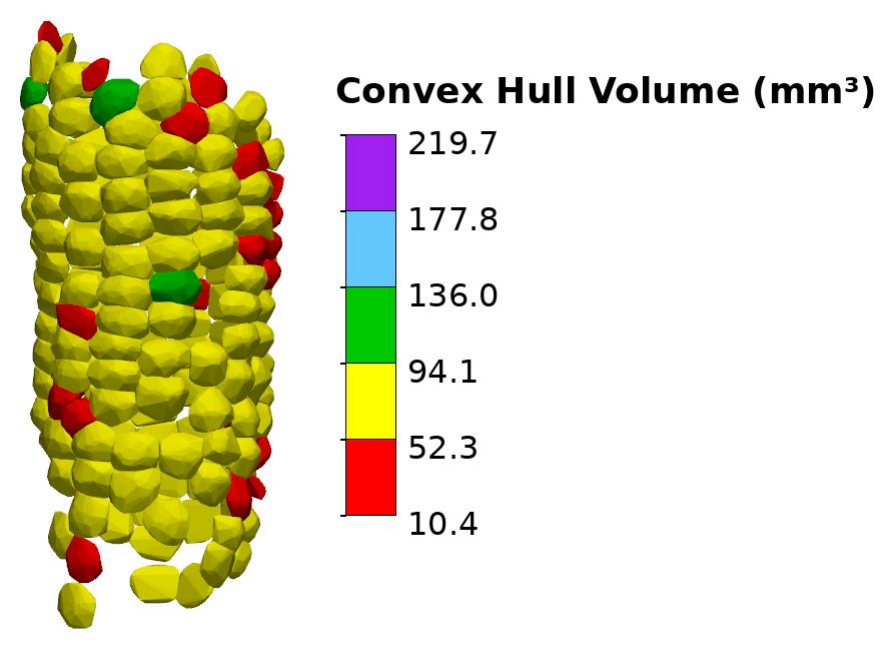}
    \caption{24-P2146\_2}
\end{subfigure}\hfill
\begin{subfigure}[b]{0.24\textwidth}
    \centering
    \includegraphics[width=\textwidth]{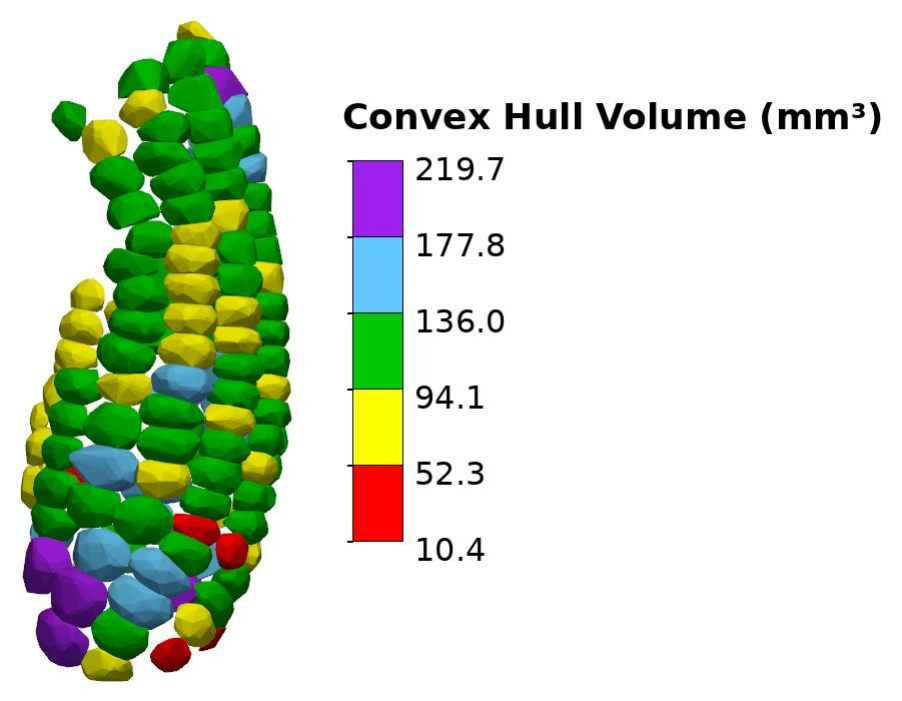}
    \caption{24-P2199\_1}
\end{subfigure}

\caption{Per-kernel convex hull volume proxy mapped onto the 3D ear point cloud for each representative ear. Color encodes convex hull volume (mm\textsuperscript{3}).}
\label{fig:volume_3d}
\end{figure}

\begin{figure}[t!]
\centering
\begin{subfigure}[b]{0.24\textwidth}
    \centering
    \includegraphics[width=\textwidth]{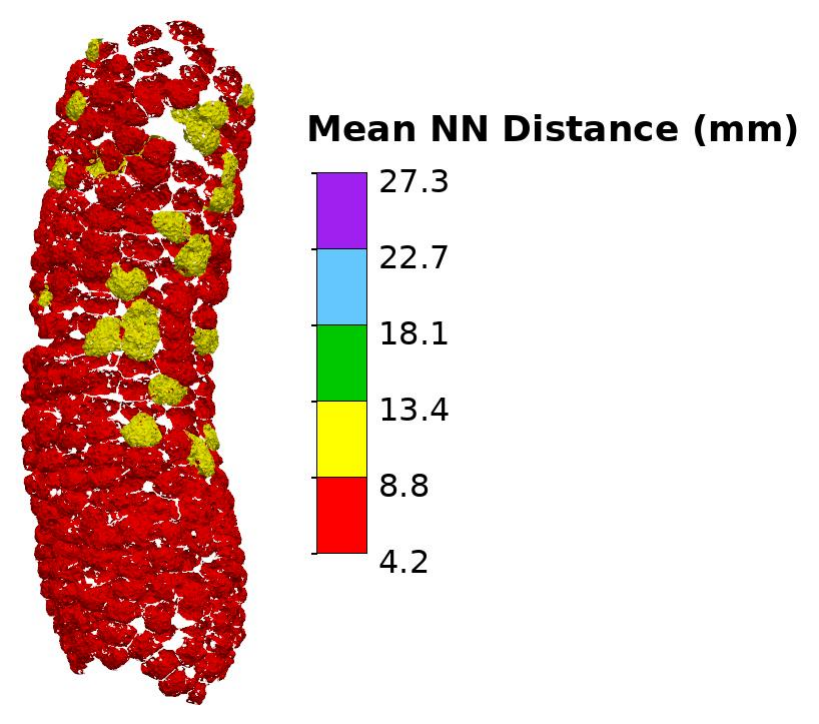}
    \caption{24-P2042\_3}
\end{subfigure}\hfill
\begin{subfigure}[b]{0.24\textwidth}
    \centering
    \includegraphics[width=\textwidth]{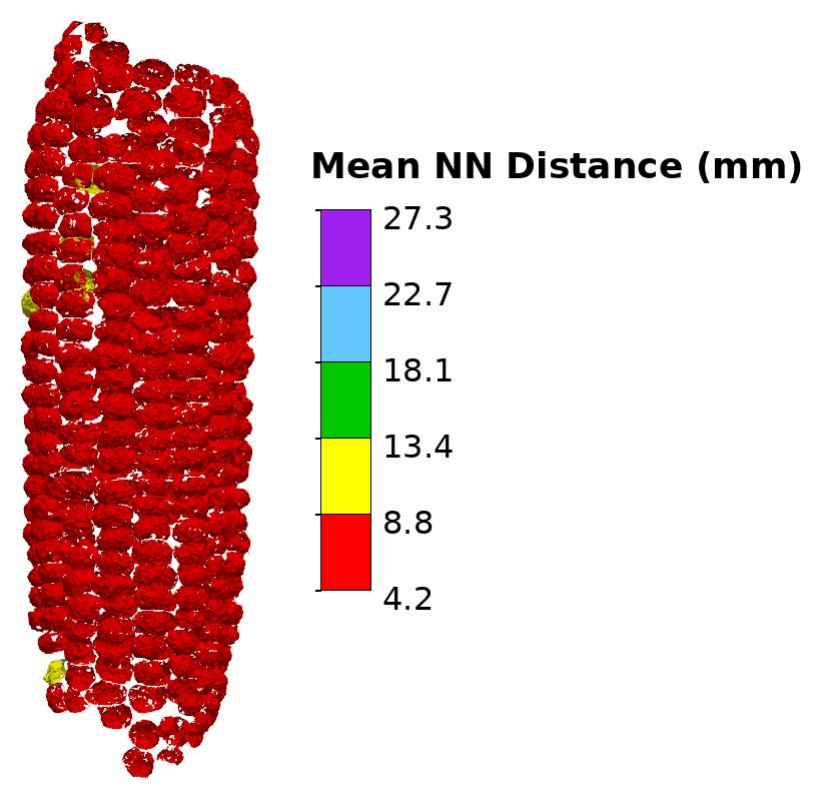}
    \caption{24-P2071\_2}
\end{subfigure}
\begin{subfigure}[b]{0.24\textwidth}
    \centering
    \includegraphics[width=\textwidth]{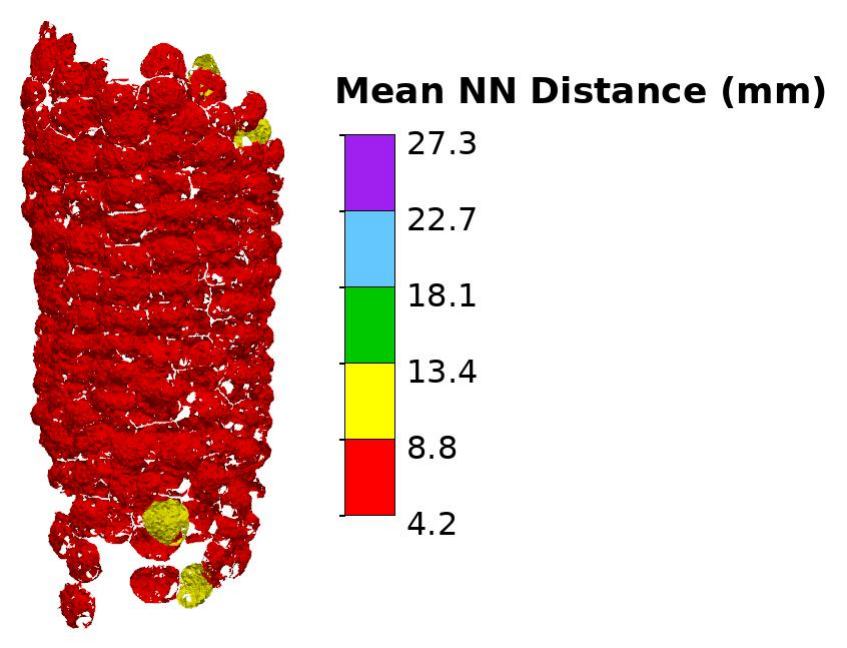}
    \caption{24-P2146\_2}
\end{subfigure}\hfill
\begin{subfigure}[b]{0.24\textwidth}
    \centering
    \includegraphics[width=\textwidth]{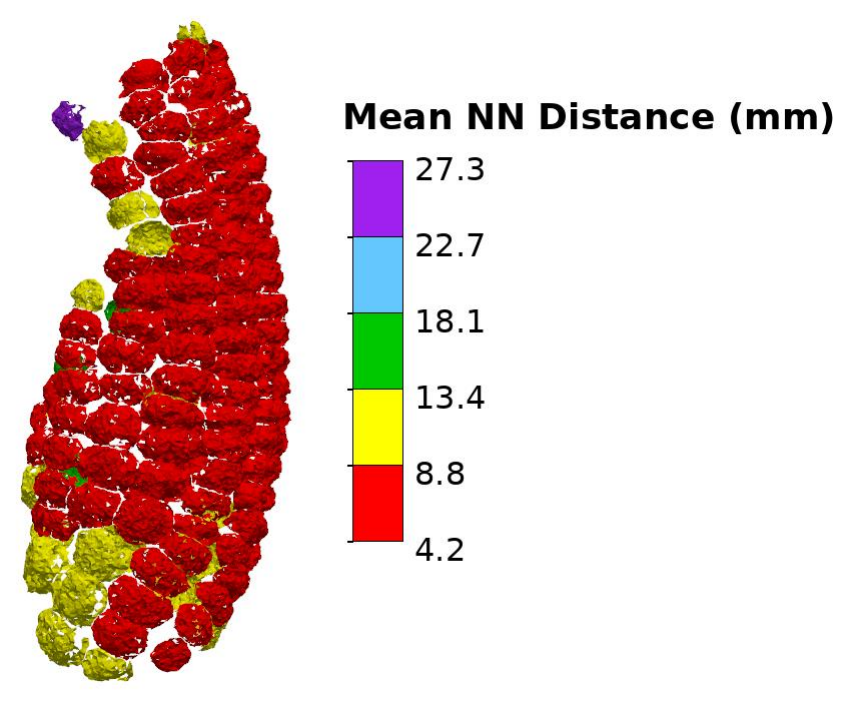}
    \caption{24-P2199\_1}
\end{subfigure}
\caption{Mean nearest-neighbor distance mapped onto the 3D ear point cloud for each representative ear. Color encodes mean 3D Gabriel graph neighbor distance (mm).}
\label{fig:nndist_3d}
\end{figure}

\begin{figure}[t!]
\centering
\begin{subfigure}[b]{0.24\textwidth}
    \includegraphics[width=\textwidth]{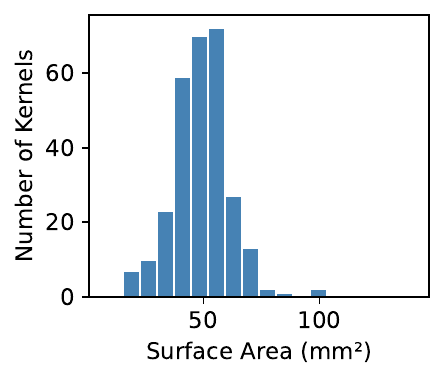}
    \caption{24-P2042\_3}
\end{subfigure}\hfill
\begin{subfigure}[b]{0.24\textwidth}
    \includegraphics[width=\textwidth]{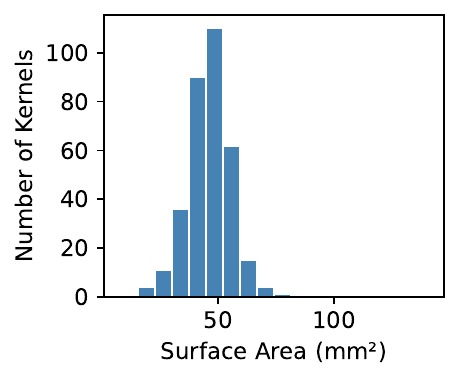}
    \caption{24-P2071\_2}
\end{subfigure}
\begin{subfigure}[b]{0.24\textwidth}
    \includegraphics[width=\textwidth]{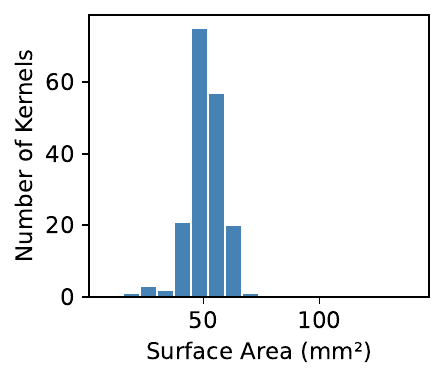}
    \caption{24-P2146\_2}
\end{subfigure}\hfill
\begin{subfigure}[b]{0.24\textwidth}
    \includegraphics[width=\textwidth]{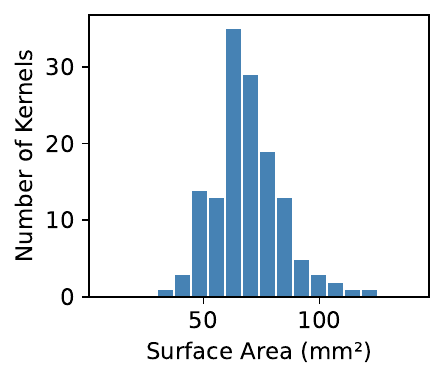}
    \caption{24-P2199\_1}
\end{subfigure}
\caption{Per-kernel 3D surface area, estimated via the Ball-Pivoting Algorithm (BPA).}
\label{fig:surface_area}
\end{figure}

\begin{figure}[t!]
\centering
\begin{subfigure}[b]{0.24\textwidth}
    \includegraphics[width=\textwidth]{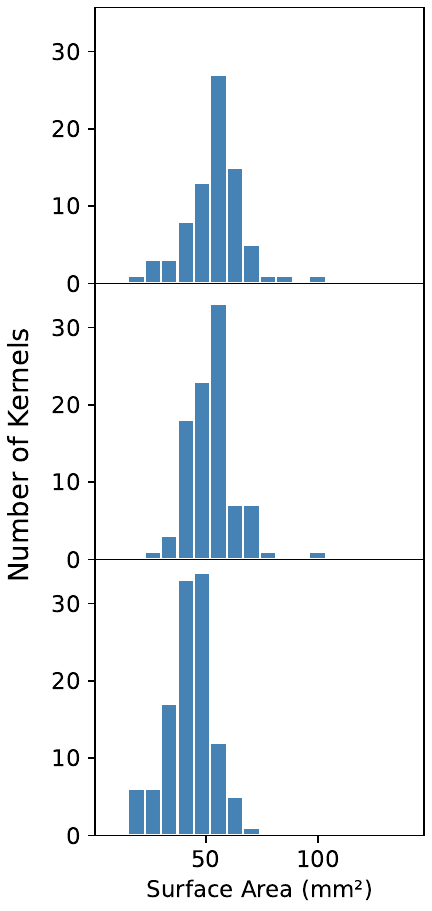}
    \caption{24-P2042\_3}
\end{subfigure}
\begin{subfigure}[b]{0.24\textwidth}
    \includegraphics[width=\textwidth]{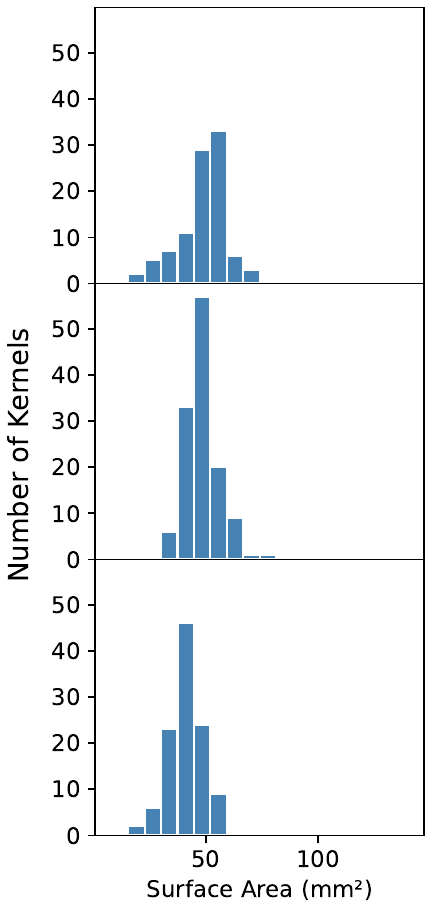}
    \caption{24-P2071\_2}
\end{subfigure}
\begin{subfigure}[b]{0.24\textwidth}
    \includegraphics[width=\textwidth]{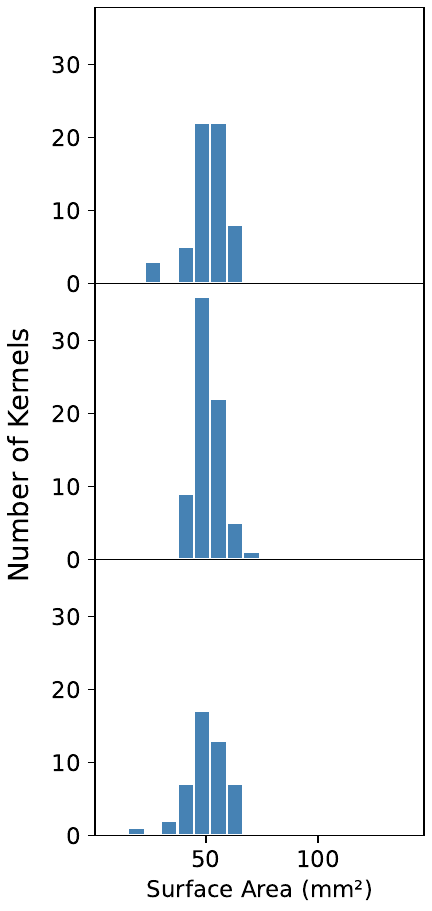}
    \caption{24-P2146\_2}
\end{subfigure}
\begin{subfigure}[b]{0.24\textwidth}
    \includegraphics[width=\textwidth]{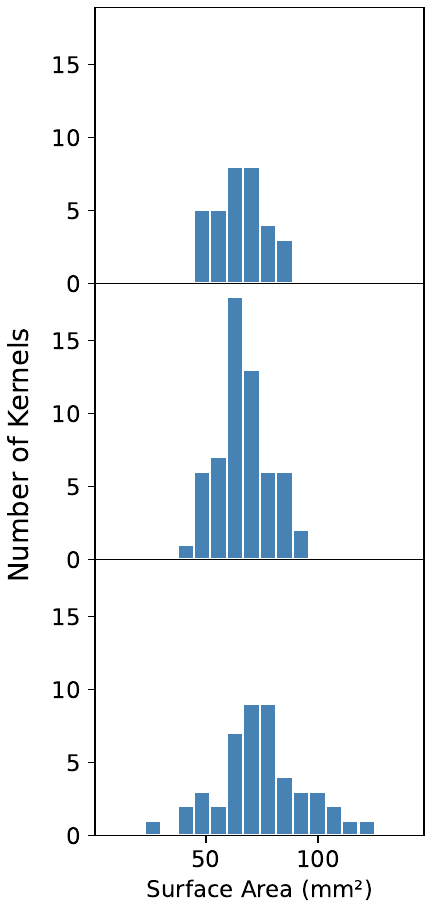}
    \caption{24-P2199\_1}
\end{subfigure}
\caption{Per-kernel surface area distributions stratified by ear third, from the base (bottom row) to the tip (top row).}
\label{fig:surface_area_thirds}
\end{figure}

\begin{figure}[t!]
\centering
\begin{subfigure}[b]{0.24\textwidth}
    \includegraphics[width=\textwidth]{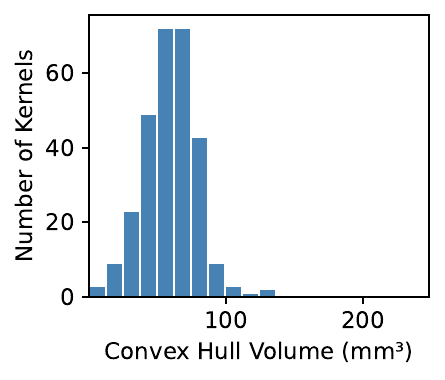}
    \caption{24-P2042\_3}
\end{subfigure}
\begin{subfigure}[b]{0.24\textwidth}
    \includegraphics[width=\textwidth]{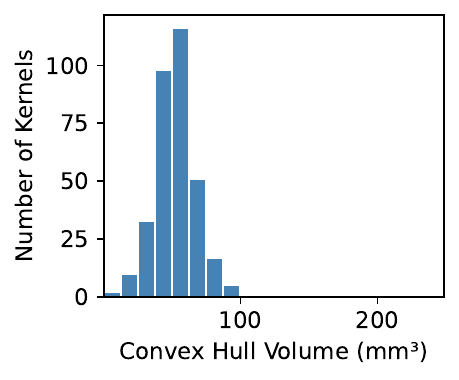}
    \caption{24-P2071\_2}
\end{subfigure}
\begin{subfigure}[b]{0.24\textwidth}
    \includegraphics[width=\textwidth]{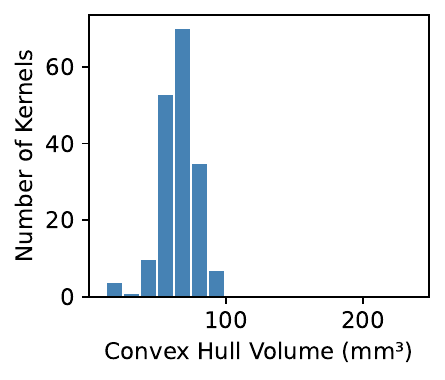}
    \caption{24-P2146\_2}
\end{subfigure}
\begin{subfigure}[b]{0.24\textwidth}
    \includegraphics[width=\textwidth]{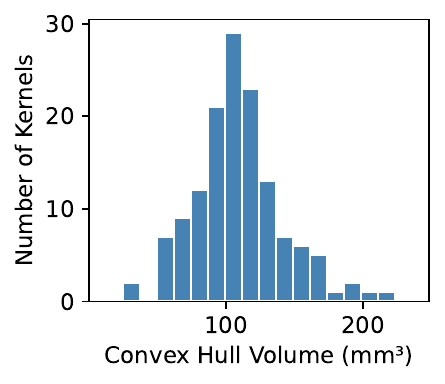}
    \caption{24-P2199\_1}
\end{subfigure}
\caption{Per-kernel convex hull volume proxy distributions.}
\label{fig:volume}
\end{figure}

\begin{figure}[t!]
\centering
\begin{subfigure}[b]{0.24\textwidth}
    \includegraphics[width=\textwidth]{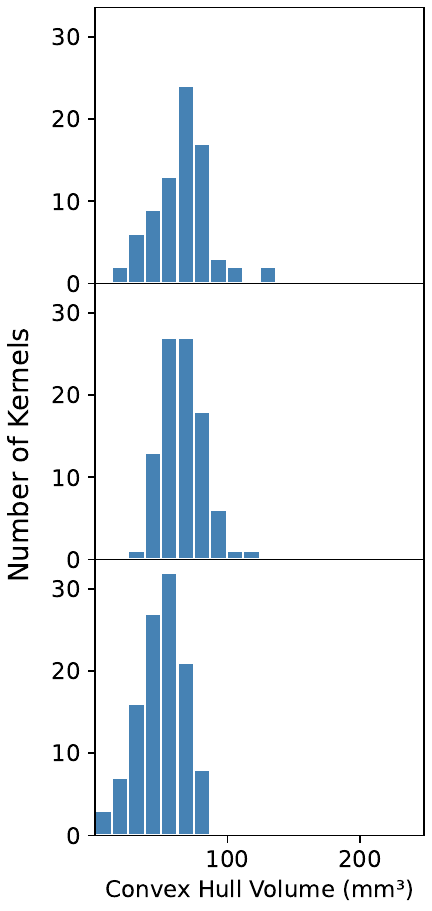}
    \caption{24-P2042\_3}
\end{subfigure}
\begin{subfigure}[b]{0.24\textwidth}
    \includegraphics[width=\textwidth]{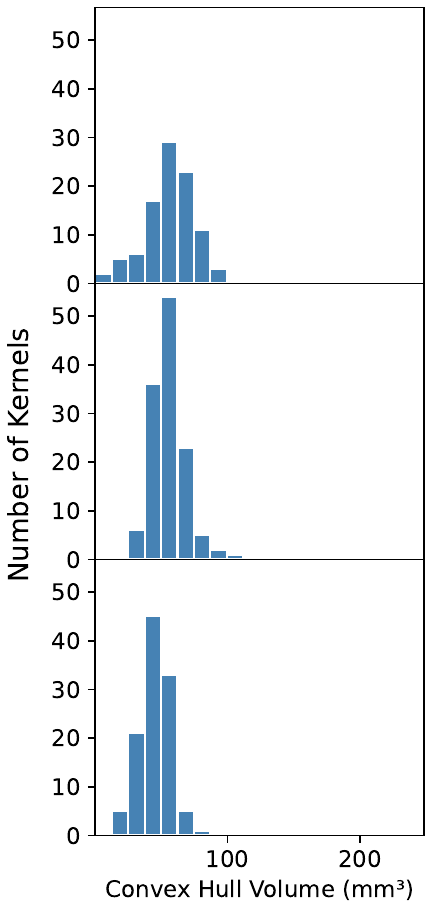}
    \caption{24-P2071\_2}
\end{subfigure}
\begin{subfigure}[b]{0.24\textwidth}
    \includegraphics[width=\textwidth]{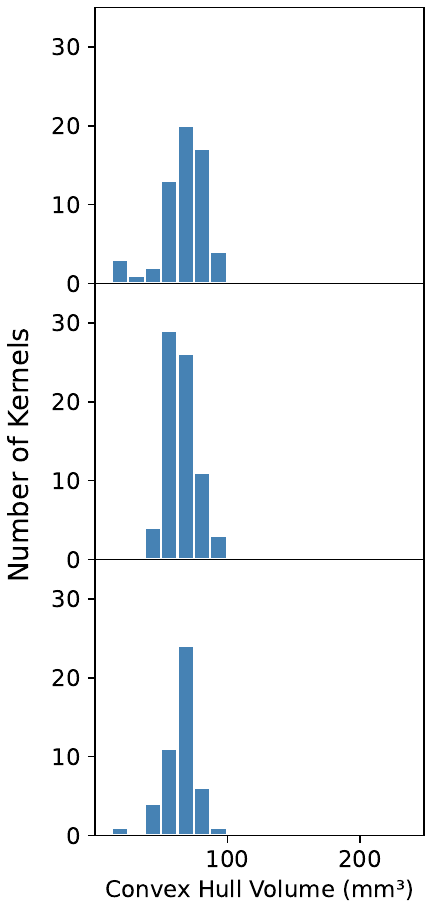}
    \caption{24-P2146\_2}
\end{subfigure}
\begin{subfigure}[b]{0.24\textwidth}
    \includegraphics[width=\textwidth]{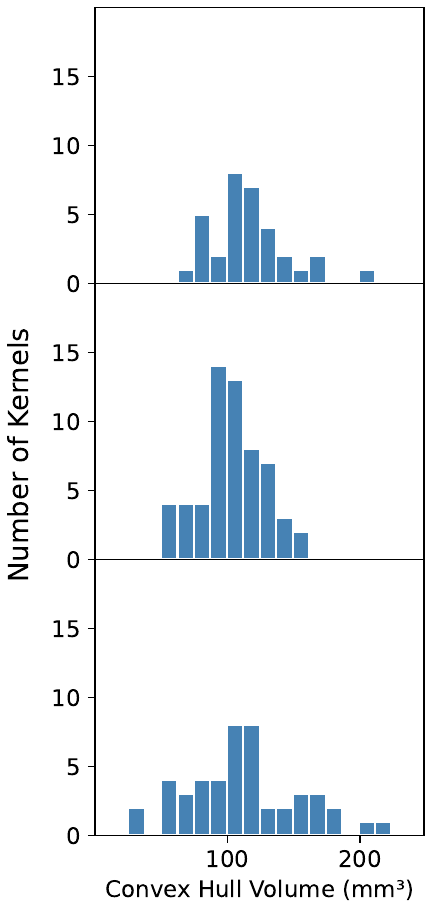}
    \caption{24-P2199\_1}
\end{subfigure}
\caption{Per-kernel convex hull volume proxy distributions stratified by ear third, from the base (bottom row) to the tip (top row).}
\label{fig:volume_thirds}
\end{figure}

\begin{figure}[t!]
\centering
\begin{subfigure}[b]{0.24\textwidth}
    \includegraphics[width=\textwidth]{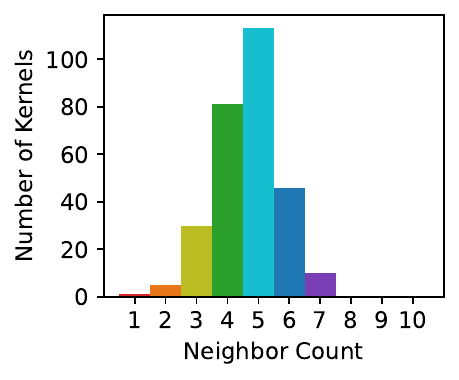}
    \caption{24-P2042\_3}
\end{subfigure}\hfill
\begin{subfigure}[b]{0.24\textwidth}
    \includegraphics[width=\textwidth]{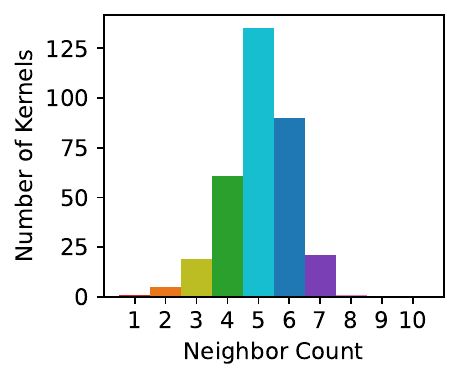}
    \caption{24-P2071\_2}
\end{subfigure}
\begin{subfigure}[b]{0.24\textwidth}
    \includegraphics[width=\textwidth]{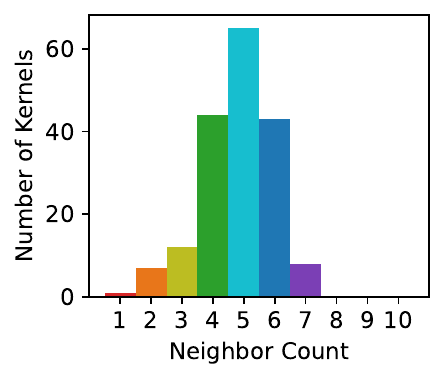}
    \caption{24-P2146\_2}
\end{subfigure}\hfill
\begin{subfigure}[b]{0.24\textwidth}
    \includegraphics[width=\textwidth]{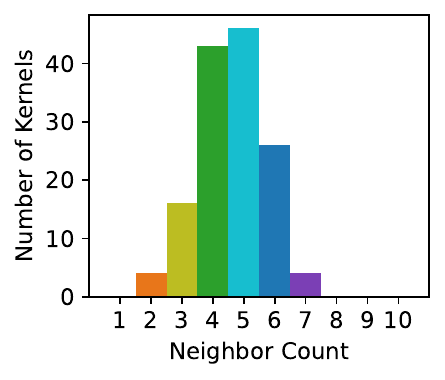}
    \caption{24-P2199\_1}
\end{subfigure}
\caption{Gabriel graph neighbor count distributions.}
\label{fig:neighbor_dist}
\end{figure}

\begin{figure}[t!]
\centering
\begin{subfigure}[b]{0.24\textwidth}
    \includegraphics[width=\textwidth]{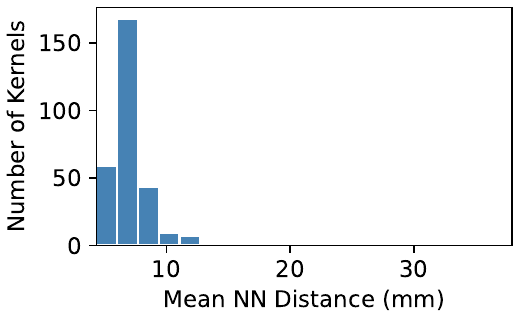}
    \caption{24-P2042\_3}
\end{subfigure}\hfill
\begin{subfigure}[b]{0.24\textwidth}
    \includegraphics[width=\textwidth]{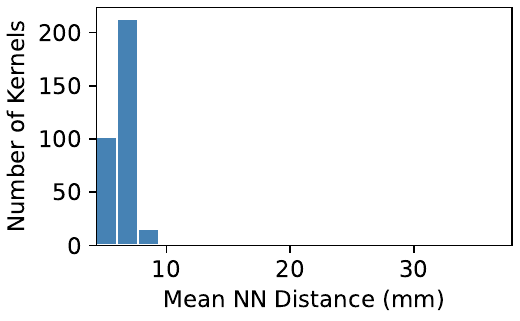}
    \caption{24-P2071\_2}
\end{subfigure}
\begin{subfigure}[b]{0.24\textwidth}
    \includegraphics[width=\textwidth]{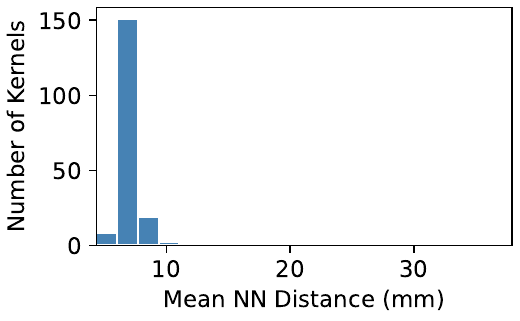}
    \caption{24-P2146\_2}
\end{subfigure}\hfill
\begin{subfigure}[b]{0.24\textwidth}
    \includegraphics[width=\textwidth]{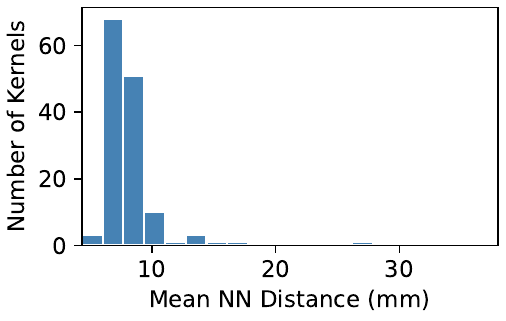}
    \caption{24-P2199\_1}
\end{subfigure}
\caption{Distributions of 3D mean nearest-neighbor distance to each Gabriel graph neighbor.}
\label{fig:mean_distance}
\end{figure}

\begin{figure}[p]
\centering
\begin{subfigure}[b]{0.49\textwidth}
    \includegraphics[width=\textwidth, keepaspectratio=true]{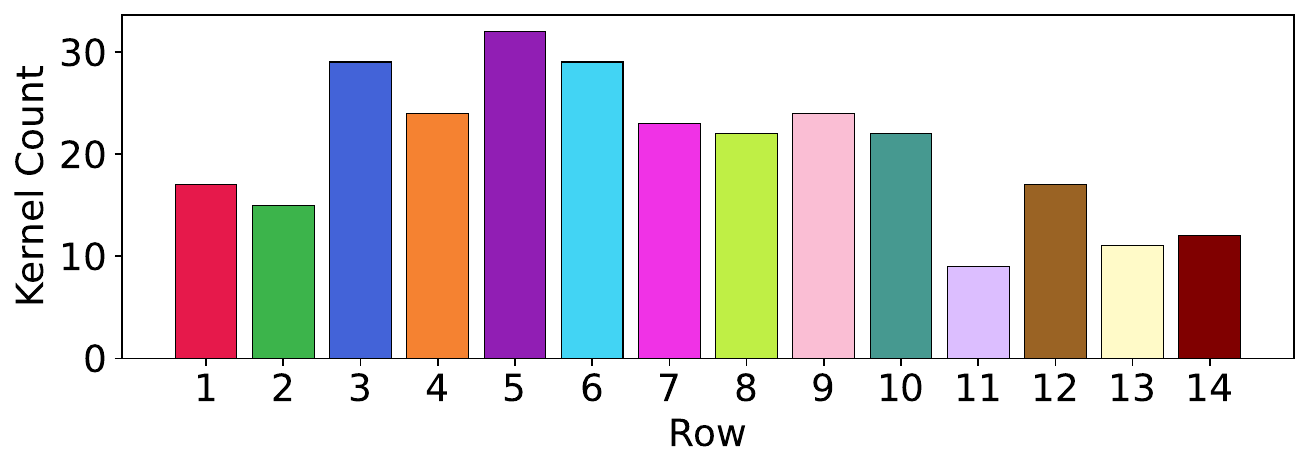}
    \caption{24-P2042\_3}
\end{subfigure}
\begin{subfigure}[b]{0.49\textwidth}
    \includegraphics[width=\textwidth, keepaspectratio=true]{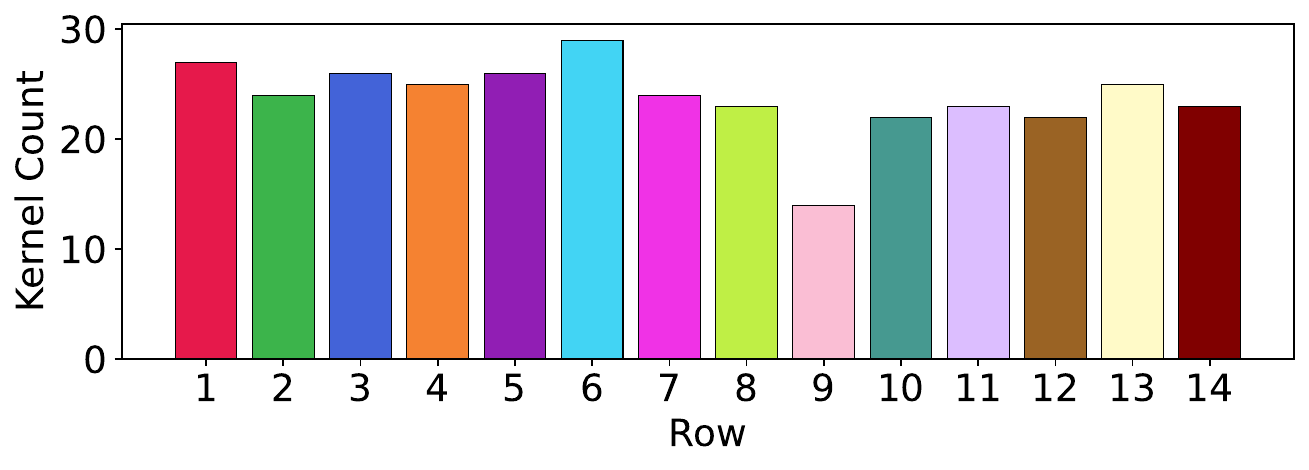}
    \caption{24-P2071\_2}
\end{subfigure}
\begin{subfigure}[b]{0.49\textwidth}
    \includegraphics[width=\textwidth, keepaspectratio=true]{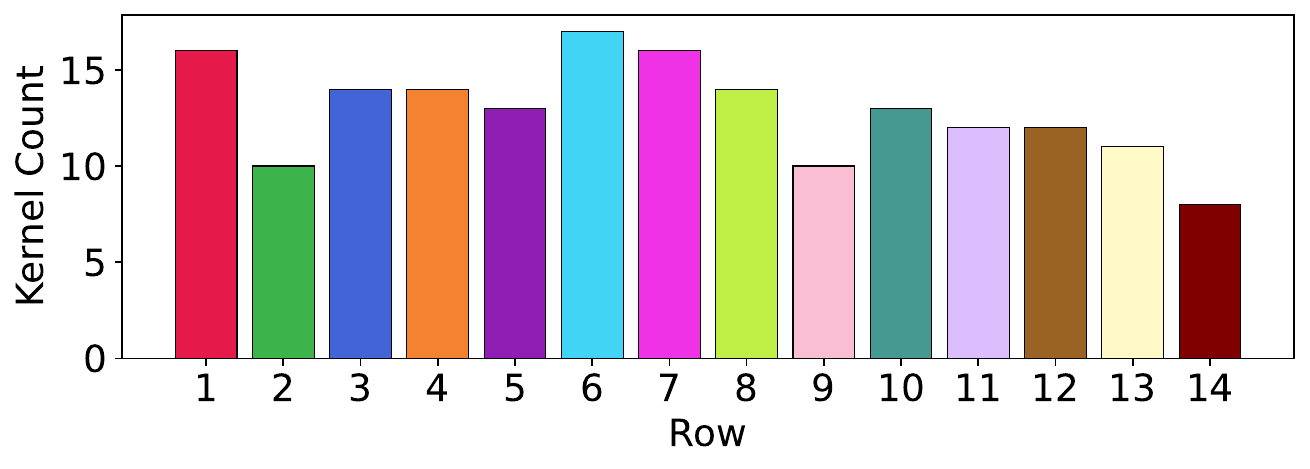}
    \caption{24-P2146\_2}
\end{subfigure}
\begin{subfigure}[b]{0.49\textwidth}
    \includegraphics[width=\textwidth, keepaspectratio=true]{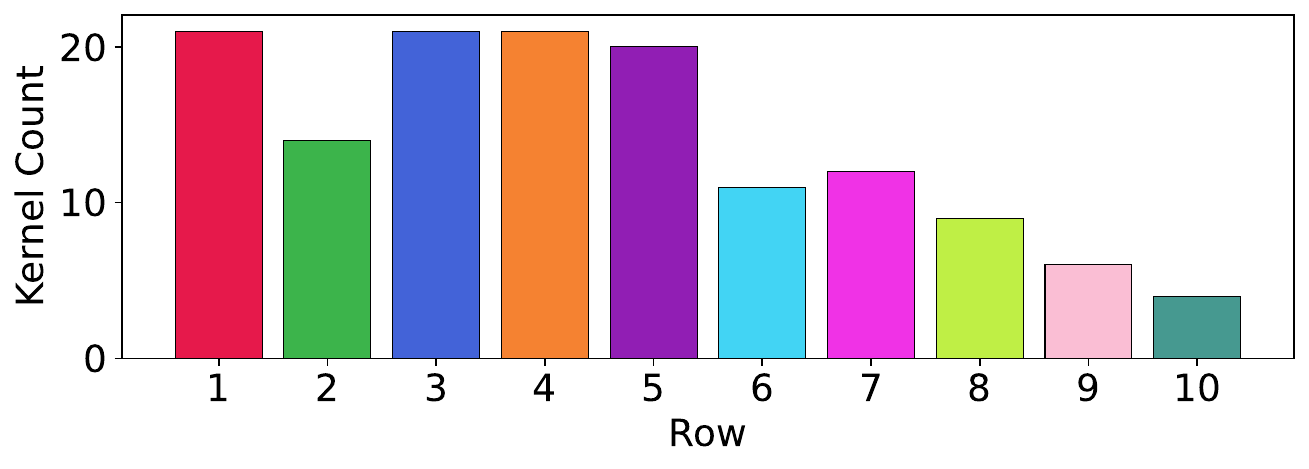}
    \caption{24-P2199\_1}
\end{subfigure}
\caption{Per-row kernel count distributions, derived from the row chain graph.}
\label{fig:row_distribution}
\end{figure}

\begin{figure}[t!]
\centering
\begin{subfigure}[b]{0.45\textwidth}
    \includegraphics[width=\textwidth]{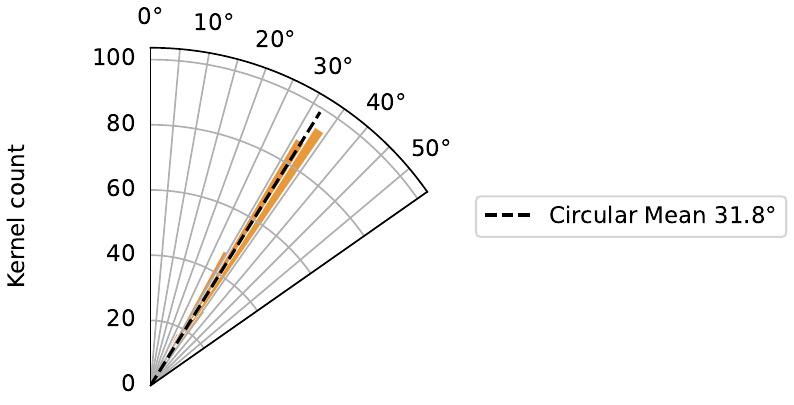}
    \caption{24-P2042\_3}
\end{subfigure}\hfill
\begin{subfigure}[b]{0.45\textwidth}
    \includegraphics[width=\textwidth]{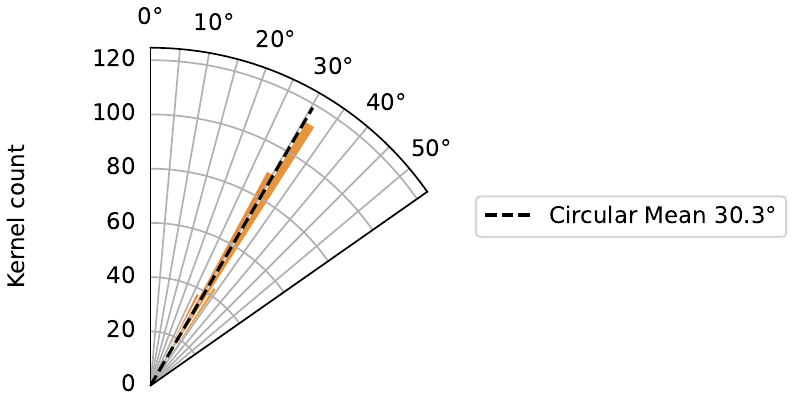}
    \caption{24-P2071\_2}
\end{subfigure}

\vspace{0.5em}

\begin{subfigure}[b]{0.45\textwidth}
    \includegraphics[width=\textwidth]{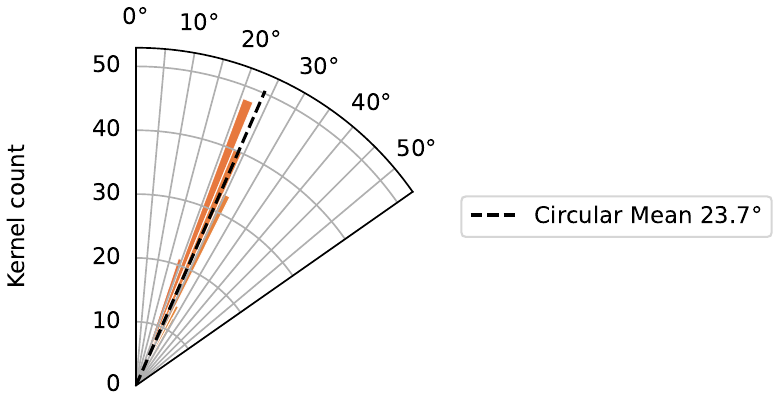}
    \caption{24-P2146\_2}
\end{subfigure}\hfill
\begin{subfigure}[b]{0.45\textwidth}
    \includegraphics[width=\textwidth]{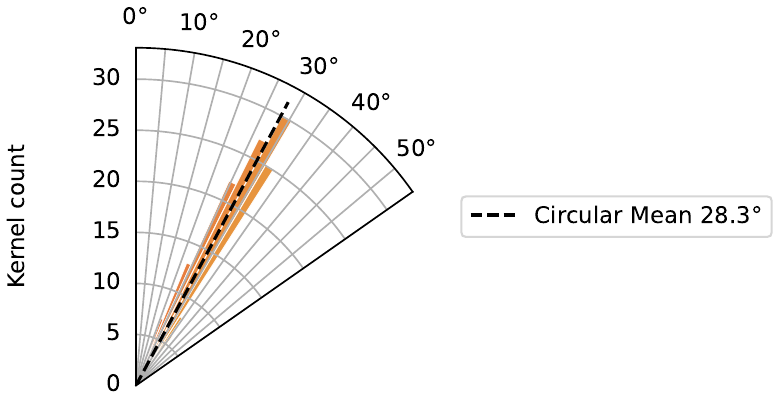}
    \caption{24-P2199\_1}
\end{subfigure}
\caption{Per-kernel circular mean hue distributions for the four representative ears. Each bar is colored to match the kernel hue angle it represents.}
\label{fig:hue}
\end{figure}

\begin{figure}[t!]
\centering
\begin{subfigure}[b]{0.24\textwidth}
    \includegraphics[width=\textwidth]{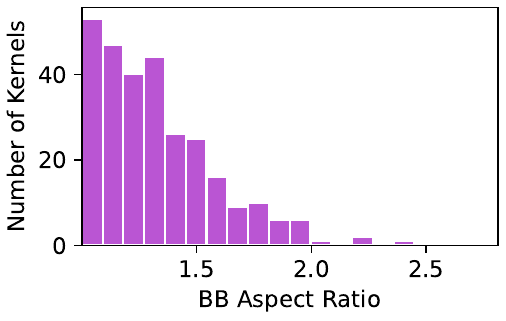}
    \caption{24-P2042\_3}
\end{subfigure}\hfill
\begin{subfigure}[b]{0.24\textwidth}
    \includegraphics[width=\textwidth]{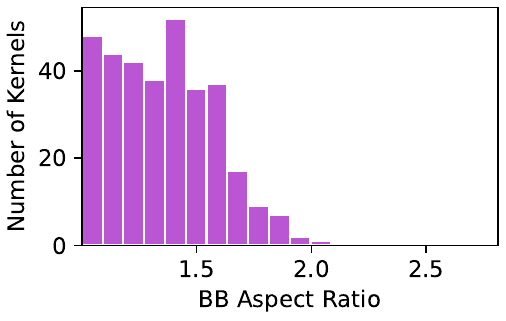}
    \caption{24-P2071\_2}
\end{subfigure}
\begin{subfigure}[b]{0.24\textwidth}
    \includegraphics[width=\textwidth]{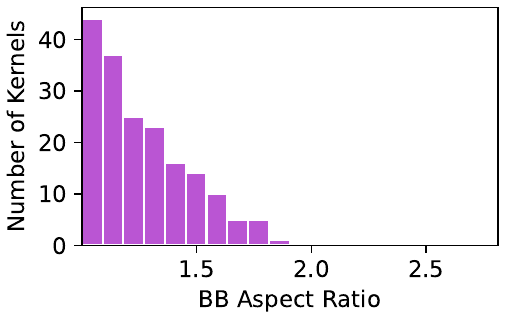}
    \caption{24-P2146\_2}
\end{subfigure}\hfill
\begin{subfigure}[b]{0.24\textwidth}
    \includegraphics[width=\textwidth]{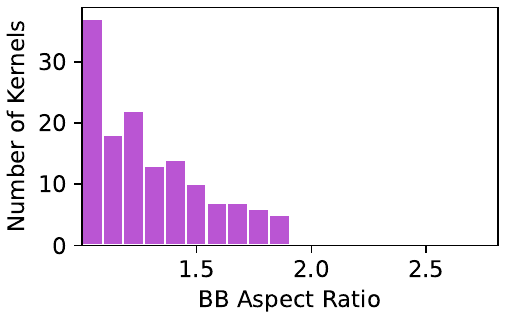}
    \caption{24-P2199\_1}
\end{subfigure}
\caption{Per-kernel 3D bounding box aspect ratio distributions.}
\label{fig:aspect_ratio}
\end{figure}

\begin{figure}[t!]
\centering
\begin{subfigure}[b]{0.24\textwidth}
    \includegraphics[width=\textwidth]{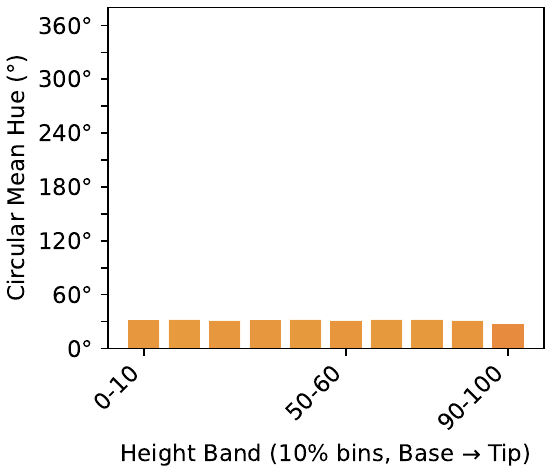}
    \caption{24-P2042\_3}
\end{subfigure}
\begin{subfigure}[b]{0.24\textwidth}
    \includegraphics[width=\textwidth]{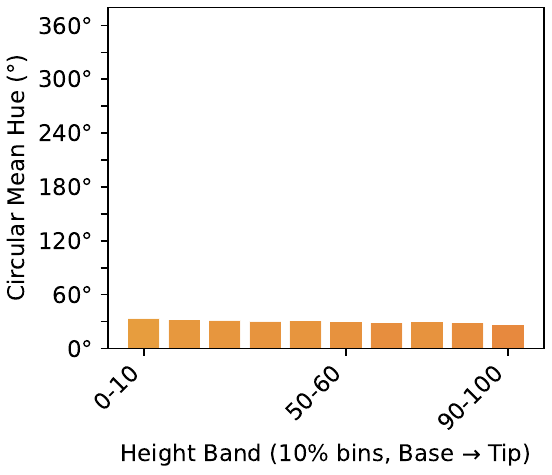}
    \caption{24-P2071\_2}
\end{subfigure}
\begin{subfigure}[b]{0.24\textwidth}
    \includegraphics[width=\textwidth]{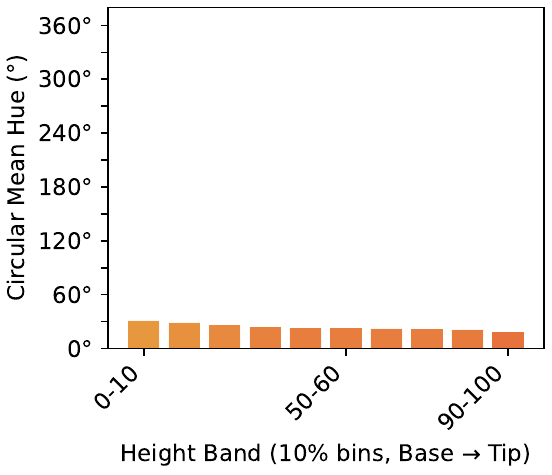}
    \caption{24-P2146\_2}
\end{subfigure}
\begin{subfigure}[b]{0.24\textwidth}
    \includegraphics[width=\textwidth]{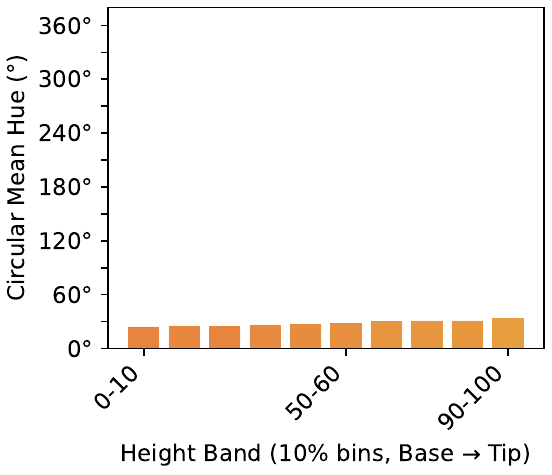}
    \caption{24-P2199\_1}
\end{subfigure}
\caption{Circular mean hue per 10\% height band (base to tip) for the four representative ears. Each bar is colored to match the kernel hue angle it represents.}
\label{fig:hue_by_height}
\end{figure}

\begin{figure}[t!]
\centering
\begin{subfigure}[b]{0.24\textwidth}
    \includegraphics[width=\textwidth]{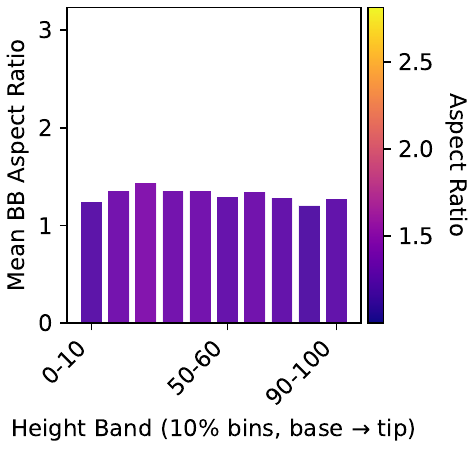}
    \caption{24-P2042\_3}
\end{subfigure}
\begin{subfigure}[b]{0.24\textwidth}
    \includegraphics[width=\textwidth]{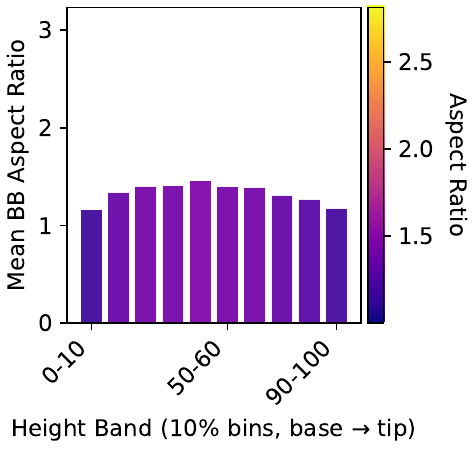}
    \caption{24-P2071\_2}
\end{subfigure}
\begin{subfigure}[b]{0.24\textwidth}
    \includegraphics[width=\textwidth]{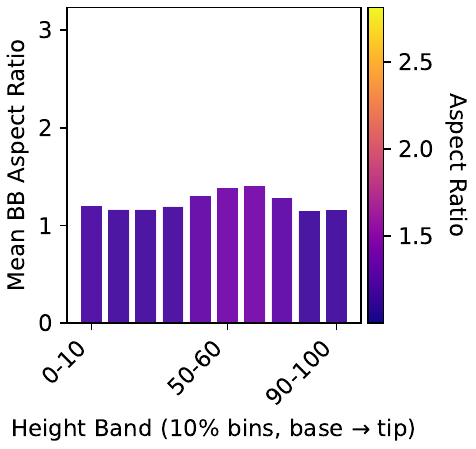}
    \caption{24-P2146\_2}
\end{subfigure}
\begin{subfigure}[b]{0.24\textwidth}
    \includegraphics[width=\textwidth]{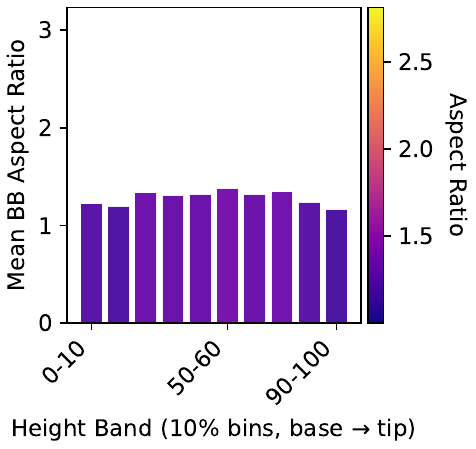}
    \caption{24-P2199\_1}
\end{subfigure}
\caption{Mean bounding-box aspect ratio per 10\% height band (base to tip).}
\label{fig:ar_by_height}
\end{figure}

\begin{figure}[p]
\centering
\begin{subfigure}{0.49\textwidth}
    \centering
    \includegraphics[width=\textwidth]{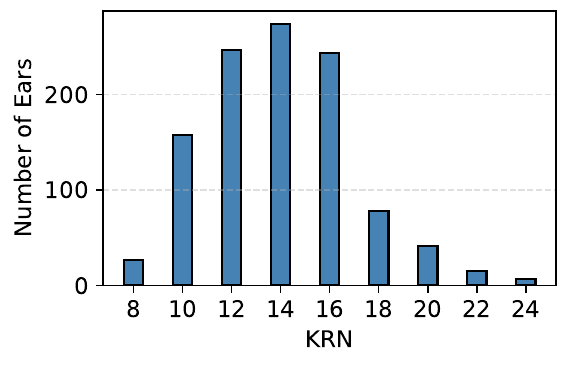}
    \caption{\footnotesize KRN distribution.}
    \label{fig:combined_krn}
\end{subfigure}\hfill
\begin{subfigure}{0.49\textwidth}
    \centering
    \includegraphics[width=\textwidth]{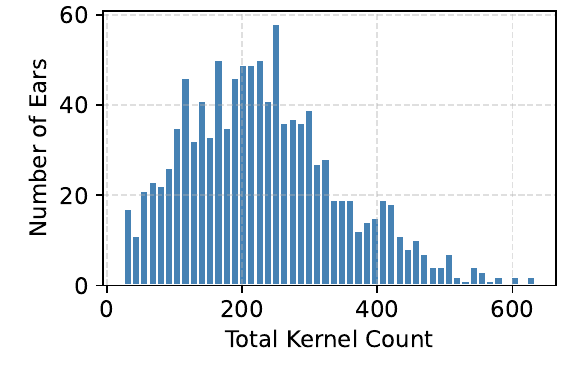}
    \caption{\footnotesize Kernel count distribution.}
    \label{fig:combined_kcount}
\end{subfigure}
\begin{subfigure}{0.49\textwidth}
    \centering
    \includegraphics[width=\textwidth]{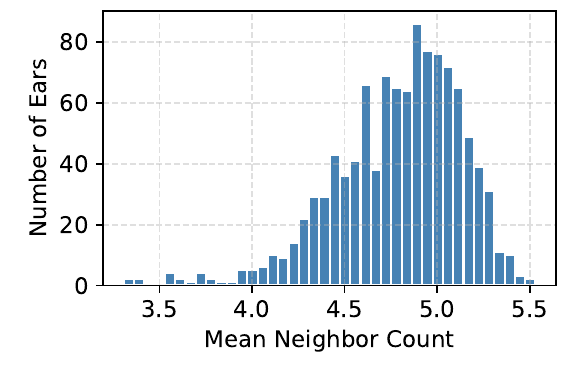}
    \caption{\footnotesize Mean Gabriel graph neighbor count distribution.}
    \label{fig:combined_neighbors}
\end{subfigure}\hfill
\begin{subfigure}{0.49\textwidth}
    \centering
    \includegraphics[width=\textwidth]{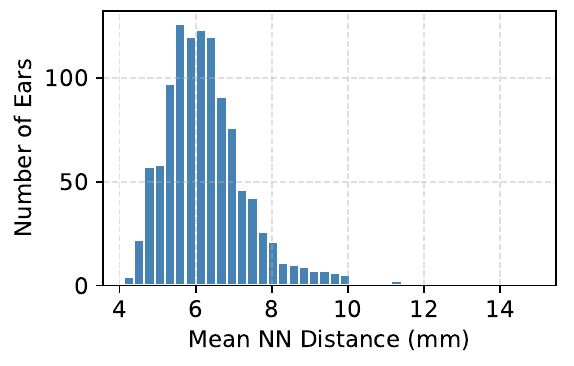}
    \caption{\footnotesize Mean 3D nearest-neighbor distance distribution.}
    \label{fig:combined_nndist}
\end{subfigure}
\begin{subfigure}{0.49\textwidth}
    \centering
    \includegraphics[width=\textwidth]{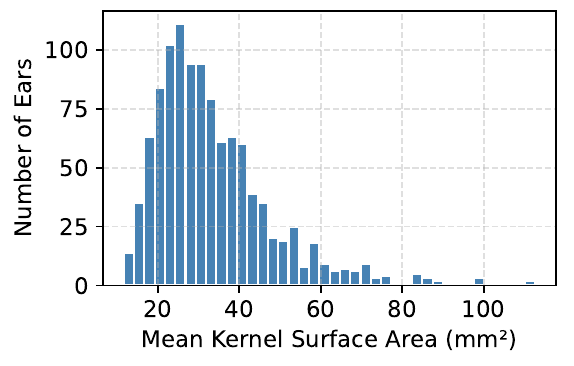}
    \caption{\footnotesize Mean kernel 3D surface area distribution.}
    \label{fig:combined_sa}
\end{subfigure}\hfill
\begin{subfigure}{0.49\textwidth}
    \centering
    \includegraphics[width=\textwidth]{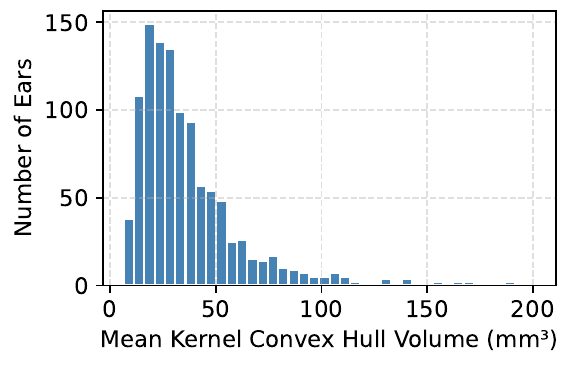}
    \caption{\footnotesize Mean kernel convex hull volume proxy distribution.}
    \label{fig:combined_volume}
\end{subfigure}
\caption{Trait distributions across all $1{,}091$ ears.}
\label{fig:combined_distributions}
\end{figure}

Representative per-ear trait distributions for the four ears detailed in the Methods are presented in Figures~\ref{fig:area_3d}--\ref{fig:ar_by_height}. These include spatially mapped per-kernel surface area on the 3D ear (\figref{fig:area_3d}), per-kernel convex hull volume proxy on the 3D ear (\figref{fig:volume_3d}), mean nearest-neighbor distance on the 3D ear (\figref{fig:nndist_3d}), per-kernel surface area histogram (\figref{fig:surface_area}), surface area stratified by ear third (\figref{fig:surface_area_thirds}), convex hull volume proxy (\figref{fig:volume}), convex hull volume proxy stratified by ear third (\figref{fig:volume_thirds}), Gabriel graph neighbor count (\figref{fig:neighbor_dist}), mean nearest-neighbor distance (\figref{fig:mean_distance}), per-row kernel count (\figref{fig:row_distribution}), circular mean hue (\figref{fig:hue}), bounding box aspect ratio (\figref{fig:aspect_ratio}), mean hue vs.\ height (\figref{fig:hue_by_height}), and mean aspect ratio vs.\ height (\figref{fig:ar_by_height}). Trait distributions over all ears are shown in \figref{fig:combined_distributions}.

% =====================================================================
% FIGURE: 3D Surface Area maps
% =====================================================================

% =====================================================================
% FIGURE: 3D Convex Hull Volume maps
% =====================================================================

% =====================================================================
% FIGURE: 3D Mean NN Distance maps
% =====================================================================

% ------ Surface area histograms ------

% ------ Surface area by thirds ------

% ------ Volume histograms ------

% ------ Volume by thirds ------

% ------ Neighbor count distributions ------

% ------ Mean neighbor distance histograms ------

% ------ Row distributions ------

% ------ Hue histograms ------

% ------ Aspect ratio histograms ------

% ------ Hue by height ------

% ------ Aspect ratio by height ------

% ------ Combined distributions ------

\FloatBarrier

\subsection{Manually Annotated Subset Validation}
\label{sec:manual_validation}

\begin{figure}[h!]
\centering
    \begin{subfigure}[b]{0.45\textwidth}
        \includegraphics[width=0.99\linewidth]{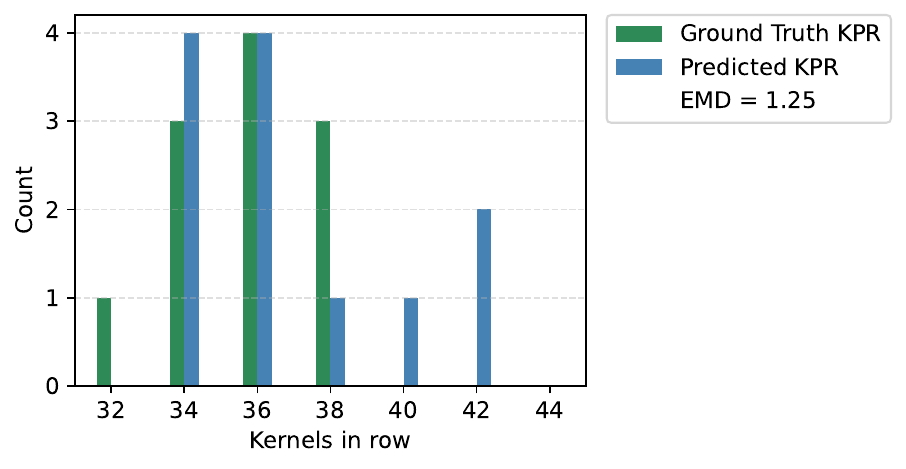}
        \caption{KPR distribution}
        \label{fig:sample1_kpr}
    \end{subfigure}
    \begin{subfigure}[b]{0.54\textwidth}
        \includegraphics[width=0.99\linewidth]{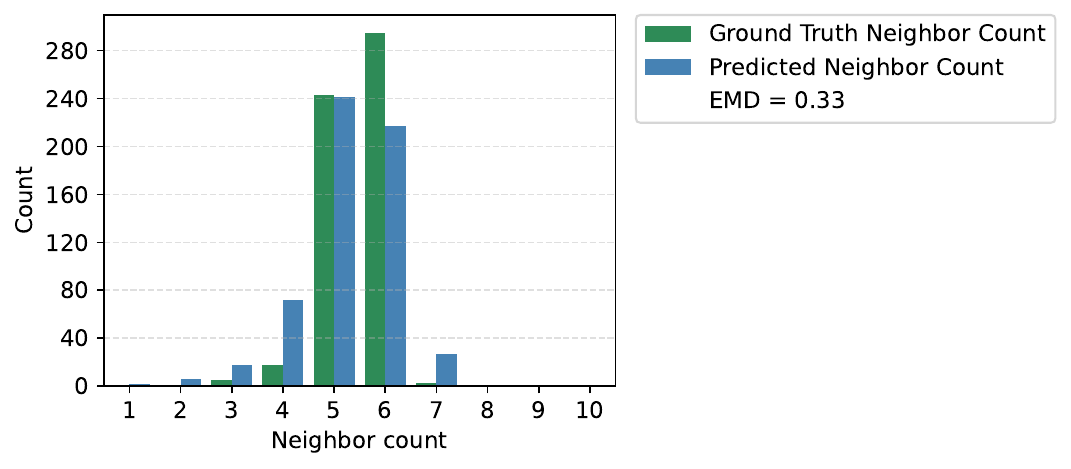}
        \caption{Neighbor count distribution}
        \label{fig:sample1_neighbor}
    \end{subfigure}
\caption{Representative kernels-per-row (KPR) and Gabriel graph neighbor-count accuracy for one ear (Sample 1) from the six-ear manually annotated subset.}
\label{fig:sample1_accuracy}
\end{figure}

Spatial arrangement and packing traits require dense per-kernel visual tracking, which is prohibitively labor-intensive to perform manually at scale. An exhaustive manual validation was therefore performed on the six-ear manually annotated subset, drawn from an independent field study with no overlap with the 1{,}091-ear reconstructed population or the 268-ear labeled dataset (\secref{sec:dataset_composition}), to verify spatial arrangement and packing accuracy. The pipeline's tracking fidelity on this subset is evaluated through kernels-per-row (KPR) and Gabriel graph neighbor-count (packing) distributions, quantified by the Earth-Mover Distance (\secref{sec:emd}). A representative example (Sample 1) is shown in \figref{fig:sample1_accuracy}. The predicted and ground-truth KRN for all six samples are summarized in \tableref{tab:krn_accuracy}, and the per-ear EMD values for both traits in \tableref{tab:emd_summary}; the remaining individual distribution plots (Samples 2--6) are provided in the \hyperref[sec:supplemental]{Supplemental Material}.

Gabriel graph neighbor-count agreement is uniformly close, with per-ear EMD between 0.33 and 0.84 (mean 0.57), every ear falling below the single-count threshold defined in \secref{sec:emd}. Kernels-per-row agreement is less consistent, with per-ear EMD between 0.75 and 7.58 (mean 3.04), concentrated in Samples 4 and 6. These are the same two ears for which KRN was underpredicted (\tableref{tab:krn_accuracy}); the mechanism linking the two is examined in \secref{sec:discussion}.

% ------ Sample 1 Distributions ------

% ------ KRN Table ------
\begin{table}[h!]
\centering
\caption{Predicted vs.\ ground-truth KRN per ear for the manually validated subset.}
\label{tab:krn_accuracy}
\begin{tabular}{ccc}
    \toprule
    Sample & Predicted KRN & GT KRN \\
    \midrule
    1 & 16 & 16 \\
    2 & 16 & 16 \\
    3 & 14 & 14 \\
    4 & 16 & 18 \\
    5 & 16 & 16 \\
    6 & 16 & 18 \\
    \bottomrule
\end{tabular}
\end{table}

% ------ EMD Table ------
\begin{table}[t!]
\centering
\caption{Earth-Mover Distance between automated and manually annotated distributions, per ear, for the six-ear manually annotated subset. Values are taken from \figref{fig:sample1_accuracy} (Sample 1) and \figref{fig:appendix_kpr} and \figref{fig:appendix_neighbors} (Samples 2--6).}
\label{tab:emd_summary}
\begin{tabular}{ccc}
    \toprule
    Sample & KPR EMD (kernels) & Neighbor Count EMD (neighbors) \\
    \midrule
    1 & 1.25 & 0.33 \\
    2 & 0.75 & 0.84 \\
    3 & 1.21 & 0.41 \\
    4 & 7.58 & 0.61 \\
    5 & 2.62 & 0.83 \\
    6 & 4.85 & 0.42 \\
    \midrule
    Mean & 3.04 & 0.57 \\
    \bottomrule
\end{tabular}
\end{table}

\FloatBarrier

\subsection{Computational Cost and Throughput}
\label{sec:compute}

The automated workflow was deployed at scale across all 1{,}091 ears to evaluate pipeline scalability. All downstream analytical stages, including cylindrical unwrapping, CLAHE contrast enhancement, CPSAM segmentation, BPA surface reconstruction, convex hull volume, hue and aspect ratio extraction, Gabriel graph packing, and FFT/row-chain-based row and kernel-per-row estimation, were executed on the Iowa State University Nova high-performance computing cluster.

A detailed breakdown of the processing throughput and hardware configurations per ear is provided in \tableref{tab:compute_cost}. Downstream trait extraction, measured from the calibrated, PCA-aligned point cloud (\secref{sec:pca}) through final trait output, averaged approximately 10 seconds per ear using a single NVIDIA A100 GPU. Applied across the full dataset of 1{,}091 ears, this component requires a total wall-clock time of approximately 3 hours. This processing footprint occurs downstream of the upstream video-to-point-cloud reconstruction stage (COLMAP~\citep{schoenberger2016sfm} pose estimation and NeRF training), which follows \citet{young2026lowcost}; NeRF training alone averages approximately 14 minutes per ear in that platform.

% ============================================================
% COMPUTATIONAL COST TABLE
% ============================================================
\begin{table}[h!]
\centering
\caption{Computational cost per ear, by pipeline stage.}
\label{tab:compute_cost}
\setlength{\tabcolsep}{4pt}
\renewcommand{\arraystretch}{0.85}
\begin{tabular}{lcc}
\toprule
\textbf{Stage} & \textbf{Hardware} & \textbf{Time per ear} \\
\midrule
NeRF training$^{a}$
& NVIDIA A100
& $\sim$14 min \\

Trait extraction (this work)
& NVIDIA A100
& $\sim$10 s \\
\bottomrule
\multicolumn{3}{l}{\footnotesize $^{a}$NeRF training only, following \protect\citet{young2026lowcost}.}
\end{tabular}
\end{table}

\FloatBarrier

Of the 268-ear labeled dataset, 100 ears were used for parameter tuning and the remaining 168 reserved as a held-out set; the tuned pipeline was subsequently applied to all 1{,}091 reconstructed ears. On the 168-ear held-out set, the pipeline achieved a kernel count $R^{2}$ of 0.921 ($\mathrm{MAPE} = 10.33\%$) and KRN within $\pm$2 rows for 95.2\% of ears, operating on point clouds from the low-cost turntable imaging platform of \citet{young2026lowcost} without task-specific model training. Trait extraction downstream of point cloud reconstruction required approximately 10 seconds per ear on a single NVIDIA A100 GPU (\secref{sec:compute}), or roughly 3 hours in total across the full 1{,}091-ear dataset, supporting the pipeline's suitability for breeding-scale deployment when paired with the low-cost acquisition hardware of \citet{young2026lowcost}.

% ============================================================
\section{Discussion and Conclusions}
\label{sec:discussion}

The three validation tiers, synthetic, held-out, and manually annotated, together isolate distinct sources of pipeline error, and the sections below interpret each in turn before addressing the pipeline's structural limitations and broader implications for breeding-scale phenotyping.

Triple-juxtaposed unwrapping eliminated seam-boundary double-counting, a failure mode inherent to naive cylindrical projection of roughly cylindrical objects such as maize ears. Most row count prediction error fell within two rows, and larger deviations arose only for ears with non-aligned rows, where the angular regularity assumption underlying the FFT approach is violated. The same constraint appeared in the six-ear manually annotated subset: the adaptive FFT predicted the exact KRN for four of the six ears, and the remaining two deviated by two rows because of minor axial row non-alignment. This same six-ear subset separates the two packing-related outputs cleanly (\tableref{tab:emd_summary}). Gabriel graph neighbor-count distributions agree closely with manual annotation on every ear, whereas kernels-per-row distributions agree less consistently, with the disagreement concentrated in Samples 4 and 6 (\secref{sec:manual_validation}). Those are precisely the two ears whose KRN was underpredicted as 16 against a ground truth of 18 (\tableref{tab:krn_accuracy}). This coupling is structural: the row chain graph merges chains until exactly $\hat{R}$ remain (\secref{sec:ear_traits}), so an underestimated $\hat{R}$ forces kernels from two true rows into a single predicted chain, inflating the predicted kernels-per-row for the merged chain and shifting the distribution away from ground truth. KPR accuracy is therefore conditional on correct KRN, and the KPR EMD reflects the downstream consequence of the row-detection limitation. Sample 5 departs from this pattern, showing moderate KPR disagreement despite correct KRN. Neighbor count, computed directly from the Gabriel graph on kernel centroids without reference to $\hat{R}$ (\secref{sec:kernel_traits}), is unaffected by this coupling, consistent with its uniformly low EMD.

Turning from row structure to kernel count itself, the kernel count MAPE of 10.33\% exceeds that reported by several 2D consumer-hardware methods, consistent with reconstruction noise compounding through cylindrical projection. NeRF reconstructions of tightly packed kernels blur the boundaries between adjacent kernels, so CPSAM sometimes merges two true kernels into one segment (undercounting) and sometimes splits a single kernel's blurred boundary into two segments (overcounting). Residuals in \figref{fig:kernel_count_results} widen with increasing ground-truth kernel count, consistent with a proportional rather than additive error. Stratifying held-out KC error by ground-truth KRN (\tableref{tab:krn_stratified_kc_error}) shows MAPE ranging from 8.26\% (KRN$=14$) to 14.62\% (KRN$=10$) across the five row classes with more than ten ears each ($n$, the number of ears in each KRN class, ranges from 13 to 46 among these five). KRN $=16$ and $18$ show the highest absolute error (MAE $=41.41$ and $40.54$ kernels), roughly double that of the adjacent KRN $=12$ and $14$ groups (MAE $=15.27$ and $20.43$) despite comparable ear counts ($n=44$, $46$, $44$, and $13$, respectively, for KRN $=12, 14, 16, 18$). Signed ME is near zero for KRN $=14$ ($-0.48$) but strongly positive for KRN $=10$ and $18$ ($+10.55$ and $+22.38$), indicating overprediction rather than symmetric scatter in those two classes. The smaller ear count for KRN $=18$ ($n=13$ ears) and the single KRN $=20$ ear ($n=1$) warrant caution in interpreting those estimates.

Beyond count and row accuracy, the pipeline recovers packing and volumetric traits that have not previously been jointly extracted from a consumer imaging workflow. Gabriel graph packing traits (neighbor count, mean neighbor distance) and volumetric traits (surface area and convex hull volume proxy) describe spatial organization that has been directly linked to the positional patterning of kernel development and abortion along the ear~\citep{oury2022}. Mean neighbor distance provides a continuous measure of local kernel packing density, with smaller values indicating more tightly packed kernels, and Gabriel graph neighbor count provides a complementary per-kernel measure of the same property. Structured-light scanning and micro-CT imaging recover volumetric kernel traits at higher resolution~\citep{sun2026mep3d, zhao2025}; the present pipeline instead provides accessible approximations of these traits from consumer-grade turntable video, extending 3D kernel phenotyping to breeding programs without specialist equipment or large capital investment. This dimensional richness is obtained at a measurable cost in count accuracy, however: \citet{fan2026openear} report a kernel count $R^{2}$ of 0.980 on consumer hardware, exceeding the 0.921 achieved here, and \citet{shi2022} report a lower KRN MAE using specialist cameras. Both operate on 2D projections of the ear surface and recover neither volumetric shape proxies nor packing geometry, so the comparison is best read as a trade: projection-based methods optimize the accuracy of the traits a projection can express, whereas the present pipeline accepts higher error on those same traits in exchange for the radial dimension that a projection discards. Which trade is preferable depends on whether a breeding program requires count precision or trait breadth.

Three further considerations bound the interpretation of the traits reported here: physical occlusion, the abstractions built into the synthetic validation dataset, and the generalizability of the imaging conditions. The first is physical occlusion. Because the camera stream captures only the visible outer surface cap of individual kernels, the lateral contact boundaries and the proximal surfaces embedded in the central cob remain hidden, and the derived convex hull volume proxy and axis-aligned bounding box under-observe the radial dimension in depth. These metrics are therefore relative shape descriptors and structural spatial proxies rather than absolute measurements: absolute kernel volume requires destructive shelling or non-destructive X-ray micro-CT, and their absolute fidelity remains untested against physical measurement. The synthetic validation establishes only that the pipeline recovers these descriptors faithfully relative to a known generating geometry. The second consideration concerns the synthetic dataset itself. The KRN and KPR MAE on the 27-ear synthetic set (7.26 rows and 11.28 kernels, \tableref{tab:pooled_metrics}) exceed those measured on the real held-out set by a wide margin, and the kernel count $R^{2}$ likewise falls to 0.769. The discrepancy is isolated to the frequency analysis step, and the elevated KPR error follows from the elevated KRN error by the same chain-merging mechanism described above for Samples 4 and 6. This performance gap is, in fact, the reverse of the typical domain-transfer problem: the mathematical regularity of the synthetic ears penalizes the adaptive frequency-domain pipeline, whereas real-world biological noise aids it, as illustrated by the harmonic-locking failure shown in \figref{fig:synthetic_fft_failure}. Highly idealized synthetic lattices, while necessary for establishing exact ground truth, introduce sharp phase boundaries and rigid structural symmetries that real ears do not exhibit. Natural biological variation and physical kernel crowding smooth out these discontinuities in real maize, so real-world ears suppress the high-frequency harmonics that mislead the peak-selection step on synthetic data. Future synthetic validation frameworks for spatial-frequency tasks may need to model biological stochasticity, such as localized kernel jitter or growth irregularities, to better approximate real-world performance. Given this gap, the synthetic dataset should be read as validating the geometric and color traits under exact ground truth, with the row-structure traits evaluated on the real held-out and manually annotated sets instead. The synthetic baseline also introduces structural abstractions: neighbor configurations are evaluated via lattice-topological adjacency instead of a Gabriel graph, and hue is defined by population mean parameters rather than the specific per-kernel draw. These control states validate the pipeline's capacity to capture relative packing density and topology rather than serving as an exact edge-by-edge or per-kernel benchmark. The third and final consideration concerns generalizability: the 1{,}091 ears were imaged on a controlled turntable platform under consistent lighting, and may not span the full range of maize ear morphologies or field conditions, so performance on varieties with markedly different kernel color, ear curvature, or packing density has not been tested here.

Beyond these bounding considerations, the axial profiles reveal spatial structure that a single per-ear mean discards. Across the four representative ears, mean aspect ratio vs.\ height profiles (\figref{fig:ar_by_height}) show modestly higher kernel elongation in the middle height bands relative to the base and tip. This trend persists at the population scale: across all 1{,}091 analyzed ears, the ear-level averaged kernel aspect ratio peaks in the middle third of the cob and decreases toward both boundaries (\tableref{tab:ar_by_thirds}), although the between-band difference is smaller than the within-band standard deviation. This pattern is consistent with a physiological gradient along the axis, potentially reflecting the sequential nature of pollination or localized resource competition during the grain-filling window rather than a uniform, ear-wide morphological constraint. Hue profiles reveal a parallel axial structure: mean hue vs.\ height profiles (\figref{fig:hue_by_height}) show color gradients that vary subtly from ear to ear along the developmental axis, and where present, these gradients may reflect positional variation in carotenoid content, as kernel carotenoid concentration has been shown to vary with position along the cob~\citep{calvobrenes2019}.

This study presents a fully automated, high-throughput pipeline extracting eleven ear- and kernel-level traits (total kernel count, KRN, KPR, surface area, convex hull volume proxy, Gabriel graph neighbor count, mean neighbor distance, circular mean kernel hue, 3D bounding box aspect ratio, mean hue vs.\ height, and mean aspect ratio vs.\ height) from calibrated 3D point clouds produced by the turntable imaging platform of \citet{young2026lowcost}, with no task-specific training. Validation spanned three tiers: the 27-ear geometric synthetic dataset, the six-ear manually annotated subset, and the 168-ear held-out set. Across these tiers, the pipeline recovers kernel count within 10.33\% MAPE, KRN within $\pm$2 rows for 95.2\% of ears, and kernel packing topology within a mean Earth-Mover Distance of 0.57 neighbors, using consumer-accessible imaging hardware and no supervised crop-specific models. Applied to 1{,}091 ears with known genotype identity, the pipeline yields a multi-trait dataset positioned for phenotype-to-genotype association analyses. That dataset spans both heritable traits such as KRN and traits sensitive to environmental stress and management such as KPR \citep{peng2011, liu2015krn, andrade2002}. No association analysis is attempted here; applying these phenotypes to genotype selection and agronomic management decisions is left to future work.

% ============================================================
\section*{Acknowledgments}
We thank Ms. Lisa Coffey of the Schnable Lab, who supervised a team of undergraduate assistants who imaged ears and generated ground truth data. Team members included: Owen Hildebrandt, Roland Stetz, Isaac Tharp, Haylee Thomas, and Ryan Wolf.

\section*{Funding Information} This work was supported by the AI Institute for Resilient Agriculture (USDA-NIFA 2021-67021-35329) and Iowa State University's Plant Science Institute.

% ============================================================
\section*{Conflict of Interest}
The authors declare no conflict of interest.

% ============================================================
\section*{ORCID}
Ritwesh A. Kumar: \url{https://orcid.org/0009-0006-1956-5273} \\
Talukder Zaki Jubery: \url{https://orcid.org/0000-0003-4107-1617} \\
Patrick Schnable: \url{https://orcid.org/0000-0001-9169-5204} \\
Adarsh Krishnamurthy: \url{https://orcid.org/0000-0002-5900-1863} \\
Baskar Ganapathysubramanian: \url{https://orcid.org/0000-0002-8931-4852}

% ============================================================
% \section*{Data Availability}
%Optional section: required if depositing a dataset to Dryad or other data depository.

\clearpage

% ============================================================
\bibliographystyle{plainnat}
\bibliography{Refs}

\clearpage
\appendix
\renewcommand{\thefigure}{A.\arabic{figure}}
\setcounter{figure}{0}
\setcounter{table}{0}
% ============================================================
\section*{Supplementary Material}
\label{sec:supplemental}

\section*{Manual Annotation Protocol}

\begin{figure}[htbp]
\centering
\includegraphics[height=0.4\textheight, keepaspectratio=true]{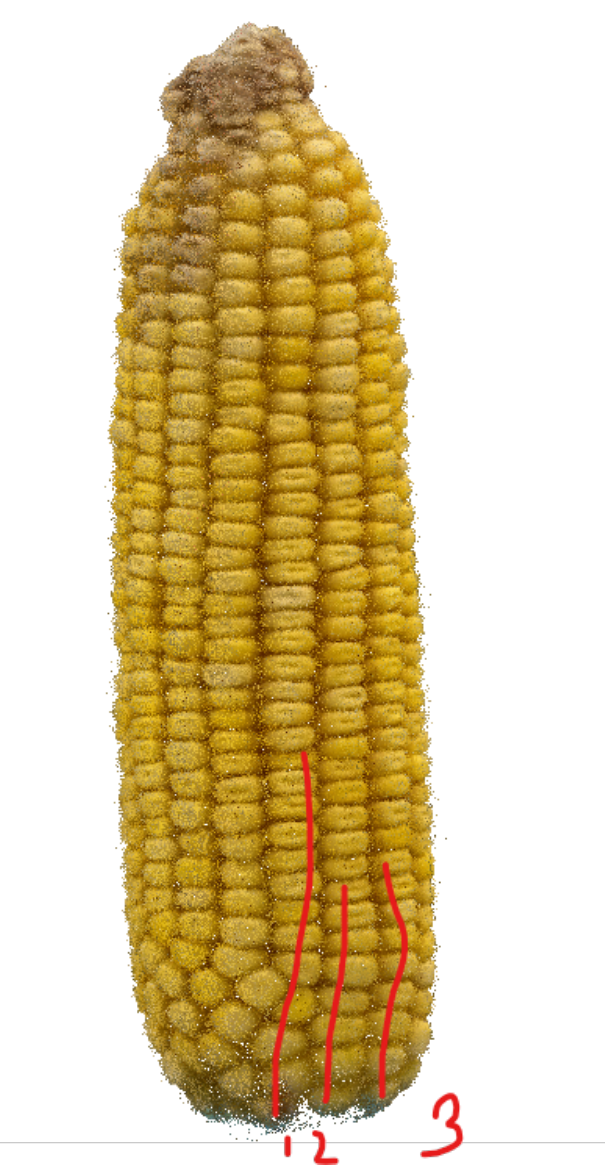}
\caption{Kernel row and kernel-position numbering protocol used for manual annotation of the six-ear manually annotated subset, illustrated for Sample 1. Traced lines mark individual rows (labeled 1, 2, 3, ...), numbered from the ear base to the tip; kernel position within each row was numbered in the same direction. All six ears were annotated according to this convention.}
\label{fig:annotation_protocol}
\end{figure}

As described in \secref{sec:dataset_composition}, both manually annotated datasets used in this study, the 268-ear labeled dataset and the six-ear manually annotated subset, followed the same base-to-tip numbering convention for kernel rows, with row 1 nearest the ear base and row numbers increasing toward the tip. Rows were counted sequentially around the full circumference of the ear, starting from an arbitrary row designated as row 1; the starting point does not affect the reported kernel row number or kernels-per-row values, which depend only on the total count and per-row kernel assignment, not on which physical row was labeled first. \figref{fig:annotation_protocol} illustrates this convention for Sample 1 of the six-ear subset: each kernel row is traced and numbered starting from row 1 at the base, with kernel position within a row numbered in the same base-to-tip direction. The two datasets differ in annotation depth and annotator count.

For the 268-ear labeled dataset, a single expert annotator recorded only total kernel count (KC) and kernel row number (KRN) per ear, without resolving individual kernel positions within each row. For the six-ear manually annotated subset, three independent annotators (two ears each) additionally recorded full per-row kernel identity and position, following the same base-to-tip convention illustrated here, enabling the exhaustive kernels-per-row and packing validation reported in \secref{sec:manual_validation}. This numbering was applied identically across all six ears in that subset, so that per-row kernel identity and position were unambiguous and consistent with the convention despite each ear being labeled by a single individual.

\section*{Additional Distribution Plots}

\begin{figure}[h!]
\centering
\begin{subfigure}[b]{0.49\textwidth}
    \centering
    \includegraphics[width=\linewidth]{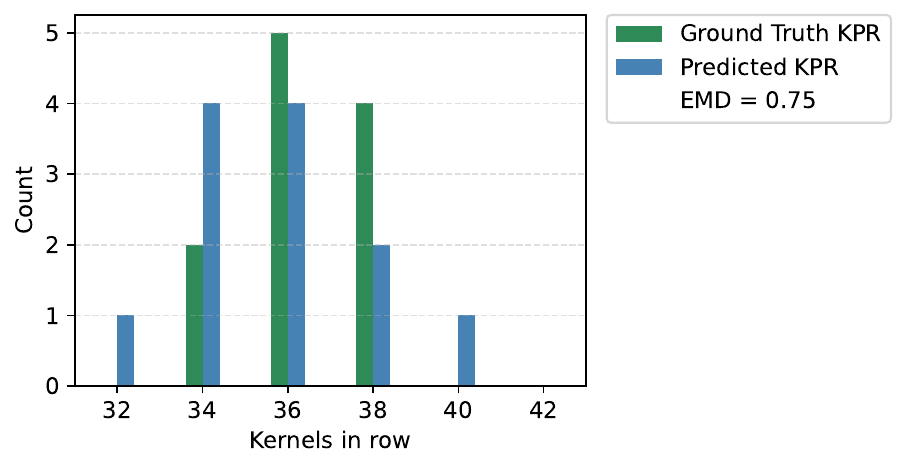}
    \caption{Sample 2: KPR distribution}
\end{subfigure}
\begin{subfigure}[b]{0.49\textwidth}
    \centering
    \includegraphics[width=\linewidth]{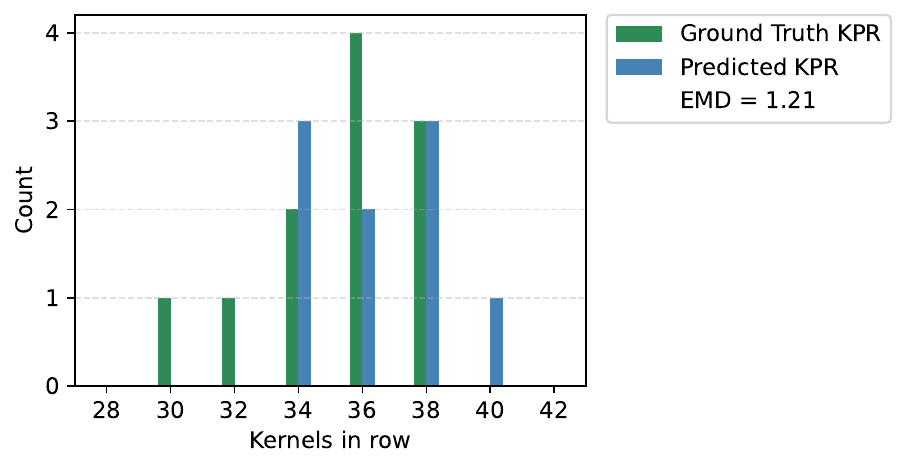}
    \caption{Sample 3: KPR distribution}
\end{subfigure}
\begin{subfigure}[b]{0.49\textwidth}
    \centering
    \includegraphics[width=\linewidth]{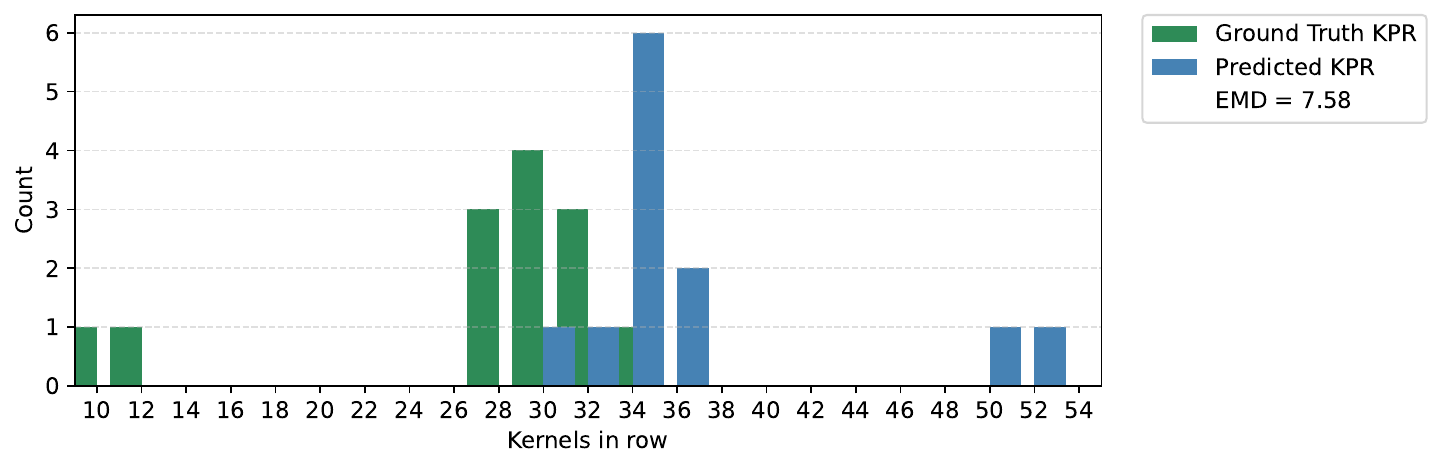}
    \caption{Sample 4: KPR distribution}
\end{subfigure}
\begin{subfigure}[b]{0.49\textwidth}
    \centering
    \includegraphics[width=\linewidth]{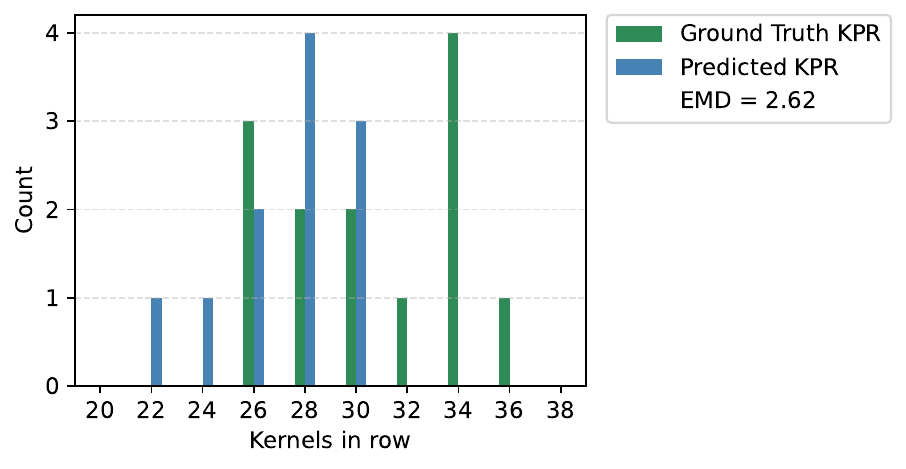}
    \caption{Sample 5: KPR distribution}
\end{subfigure}
\begin{subfigure}[b]{0.49\textwidth}
    \centering
    \includegraphics[width=\linewidth]{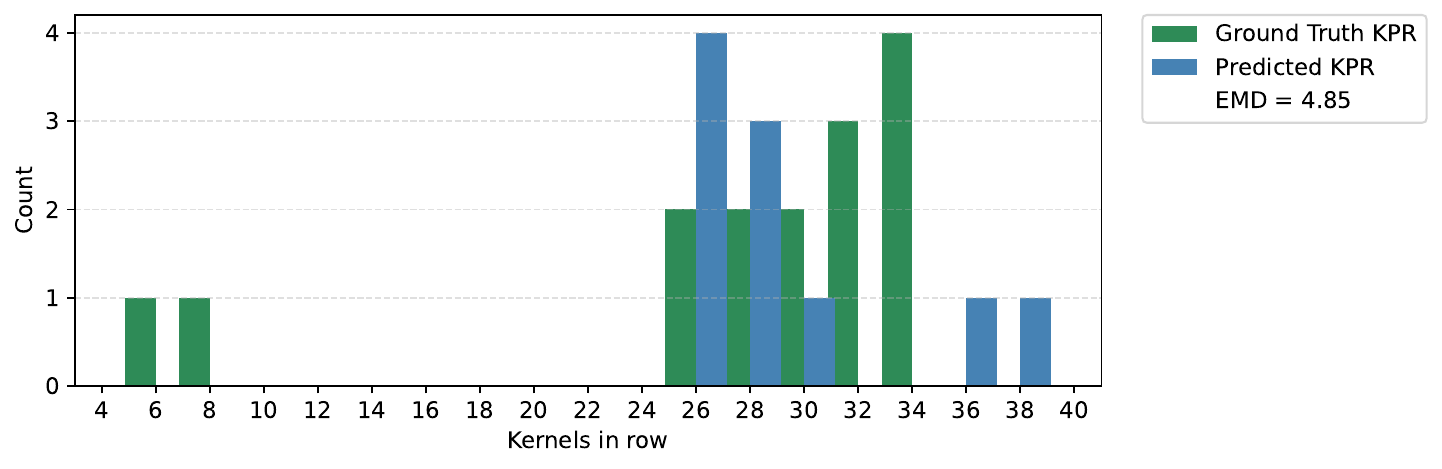}
    \caption{Sample 6: KPR distribution}
\end{subfigure}
\caption{KPR accuracy distributions for Samples 2--6.}
\label{fig:appendix_kpr}
\end{figure}

\begin{figure}[t!]
\centering
\begin{subfigure}[b]{0.49\textwidth}
    \centering
    \includegraphics[width=\linewidth]{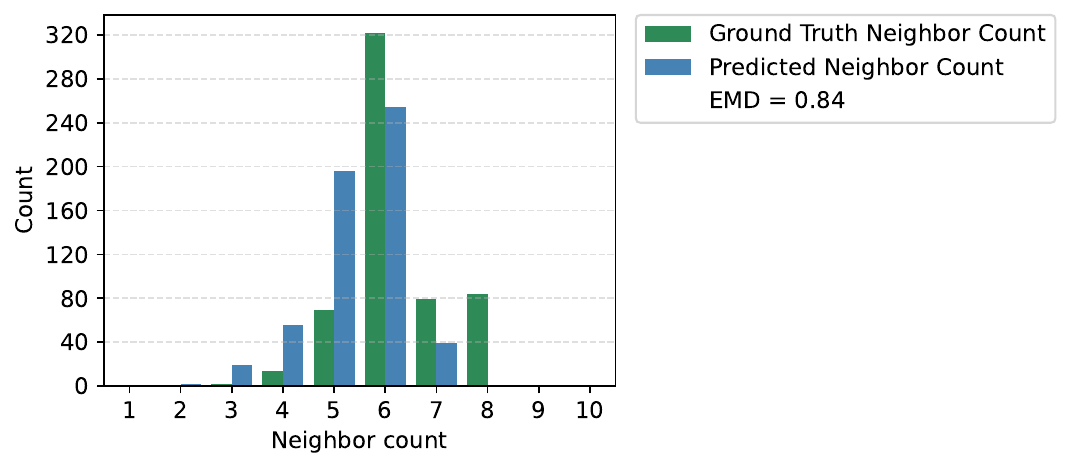}
    \caption{Sample 2: Neighbor count distribution}
\end{subfigure}
\begin{subfigure}[b]{0.49\textwidth}
    \centering
    \includegraphics[width=\linewidth]{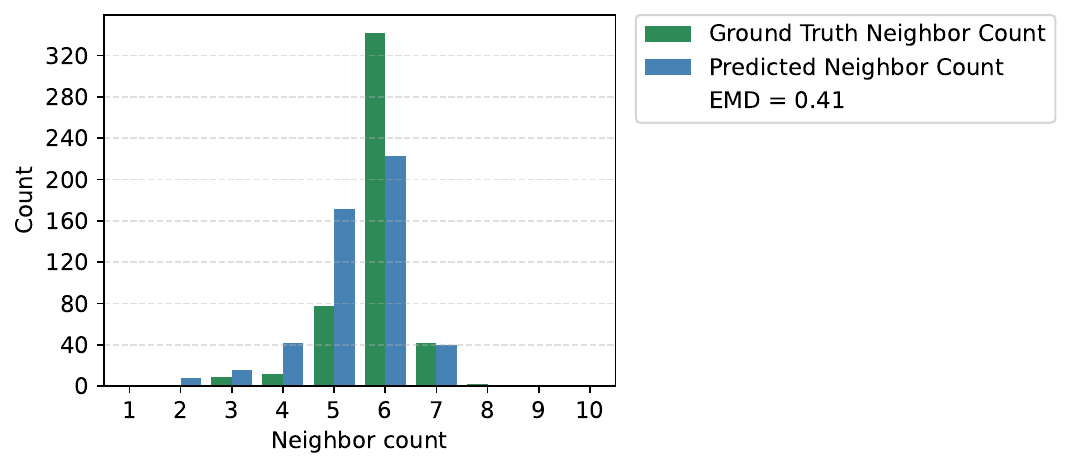}
    \caption{Sample 3: Neighbor count distribution}
\end{subfigure}
\begin{subfigure}[b]{0.49\textwidth}
    \centering
    \includegraphics[width=\linewidth]{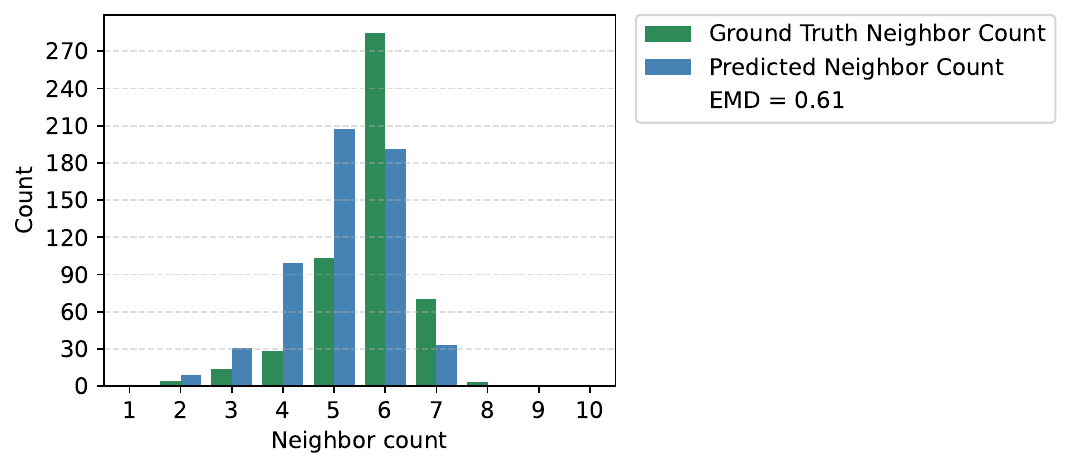}
    \caption{Sample 4: Neighbor count distribution}
\end{subfigure}
\begin{subfigure}[b]{0.49\textwidth}
    \centering
    \includegraphics[width=\linewidth]{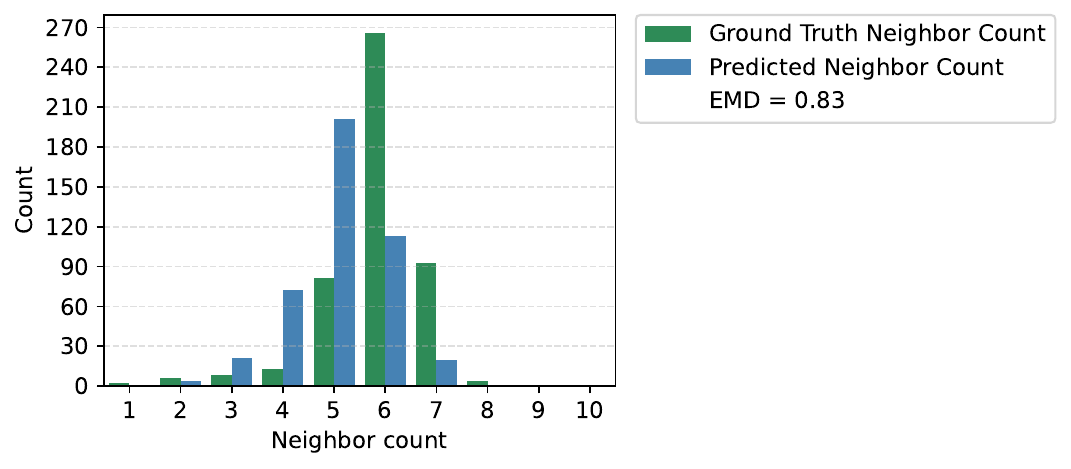}
    \caption{Sample 5: Neighbor count distribution}
\end{subfigure}
\begin{subfigure}[b]{0.49\textwidth}
    \centering
    \includegraphics[width=\linewidth]{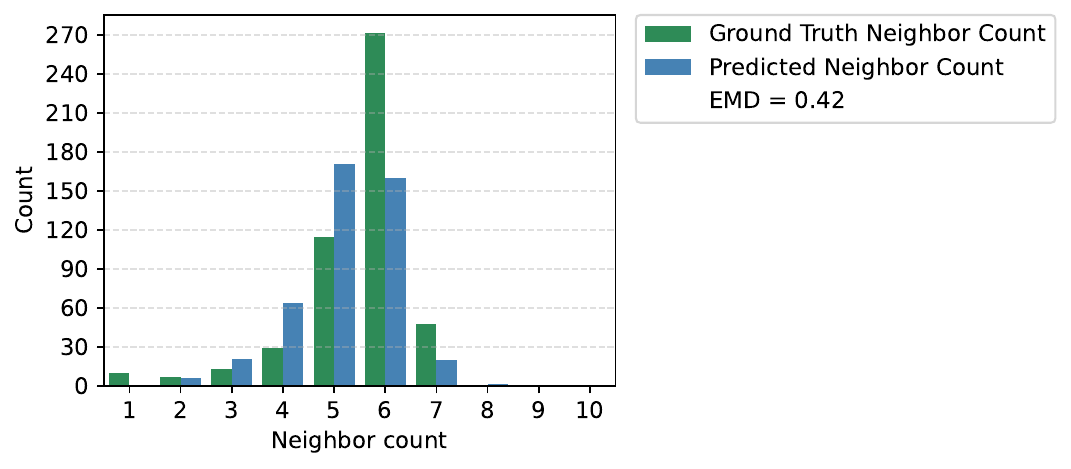}
    \caption{Sample 6: Neighbor count distribution}
\end{subfigure}
\caption{Neighbor-count accuracy distributions for Samples 2--6.}
\label{fig:appendix_neighbors}
\end{figure}

This section contains the additional distribution plots validating the tracking and packing performance of the automated pipeline. While Sample 1 is detailed above, \figref{fig:appendix_kpr} and \figref{fig:appendix_neighbors} complete the evaluation across the entire six-ear manually annotated subset introduced in \secref{sec:manual_validation}. Specifically, they present the remaining kernels-per-row (KPR) and Gabriel graph neighbor-count accuracy distributions for Samples 2--6. The corresponding Earth-Mover Distances are collected in \tableref{tab:emd_summary}.

% ==========================================
% SECTION 1: ALL KPR ACCURACY DISTRIBUTIONS
% ==========================================

% ==========================================
% SECTION 2: ALL NEIGHBOR COUNT DISTRIBUTIONS
% ==========================================

\clearpage

\section*{Corn Ear 3D Reconstruction Using Stationary Camera}
\label{sec:video_to_pcd}
This section outlines the steps, adapted from \citet{young2026lowcost}, for reconstructing the corn ear in 3D from a stationary camera.

\subsubsection*{360\textdegree{} Video Acquisition}
\label{sec:video}

\begin{figure}[t!]
\centering
\begin{subfigure}[b]{0.22\textwidth}
    \includegraphics[width=\textwidth]{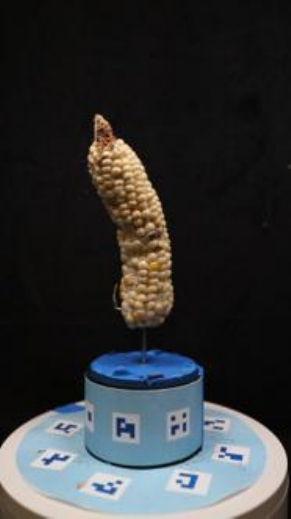}
    \caption{24-P2042\_3}
\end{subfigure}\hfill
\begin{subfigure}[b]{0.22\textwidth}
    \includegraphics[width=\textwidth]{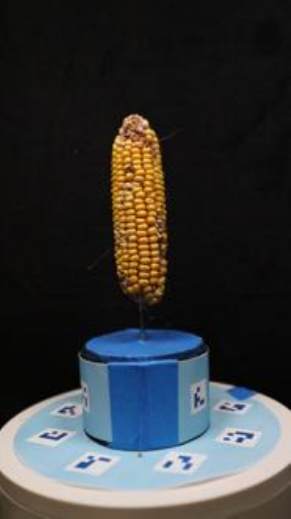}
    \caption{24-P2071\_2}
\end{subfigure}\hfill
\begin{subfigure}[b]{0.22\textwidth}
    \includegraphics[width=\textwidth]{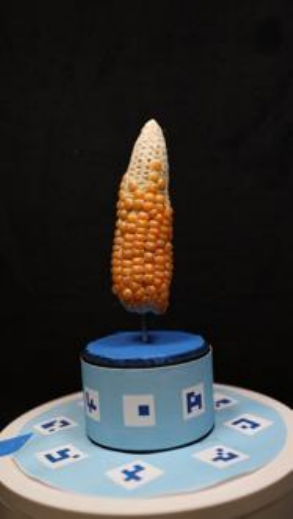}
    \caption{24-P2146\_2}
\end{subfigure}\hfill
\begin{subfigure}[b]{0.22\textwidth}
    \includegraphics[width=\textwidth]{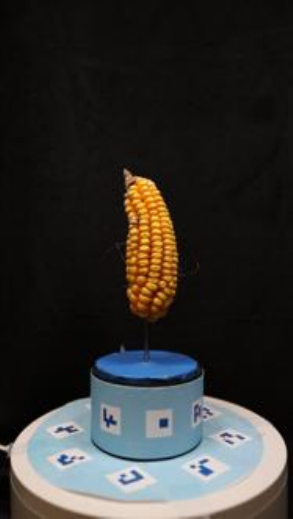}
    \caption{24-P2199\_1}
\end{subfigure}
\caption{Representative video frames showing each of the four corn ears mounted on the motorized rotating turntable prior to 360\textdegree{} video acquisition, following the imaging setup of \protect\citet{young2026lowcost}. The known diameter of the cylindrical sample holder provided the metric reference for subsequent distance calibration.}
\label{fig:turntable}
\end{figure}

Each ear was placed on a motorized rotating turntable against a dark background. A stationary camera captured a continuous 360\textdegree{} video as the ear completed a full rotation. The known diameter of the cylindrical sample holder served as a metric reference for subsequent distance calibration. Raw video frames were extracted and used as input to the 3D reconstruction pipeline. \figref{fig:turntable} shows representative frames of each of the four representative ears mounted on the turntable prior to acquisition.

% ------ Video / turntable setup ------

\subsubsection*{Camera Pose Estimation via COLMAP}

Raw video frames were processed using COLMAP \citep{schoenberger2016sfm}, a Structure-from-Motion (SfM) and Multi-View Stereo (MVS) pipeline. COLMAP estimated the intrinsic and extrinsic camera parameters for each frame, producing a sparse 3D reconstruction and a set of camera poses describing the camera's position and orientation relative to the scene at each frame. These poses served as input to the NeRF reconstruction described next.

\subsubsection*{Neural Radiance Field Reconstruction}

A Neural Radiance Field (NeRF)~\citep{mildenhall2021nerf} was trained on the video frames and their corresponding camera poses to produce a dense, photorealistic 3D representation of the scene. The trained NeRF was then queried to extract a dense 3D point cloud of the full scene, including the ear and turntable.

\subsubsection*{Ear Isolation via Density-Based Separation}

The raw scene point cloud contains points from both the turntable and the corn ear, which differ in spatial point density: the corn ear forms a compact, high-density cluster, whereas the turntable and surrounding structures are more sparsely distributed. A density-based separation method retained only the high-density cluster corresponding to the ear.

\subsubsection*{Distance Calibration}
\label{sec:calibration}

Because NeRF reconstructions are defined up to an arbitrary scale factor, the point cloud was calibrated to physical units (millimeters) using the known diameter of a cylindrical sample holder visible in every frame. The scale factor was computed as the ratio of the known holder diameter to its measured diameter in the reconstructed point cloud, and all point coordinates were multiplied by this factor to yield a metrically accurate point cloud in millimeters.

\end{document}